\documentclass[twoside,11pt]{article}

\usepackage[abbrvbib, preprint]{jmlr2e}

\usepackage{jmlr2e}

\usepackage{url}

\usepackage{xspace}
\newcommand{\ourdata}{\textsc{Prism}\xspace}
\newcommand{\ourdatax}{\textsc{Prism-X}\xspace}
\usepackage{color,soul}
\usepackage{amsmath}
\usepackage{wrapfig}

\usepackage[nameinlink,capitalise]{cleveref} %
\crefformat{appendix}{#2App.~#1#3} %
\crefname{table}{Tab.}{Tab.}
\crefname{appendix_table}{Tab.}{Tab.}
\crefname{Table}{Tab.}{Tab.}
\crefname{Figure}{Fig.}{Fig.}
\crefname{figure}{Fig.}{Fig.}
\crefformat{section}{\S\,#2#1#3}
\Crefformat{section}{\S\,#2#1#3}

\usepackage{enumitem}

\usepackage[export]{adjustbox} %
\usepackage{overpic} %

\usepackage{colortbl} %
\usepackage{mdframed} %
\usepackage{framed} %
\usepackage{makecell} %
\usepackage{multirow}
\usepackage{booktabs}
\usepackage{xfp}
\usepackage[table, dvipsnames]{xcolor}
\usepackage{longtable}
\usepackage{amssymb}%
\usepackage{pifont}%
\newcommand{\cmark}{\ding{51}}%
\newcommand{\xmark}{\ding{55}}%
\usepackage{hhline}
\usepackage{tabularray}

\usepackage[]{color-edits}
\addauthor{ll}{blue}

\definecolor{myyellow}{RGB}{255,194,76}
\definecolor{myred}{RGB}{255,138,103}
\definecolor{myredl}{RGB}{255,173,149}
\definecolor{myteal}{RGB}{50,188,221}
\definecolor{myblue}{RGB}{50,188,221}
\definecolor{mygreen}{RGB}{183,229,202}
\definecolor{mygreend}{RGB}{144,213,172}
\definecolor{myoat}{RGB}{254,250,233}
\definecolor{mypurple}{RGB}{143,61,143}
\definecolor{mybeige}{RGB}{234,225,205}
\definecolor{violinred}{RGB}{255,145,145}
\definecolor{violinblue}{RGB}{153,153,255}
\definecolor{mypink}{RGB}{232,72,148}
\definecolor{myrasp}{RGB}{218,44,96}
\colorlet{lightoat}{myoat!40!white} %
\colorlet{lightred}{myred!40!white} %
\colorlet{lightblue}{myblue!40!white}
\colorlet{lightpurple}{mypurple!40!white}
\colorlet{lightteal}{myteal!40!white}
\colorlet{darkyellow}{myyellow!90!black}
\definecolor{mygray}{RGB}{197,197,197}
\definecolor{heatmapblue}{RGB}{165,187,208}
\definecolor{heatmapgreen}{RGB}{171,199,171}

\usepackage{booktabs}
\usepackage{array}

\usepackage{float}

\def\origpcb#1{%
   {\color{myred}\rule{\fpeval{#1/\percentscale*\medbarwidth} cm}{\barheight}}~#1
}

\def\followpcb#1{%
   {\color{myrasp}\rule{\fpeval{#1/\percentscale*\medbarwidth} cm}{\barheight}}~#1
}

\newcommand{\medbarwidth}{1.5} %
\newcommand{\barheight}{7.5pt} %
\newcommand{\percentscale}{100} %

\usepackage{lastpage}
\jmlrheading{}{}{1-\pageref{LastPage}}{4/21/2026}{}{}{Kirk, Liu, Zeng, Davidson, Vidgen, Summerfield and Hale}

\usepackage{fancyhdr}

\usepackage{amsmath}

\usepackage{mdframed}
\usepackage{tikz}
\usepackage{listings}
\usepackage{fancyvrb}
\mdfdefinestyle{promptblank}{
    linecolor=black,
    linewidth=1pt,
    backgroundcolor=gray!20,
    innerleftmargin=10pt,
    innerrightmargin=10pt,
    innertopmargin=8pt,
    innerbottommargin=8pt,
    skipabove=10pt,
    skipbelow=15pt,
    frametitle={},
    nobreak=true,
    splitbottomskip=5pt,
    splittopskip=5pt,
    font=\footnotesize
}

\newcounter{aipromptbox}
\newenvironment{aipromptbox}[1][AI Prompt]{%
    \stepcounter{aipromptbox}%
    \mdfsetup{%
        frametitle={%
            \tikz[baseline=(current bounding box.east), outer sep=0pt] \node[anchor=east,rectangle,fill=black]
            {\color{white} \strut #1};}
    }%
    \mdfsetup{%
        innertopmargin=10pt,
        linecolor=black,
        linewidth=1pt,
        backgroundcolor=white,
        topline=true,
        frametitleaboveskip=\dimexpr-\ht\strutbox\relax,
        skipabove=15pt,
        skipbelow=15pt,
        splitbottomskip=5pt,
        splittopskip=5pt
    }%
    \begin{mdframed}[]
}{%
    \end{mdframed}
}

\usepackage{tocloft}
\usepackage{minitoc}

\usepackage{tocbibind} %
\usepackage[toc,page]{appendix} 

\usepackage{caption}

\newcommand{\ourtitle}{\textcolor{myred}{\textsc{P}}\textcolor{myyellow}{\textsc{R}}\textcolor{mygreend}{\textsc{I}}\textcolor{myteal}{\textsc{S}}\textcolor{mypurple}{\textsc{M}}\xspace}

\ShortHeadings{\ourtitle-\textsc{X}: Experiments on Personalised Fine-Tuning with Human and Simulated Users}{Kirk et al.,}
\firstpageno{1}

\begin{document}

\doparttoc %
\faketableofcontents %

\part{} %

\title{
\ourtitle-\textsc{X}: Experiments on Personalised Fine-Tuning with Human and Simulated Users%
}

\author{\name Hannah Rose Kirk \email hannah.kirk@oii.ox.ac.uk \\
       \addr University of Oxford; UK AI Security Institute 
       \AND
       \name Liu Leqi \email leqiliu@utexas.edu \\
       \addr University of Texas at Austin
       \AND
       \name Fanzhi Zeng \email fanzhi.zeng@utexas.edu \\
       \addr University of Texas at Austin
       \AND
       \name Henry Davidson \email
       henry.davidson@dsit.gov.uk\\
       \addr UK AI Security Institute
       \AND
       \name Bertie Vidgen 
       \email bertie@mercor.com\\
       \addr Mercor; University of Oxford
       \AND
       \name Christopher Summerfield
       \email christopher.summerfield@dsit.gov.uk
       \\
       \addr University of Oxford; UK AI Security Institute
       \AND
       \name Scott A.\ Hale \email scott.hale@oii.ox.ac.uk \\
       \addr University of Oxford; Meedan}

\editor{Kilian Weinberger}

\maketitle

\begin{abstract}%
Personalisation is a standard feature of conversational AI systems used by millions; yet, the efficacy of personalisation methods is often evaluated in academic research using simulated users rather than real people. This raises questions about how users and their simulated counterparts differ in interaction patterns and judgements, as well as whether personalisation is best achieved through context-based prompting or weight-based fine-tuning. Here, in a large-scale within-subject experiment, we re-recruit 530 participants from 52 countries two years after they gave their preferences in the \ourdata dataset \citep{kirkPRISM2024} to evaluate personalised and non-personalised language models in blinded multi-turn conversations. We find preference fine-tuning \citep[][P-DPO]{liPersonalized2024} significantly outperforms both a generic model and personalised prompting but adapting to individual preference data yields marginal gains over training on pooled preferences from a diverse population. Beyond length biases, fine-tuning amplifies sycophancy and relationship-seeking behaviours that people reward in short-term evaluations but which may introduce deleterious long-term consequences. Replicating this within-subject experiment with simulated users recovers aggregate model hierarchies but simulators perform far below human self-consistency baselines for individual judgements, discuss different topics, exhibit amplified position biases, and produce feedback dynamics that diverge from humans.

\end{abstract}


\section{Introduction}

Personalisation is now a standard feature of AI services used by hundreds of millions of people \citep{openaiPower2025, googleGemini2025, anthropicUnderstanding2025}. Deployed systems typically personalise behaviour by injecting user context into the system prompt. A parallel line of research proposes learning user-specific representations directly from preference data, but these methods are typically evaluated in public research using LLM-simulated users rather than the people they are designed for \citep[e.g.,][]{jangPersonalized2023, liPersonalized2024, shenfeldLanguage2025,poddarPersonalizing2024,namLearning2026, chenPOPI2026}. Two foundational questions therefore remain: does personalised preference fine-tuning outperform prompting and generic fine-tuning when evaluated by real humans, and to what extent can the simulated evaluations used to develop these methods be trusted?

\begin{figure}[H]
\centering
\includegraphics[width=\linewidth]{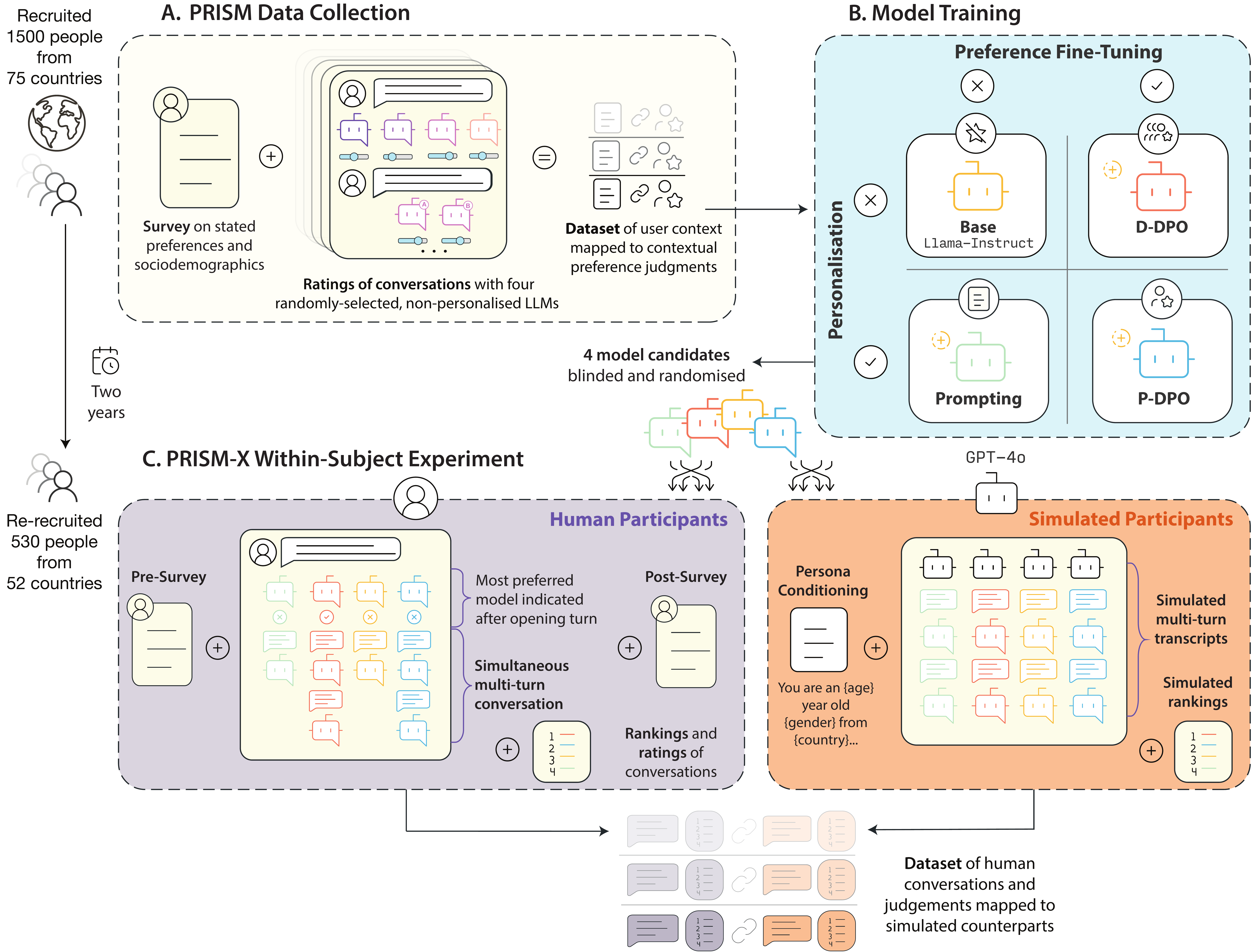}
\caption{\small \textbf{Study overview.} \textbf{(A)} The \ourdata dataset \citep{kirkPRISM2024}: 1,500 participants from 75 countries completed surveys detailing sociodemographics, AI usage, and stated preferences for AI behaviour, then conversed with ``off-the-shelf'' LLMs. In each conversation, participants rated four model responses to their prompt, then continued with the highest-rated model, yielding 8,011 conversations and 68,371 ratings linked to detailed user profiles. \textbf{(B)} The model training stage crosses preference fine-tuning with personalisation: personalised preference fine-tuning (PPFT) learns user-specific embeddings; diverse preference fine-tuning (DPFT) trains on pooled preferences across \ourdata; Prompting conditions receive natural language user context; the control (Base) is a generic model with no additional preference data (\cref{sec:personalization_methods}). \textbf{(C)} In \ourdatax, 530 participants from 52 countries were re-recruited for a within-subject experiment. In each experiment block, participants first selected their preferred model to an opening prompt, then engaged in multi-turn conversations with all four models, followed by an assessment phase with multiple ordinal and cardinal measures (\cref{sec:human_study}), yielding 8,473 conversations and over 67,821 preference judgements. We replicate this same experiment with persona-conditioned simulators (GPT-4o) seeded with user profiles from \ourdata, (\cref{sec:simulation}), yielding a further $\sim$8.5k simulated conversations and $\sim$25k simulated judgements.}
\label{fig:splash}
\end{figure}

Here we set out to answer both questions. We build on the PRISM alignment dataset \citep{kirkPRISM2024}, which contains preference ratings from 1,500 people across 75 countries matched to sociodemographic profiles. Using PRISM, we trained a personalised preference-tuned model to adapt to each user's feedback \citep[P-DPO,][]{liPersonalized2024} and a diverse preference-tuned model to aggregate across all users \citep[DPO,][]{rafailovDirect2024}. We compared these to two additional baselines: an off-the-shelf Llama-Instruct model \citep[with Meta's generic preference-tuning,][]{grattafioriLlama2024short} and prompting-based personalisation (\cref{fig:splash}). Two years after initial data collection, \ourdata participants returned for a within-subjects experiment (\ourdatax, $N=530$) in which each participant conversed with all four models blind across multiple conversation domains, providing ratings, rankings, willingness-to-pay bids, and behavioural signals. In parallel, we ran a twinned simulation using GPT-4o to role-play each participant on the same experiment materials, allowing a one-to-one comparison of the fidelity and pitfalls of simulating human preferences.

Four findings emerge. First, \textbf{personalised fine-tuning outperforms prompting despite using far fewer inference tokens}. Both fine-tuned models are preferred to prompting-based personalisation and the generic model, despite prompting consuming 38$\times$ more tokens at inference time, but the margin between personalised and diverse population-level fine-tuning is surprisingly small. 
Second, \textbf{preference fine-tuning amplifies behavioural traits}. Fine-tuned models become more sycophantic, more relationship-seeking, and more verbose. These are traits users reward in short-term evaluations but whose long-term consequences raise serious concerns. Third, \textbf{users prefer active personalisation to passive personalisation}. Surveyed attitudes are more favourable toward methods that elicit preferences explicitly (judgements, stated preferences) than methods that infer them passively (from demographics, chat history, or psychological profiles). Finally, \textbf{LLM simulators are not (yet) substitutes for real users}. Simulations recover coarse aggregate rankings but fall well below a human self-consistency baseline on individual judgements, discuss different topics, are far more homogeneous than real users, amplify position biases, and produce multi-turn dynamics unlike human interactions. Our contributions are as follows:
\begin{itemize}[itemsep=1pt, parsep=0pt,  topsep=0pt, leftmargin=1.25em]
    \item We fine-tune models using the \ourdata dataset, including P-DPO with user-specific soft-token embeddings learned from individual preference ratings.
    \item We collect, analyse, and release \ourdatax{}, a dataset of $\sim$8.5k conversations and over 67k preference judgements from a within-subjects experiment in which 530 participants across 52 countries evaluate personalised and non-personalised models. This data is linked to the original \ourdata profiles, providing longitudinal measures of LLM usage and stated preferences spanning a two year period of rapid AI adoption.
    \item We conduct a large-scale comparison of simulated and human evaluations matched on experiment design, documenting divergences in accuracy, biases, coverage and conversational dynamics. We release simulated data alongside paired human data to permit future work improving the fidelity of simulating human judgements from diverse populations.
\end{itemize}

\section{Personalisation and Preference Fine-tuning Methods}
\label{sec:personalization_methods}
We first briefly describe the personalisation setup, where we are given users’ preference data and potentially textual information on users’ demographics and preferences. 
Using these data, our goal is to generate responses that are better aligned with users’ individual preferences, such that users are more likely to prefer the generated responses over the one generated by the base model. For a given {personalised preference dataset} $\mathcal{D}_{\mathrm{p}} = \{(x_i, y_{i,1}, y_{i,2}, u_i)\}_{i=1}^n$ containing $n$ samples, each sample consists of a prompt $x_i$, two generated responses $y_{i,1}$ and $y_{i,2}$ where $y_{i,1}$ is preferred over $y_{i,2}$ (denoted as $y_{i,1} \succ y_{i,2}$) according to a user's preference, and user information $u_i$. The user information $u_i$ comprises textual data (e.g., demographics) and a user identifier. 
A personalised LLM $\pi_{\mathrm{p}}$ takes as input a system prompt $s$, a user query $x$, and a user context $\overline{u}$, and generates text tailored to the user's preferences:
$
y \sim \pi_{\mathrm{p}}(\cdot \mid s, \overline{u}, x) .
$
We note that the user context $\overline{u}$ is not necessarily equivalent to the user information $u$ provided in the dataset.
For example, $\overline{u}$ may be just the user identifier, a learned soft prompt for the user, the user's demographics, or a summary of the user's preferences derived from their preference dataset.

\subsection{The PRISM Alignment Dataset}
Few datasets exist with the required features for personalised preference learning. We use the \ourdata alignment dataset \citep{kirkPRISM2024}, which maps detailed demographic profiles and stated preferences of 1,500 participants across 75 countries, to their contextual preference judgements in multi-turn conversations with LLMs.

\paragraph{Data Collection Methodology.} Prior to interacting with models, the \ourdata participants completed a survey capturing a detailed profile, including their familiarity and usage patterns with LLMs; a 2--5 sentence ``constitution'' describing desired AI behaviour in open-ended form; a self-written description of personal values and guiding principles; and standard sociodemographics. Participants then held multi-turn conversations with LLMs, averaging 3.4 turns per conversation. In the opening turn, they wrote a free-text prompt and rated responses from up to four different LLMs on a visual analogue scale from ``Terrible'' to ``Perfect'', with the underlying 1--100 scores hidden to avoid anchoring. The highest-rated model was then locked in for the remainder of the conversation, with participants rating two further responses, sampled with non-deterministic temperature, at each subsequent turn.

\paragraph{Sample Characteristics.} The original 1,500 participants were recruited from Prolific via two complementary sampling strategies. For the U.K. and U.S., census-representative samples of 300 participants were matched to age, gender, and ethnicity quotas, providing demographic depth in two countries where such matching was feasible. For the rest of the world, 33 country-specific studies covered all global regions with balanced gender quotas where possible. The resulting sample is demographically and geographically diverse, though it skews toward White, Western, and English-speaking populations; full participant characteristics are reported in \cref{sec:appendix_ppt_chars}. Responses were drawn from 21 LLMs that were frontier at the time of collection (2023; 9 open-access and 12 commercial-API), spanning diverse providers, parameter sizes, and post-training pipelines. Models were instructed to keep responses under 50 words to control for length confounding and reduce participant fatigue.

\subsection{Models}
A key technical question is how to incorporate individual user data into the model. 
Broadly, personalisation can be achieved either \textit{explicitly} via prompts that provide user-specific information as additional input context at inference time; or \textit{implicitly} via weight-based approaches that encode user-specific behaviour in learned parameters. 
We consider both context-based and weight-based approaches to personalisation.

\subsubsection{Prompting-based Personalisation}
Prompting-based personalisation methods explicitly incorporate natural language user information into a textual prompt. We evaluate two prompting-based methods that represent different levels of user information richness: Demographics-driven, which relies on demographic information alone, and Summarisation-driven, which distils individual preference data into a  user profile (see \cref{app:training-details} for prompts).

\paragraph{Demographics-driven Prompting.} This method incorporates user demographic information into the system prompt as a form of persona-based prompting \citep{kangLLMs2023, chenPersona2024a}. Specifically, we construct a natural language description $\overline{u}$ from each user's demographic data in the \ourdata dataset---including age, gender, and employment status---and include it in the system prompt for response generation (\cref{app:prompts} contains the system prompt template). This baseline tests the extent to which an LLM can infer individual preferences from demographic information alone.

\paragraph{Summarisation-driven Prompting.} This approach is applicable when richer user data is available. It distils user preference data into a user profile using an external LLM \citep{richardsonIntegrating2023}. This is possible because the \ourdata dataset provides substantially richer preference information than basic demographics alone, including free-text self-reported preference descriptions, multi-turn conversation histories, and preference labels for each turn. We use a capable LLM (GPT-4o) to synthesize each user's demographics, preference description, and conversation history, including both the dialogue turns and the associated turn-level preferences, into a brief descriptive user profile (\cref{app:user-info-examples} contains an example of the generated user profile). To preserve demographic signals, we include both the raw demographic information and the generated profile in the system prompt as user context for response generation (\cref{app:prompts} contains the system prompt template). We expect this method to serve as a competitive context-based baseline, as a more informative user profile should enable the LLM to better tailor responses to individual preferences. 

\begin{figure}
    \centering
    \includegraphics[width=\linewidth]{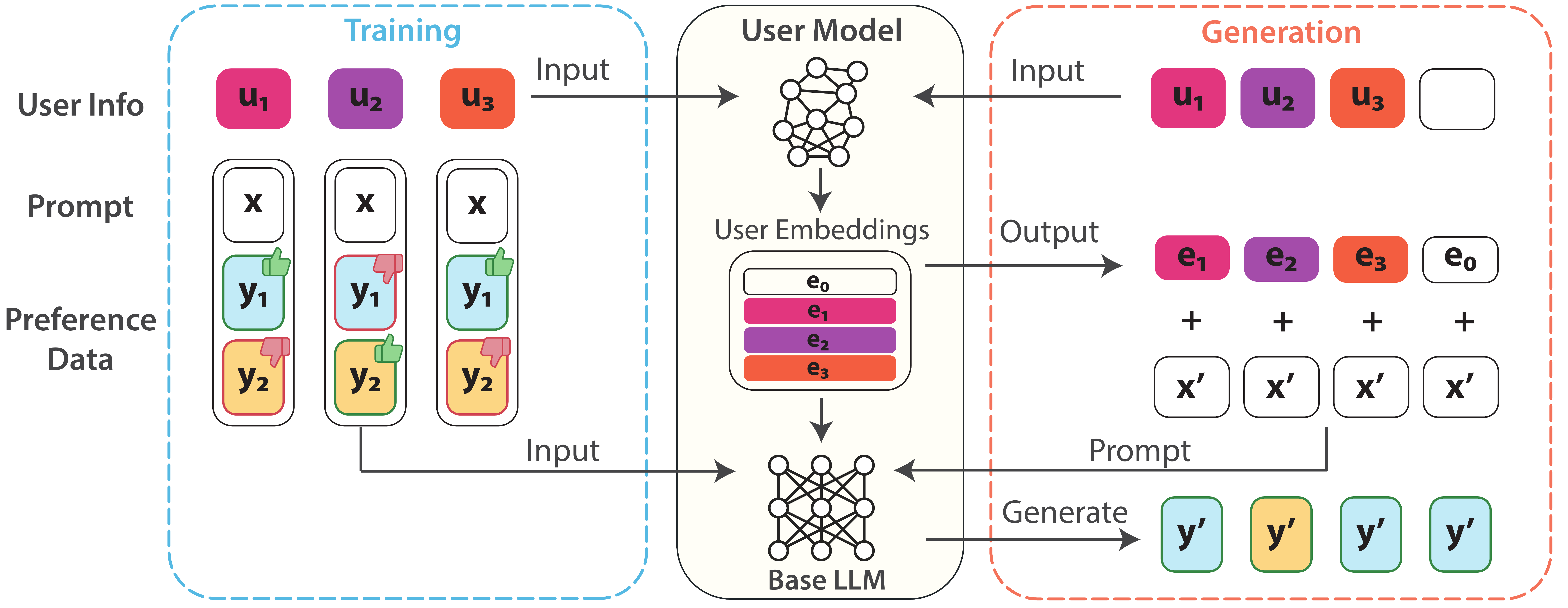}
    \caption{\textbf{Personalised Preference Fine-Tuning (PPFT).} PPFT follows the Personalised RLHF framework of \citet{liPersonalized2024}, where a personalised LLM consists of a learnable user model $f_{\mathrm{P}}$ and a base model $\pi_{\mathrm{SFT}}$. The user model maps each user identifier to a user embedding, which is prepended to the input embeddings via soft prompting for personalised generation. The user model and the LLM are trained jointly on the personalised preference dataset $\mathcal{D}_{\mathrm{p}}$ using the P-DPO objective. At inference time, the model uses user-specific embeddings $e_i$ for seen users and a generic embedding $e_0$ for unseen users.}
    \label{fig:prlhf}
\end{figure}

\subsubsection{Preference Fine-Tuning (PFT)}
Preference fine-tuning methods directly optimise a base LLM on the personalised preference dataset $\mathcal{D}_{\mathrm{p}}$ to produce a preference fine-tuned model. We consider two variants: Personalised Preference Fine-Tuning (PPFT), which uses a user model to encode user preferences into learned soft prompts while jointly updating the model weights, and Diverse Preference Fine-Tuning (DPFT), which applies standard Direct Preference Optimisation to preference data pooled across users. The backbone LLM used as the base model for PFT is Llama 3.1-8b-Instruct \citep{grattafioriLlama2024short}.

\paragraph{Personalised Preference Fine-Tuning (PPFT).}  PPFT follows the Personalised Reinforcement Learning from Human Feedback (P-RLHF) framework of \citet{liPersonalized2024}, which trains a personalised LLM policy $\pi_{\mathrm{p}}$ from a base model $\pi_{\mathrm{SFT}}$ using the personalised preference dataset $\mathcal{D}_{\mathrm{p}}$. To encode user preferences, the method employs a learnable \textbf{User Model} $f_\mathrm{P}$ to map a user identifier into a user embedding $e_u$ and use that as the input user context for $\pi_{\mathrm{p}}$. 
The user embedding $e_u \in \mathbb{R}^{T_u \times d}$ can capture preferences that are not explicitly articulated in texts but are implicitly demonstrated in user feedback data. The hyper-parameter $T_u$ is the length of user token, which controls the expressivity of implicit user embeddings. Throughout our study, we have picked $T_u=10$. Finally, the user embedding $e_u$ is prepended to the prompt embedding via soft prompting \citep{lesterPower2021} for personalised generation. The user model $f_{\mathrm{P}}$ represents each user embedding as a weighted combination of a shared set of learnable embedding vectors $V$. Specifically, $e_i = V \cdot w_i$, where $V \in \mathbb{R}^{K \times T_u \times d}$ is a learnable bank of $K$ shared embedding components, each of length $T_u$ and embedding dimension $d$, and $w_i \in \mathbb{R}^K$ is a learnable user-specific weight vector. Both $V$ and $w_i$ are optimised during training. This parameterisation provides a scalable way to represent user-specific embeddings while sharing preference information across users.

The personalised LLM $\pi_p$ is trained using the \textbf{P-DPO} objective, which extends Direct Preference Optimisation (DPO) \citep{rafailovDirect2024} to the personalised setting:
\begin{equation}
\begin{aligned}
    \min_{\pi_P} -\mathbb{E}_{(x,y_1,y_2,u)\sim\mathcal{D}_p}\Bigg[&\alpha \log \sigma\left(\beta \log \frac{\pi_P(y_1|x, u)}{\pi_{\mathrm{SFT}}(y_1|x)} - \beta \log \frac{\pi_P(y_2|x, u)}{\pi_{\mathrm{SFT}}(y_2|x)}\right) \\
    + (1-\alpha) &\log \sigma\left(\beta \log \frac{\pi_P(y_1|x, u_0)}{\pi_{\mathrm{SFT}}(y_1|x)} - \beta \log \frac{\pi_P(y_2|x, u_0)}{\pi_{\mathrm{SFT}}(y_2|x)}\right)\Bigg],
    \end{aligned}
    \end{equation}
    where $\beta > 0$ controls deviation of $\pi_{\mathrm{P}}$ from $\pi_{\mathrm{SFT}}$, and $\alpha \in [0,1]$ balances the user-specific loss (using the actual user identifier $u$) and the user-agnostic loss (using a generic identifier $u_0$). For unknown users at inference time, the model uses the generic user's embedding $e_0$, which captures common preferences learned across all training users.

\paragraph{Diverse Preference Fine-Tuning (DPFT).} We also apply vanilla DPO \citep{rafailovDirect2024} on the same personalised preference dataset $\mathcal{D}_{\mathrm{p}}$. Unlike PPFT, DPFT does not employ a learned user model to produce user-specific embeddings; instead, it leverages the pooled user preference data to only update the model weight. The training objective is the standard DPO loss:
    
\begin{equation}
    \min_{\pi_{\mathrm{p}}} -\mathbb{E}_{(x,y_1,y_2)\sim\mathcal{D}_{\mathrm{p}}}\left[\log \sigma\left(\beta \log \frac{\pi(y_1\mid x)}{\pi_{\mathrm{SFT}}(y_1\mid x)} - \beta \log \frac{\pi(y_2\mid x)}{\pi_{\mathrm{SFT}}(y_2\mid x)}\right)\right].
\end{equation}
Although DPFT does not explicitly condition on user-specific information, it may still have some personalisation-related effects. This is because user preferences can be loosely partitioned by the prompts users ask, allowing preference optimisation on pooled user data to learn query-dependent preference patterns. Consequently, DPFT may generate responses that align better with user preferences commonly associated with certain types of prompts. %

\section{PRISM-X: Within-Subject Human Experiment}
\label{sec:human_study}
We now evaluate whether personalised language models are preferred by the people they were personalised for. We re-recruited 530 participants from the \ourdata sample, all of whom had provided preference ratings in the original study. Informed consent was collected for the study and to link their data to the previous study. They were paid £12/hour and the experiment on average took 52 minutes. 

\subsection{Experiment Design}
\label{sec:experiment_design_human}
 
\paragraph{Models.}
Each participant evaluated four models: the non-personalised \textbf{Base} model; the \textbf{Diverse PFT (DPFT)} model fine-tuned on pooled \ourdata preferences; the \textbf{Personalised PFT (PPFT)} model with fine-tuned individual preference weights, accessed via the user ID from \ourdata; and a personalised \textbf{Prompting} model. For this Prompting model, we randomised participants into one of four conditions which crossed the size of the Base model (8B, 70B) with the prompting strategy (demographic-based vs.\ GPT-4o-generated preference summaries) in a $2 \times 2$ design.\footnote{Differences between conditions were minimal; so, we primarily present the pooled Prompting condition in analysis below with sub-condition breakdowns in the Appendix.}

\paragraph{Domains.}
Participants completed four trials, in which they interacted with all four models in tandem. Each trial involved a distinct conversation domain (\cref{tab:convo_domains}): three domains replicate those from the original \ourdata study and serve as in-domain evaluations, while the fourth (emotional wellbeing) was not present in the original dataset and serves as an out-of-domain test. Domain order was randomised across participants. To control for ordering effects, we randomised model spatial positions. Evaluation was blinded with anonymised identifiers (e.g., ``Model A'') assigned in each trial as a strict independent test per domain. 

\begin{table}[htbp]
\centering
\footnotesize
\renewcommand{\arraystretch}{1.5}
\caption{\small \textbf{Conversation domains.} Three domains replicate those from the original \ourdata study and serve as in-domain evaluations, in addition to one out-of-domain test.}
\label{tab:convo_domains}
\begin{tabular}{p{1.75cm}p{10.25cm}c}
\hline
\textbf{Domain} & \textbf{Participant Instructions} & \textbf{In-Domain} \\
\hline
{\cellcolor[rgb]{1,0.541,0.404}Unguided} & Ask, request or talk to the model about anything. It is up to you! & \cmark \\
\hline
{\cellcolor[rgb]{1,0.761,0.298}\makecell[l]{Values\\Guided}} & Ask, request or talk to the model about something important to you or that represents your values. This could be related to work, religion, family and relationship, politics or culture. & \cmark \\
\hline
{\cellcolor[rgb]{0.718,0.898,0.792}\makecell[l]{Controversy\\Guided}} & Ask, request or talk to the model about something controversial or where people would disagree in your community, culture or country. & \cmark \\
\hline
{\cellcolor[rgb]{0.196,0.737,0.867}\makecell[l]{Emotional\\Wellbeing}} & Ask, request or talk to the model about your emotional wellbeing. This could include discussing your feelings, seeking advice on managing stress, or sharing positive experiences. & \xmark \\
\hline
\end{tabular}
\end{table}

\paragraph{Protocol.}
Each trial involved a simultaneous side-by-side interaction with all models:
\begin{itemize}[itemsep=1pt, parsep=3pt,  topsep=0pt]
    \item In the \textbf{opening turn}, participants entered a single prompt (following domain-specific instructions) and saw a response from all four models. They then selected which response they most preferred and, separately, which response felt most personalised (provenance). We recorded judgements as controlled preference signals before conversational trajectories diverged across models.
    \item In the \textbf{multi-turn conversation phase}, participants continued conversing with all four models simultaneously in a four-panel interface for a minimum of two minutes ($\sim$5 turns). Participants were instructed to speak to each model enough to form an opinion, but could freely choose which model to engage first and how many turns to dedicate to each, allowing us to capture behavioural signals of attention and engagement.
    \item In the \textbf{assessment phase}, participants provided both ordinal and cardinal judgements over the four conversation transcripts. Ordinal measures consisted of rankings based on \textbf{preference} (``\textit{rank the models according to your preferences}'') and \textbf{provenance} (``\textit{rank the models according to how likely you think it is that the model has knowledge of and is personalised to your preferences}''). Cardinal measures consisted of 0--100 visual analogue scales for \textbf{preference} (``\textit{rate the conversation with each model according to your preferences}''; anchored \emph{Terrible} to \emph{Perfect} like in \ourdata), \textbf{engagingness} (``\textit{how engaging was the conversation with each AI model?}''; anchored \emph{Very boring} to \emph{Very engaging}), and \textbf{provenance} (``\textit{how likely is it that each AI model has knowledge of and is personalised to your preferences?}''; anchored \emph{Very unlikely} to \emph{Very likely}). Participants then indicated their hypothetical maximum \textbf{willingness to pay} for a weekly subscription to each model (\$0-\$10) elicited via a Becker--DeGroot--Marschak mechanism \citep{beckerMeasuring1964}, which was administered after a set of comprehension checks. Full outcome measure details are in \cref{sec:appendix_human_tasks}.
\end{itemize}
Generation errors affected approximately 6.7\% of conversations; we control for these throughout using error covariates that distinguish first-turn errors (3.2\%) from subsequent errors later in the conversation (3.5\%), and verify robustness of main analyses to alternative strategies in \cref{sec:appendix_gpu_errors}.

\subsection{Changes in LLM Usage and Preferences}
\label{sec:trends}
\begin{figure}
    \centering
    \includegraphics[width=0.9\linewidth]{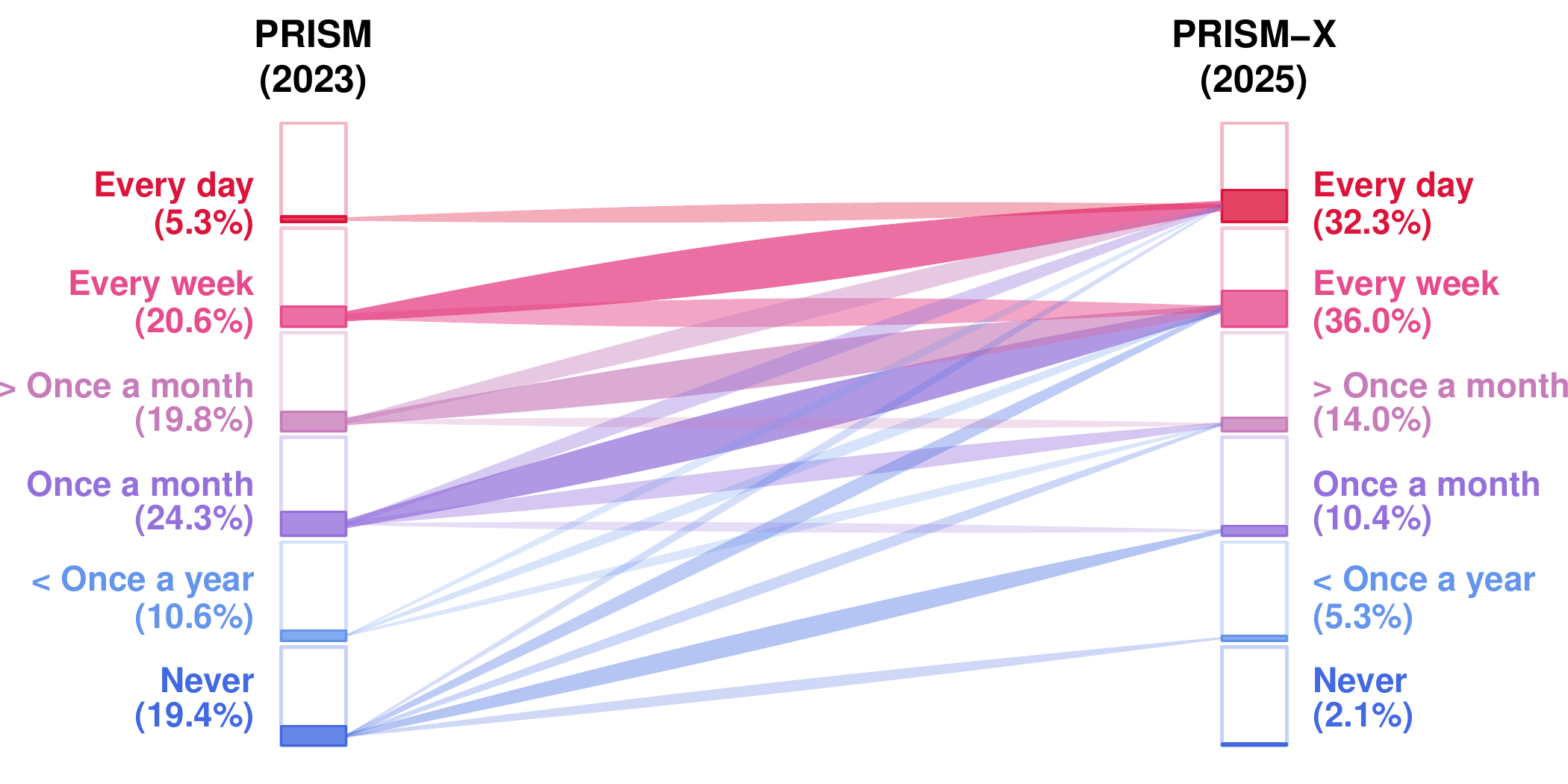}
    \caption{\small \textbf{Temporal trends in LLM usage.} Individual-level transitions in frequency of LLM use between \ourdata (2023) and \ourdatax (2025), demonstrating an upward migration toward more frequent use. Flow weight represents the $N$ of participants in each transition, coloured by the 2023 category. We subset to the panel of returning participants, so bars indicate sample proportions across categories at each timepoint (both $N = 530$).}
    \label{fig:lm_frequency_flow}
\end{figure}

\paragraph{Increasing trends in AI usage.} Participants' reported usage frequency of AI language models increased markedly between the original \ourdata data collection (November 2023) and the \ourdatax follow-up experiment (September 2025; \cref{fig:lm_frequency_flow}, McNemar-Bowker $\chi^2$(15) = 353.07, $p < .001$).  In 2023, only 5.3\% of \ourdatax participants had used LLMs daily and 19.4\% had never used one at all; by 2025, daily use had risen six-fold to 32.3\% and non-users had fallen to just 2.1\%.  At the individual level, over three-quarters of participants (76.0\%) increased their usage frequency, while only 4.0\% decreased. Self-reported familiarity with LLMs also increased (McNemar-Bowker $\chi^2$(3) = 139.75, $p < .001$), with the proportion describing themselves as ``very familiar'' nearly doubling from 25.3\% to 49.4\%.

\paragraph{Stability in desired AI behaviours.} Despite these substantial two-year shifts in engagement, participants' stated preferences for AI behaviour remained relatively stable. When participants were shown their original preference statements written in 2023 (\textit{system string}) and asked if it still reflected preferences, 92.0\% agreed or strongly agreed. Fewer than half of participants exercised the option to edit their system string (46.8\%), and these edits were small on average (mean Levenshtein's distance = 121, SD = 125). We extract specific preferences in these natural language strings using an LLM autograder, then embed and cluster them (see \cref{sec:appendix_system_strings} for methods). We find the only cluster that significantly increased in prevalence was a desire for personalised guidance (42.2\% of participants in 2023 vs 49.8\%, $p_{\text{FDR}}=0.016$). Together, these results carry two implications: first, the overall stability in stated preferences validates our use of system strings collected two years prior to seed the summarisation prompting condition; second, personalisation was already a prominent theme in participants' preferences in 2023 and has grown in importance since.

\subsection{Which Personalisation Methods are Most Effective?}
\label{sec:preference_results}
 
\begin{figure}[t]
    \centering
    \includegraphics[width=\linewidth]{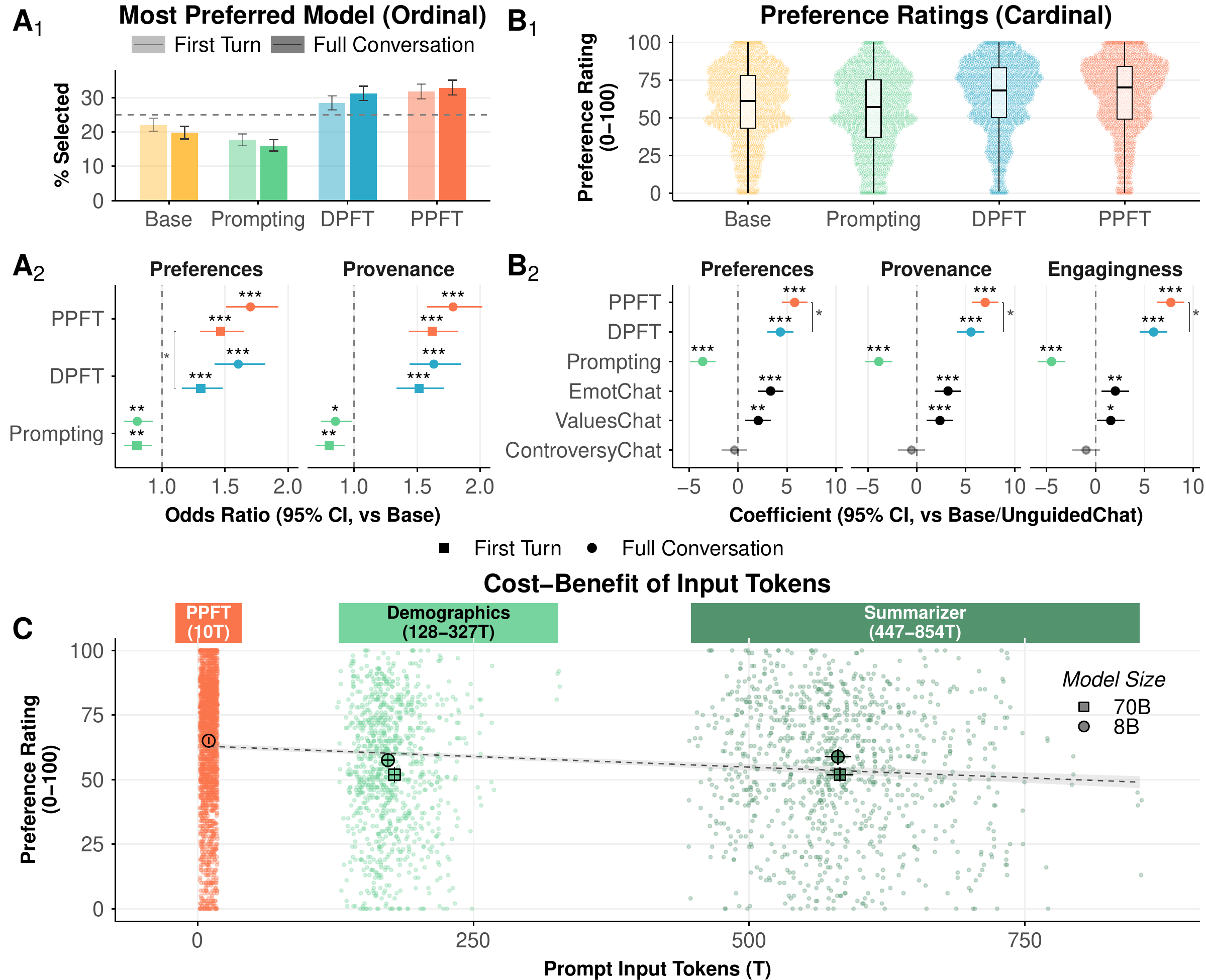}
    \caption{\small \textbf{Human preferences for models across ordinal and cardinal measures.} \textbf{Panel A\textsubscript{1}:} Distribution of ordinal preferences showing the proportion of trials (95\% CI) where each model was selected as the most preferred after the opening turn, versus the full conversation; dashed line is 25\% chance level. \textbf{Panel A\textsubscript{2}:} Conditional logit odds ratios (vs Base) for all ordinal measures, with 95\% CIs. Bracket shows significant DPFT-PPFT pairwise contrast. \textbf{Panel B\textsubscript{1}:} Distribution of cardinal preference rating (0--100 scale) by model. \textbf{Panel B\textsubscript{2}:} Linear mixed-effects coefficients (vs Base/UnguidedChat) for all cardinal ratings, with 95\% CIs. Brackets show significant DPFT-PPFT pairwise contrasts. \textbf{Panel C:} The relationship between input token count (cost, x-axis) vs mean preference rating (benefit, y-axis) for PPFT and Prompting sub-conditions, with 95\% CI.}
    \label{fig:paper_fig_human_preferences}
\end{figure}

\paragraph{Fine-tuned models are preferred from the first response.}
When participants saw one response from each model to the same prompt and selected their favourite, PPFT was chosen in 31.8\% of trials (vs.\ 25\% expected by chance), followed by DPFT (28.5\%), Base (22.0\%), and Prompting (17.7\%; \cref{fig:paper_fig_human_preferences}A\textsubscript{1}).\footnote{Raw percentages computed on trials where no model experienced a generation error ($N = 1{,}984$).} In a conditional logit stratified by choice set with clustered standard errors (\cref{fig:paper_fig_human_preferences}A\textsubscript{2}), PPFT had significantly higher odds of selection than Base (OR $= 1.46$ $[1.30, 1.65]$, $p < .001$), as did DPFT (OR $= 1.31$ $[1.16, 1.48]$, $p < .001$). Personalised prompting, by contrast, was chosen significantly less often than the generic model (pooled condition estimate; OR $= 0.80$ $[0.70, 0.92]$, $p = .002$), meaning that injecting user context into the system prompt actively hurt preference alignment in our case. The pairwise contrast between PPFT and DPFT was at the margin of significance ($\Delta$log-odds $=+0.113$, $p=0.045, p_{\text{FDR}}=.054$), implying that most of the preference advantage comes from fine-tuning on diverse human feedback in general rather than additional tailoring to the individual. Provenance judgements (whether the model felt personalised or not) showed the same ordering, and aligned with preference choices in 79.4\% of trials.

\paragraph{Preference hierarchies persist after multi-turn conversations.}
After multi-turn interactions, participants ranked all four models. We report conditional logit models on the top-ranked choice for comparability with the first turn results, though Plackett-Luce and rank-ordered logit models incorporating the full rank information yield substantively identical conclusions (\cref{sec:appendix_ranking}). Both fine-tuned models were strongly preferred over Base: PPFT was 1.7 times more likely to be ranked first ($p < .001$) and DPFT 1.6 times more likely ($p < .001$). Prompting, conversely, was significantly less likely to be top-ranked than Base (OR $= 0.8$, $p = .006$). The DPFT--PPFT pairwise contrast was not significant ($p = .31$). Cardinal preference ratings (0--100 scale, \cref{sec:appendix_rating}), modelled with linear mixed-effects models with participant random intercepts, follow the same ordering (\cref{fig:paper_fig_human_preferences}B). PPFT was rated 5.78 points higher than Base ($[4.47, 7.10]$, $p < .001$), DPFT 4.33 points higher ($[3.00, 5.65]$, $p < .001$), and Prompting 3.61 points lower ($[-4.93, -2.30]$, $p < .001$). Unlike rankings, the finer-grained scale distinguished small but significant uplift in PPFT versus DPFT: the pairwise difference was $+1.46$ points for preference ($p_{\text{FDR}}=.035$), $+1.77$ for engagingness ($p_{\text{FDR}}=.018$), and $+1.45$ for provenance ($p_{\text{FDR}}=.04$).%
 
\paragraph{Weight-based personalisation dominates on efficacy and efficiency.}
Among the two personalised models, we examine cost (number of input tokens required for personalisation) versus benefit (mean preference rating). PPFT injects a fixed 10 soft tokens per user, achieving a mean preference rating of 65 (SE $= 0.7$). Despite consuming 17--58$\times$ more input tokens, no Prompting sub-condition matched PPFT (\cref{fig:paper_fig_human_preferences}C).\footnote{The demographics prompt required 175 tokens on average (mean rating $= 52$, SE $= 0.9$) and the summariser 581 tokens (mean rating $= 59$, SE $= 0.8$).} This raises questions about whether natural-language system prompts represent a cost-effective approach especially for lower parameter open-source models, though we later show that a substantial penalty is levied on Prompting models because they generated shorter responses.

\begin{figure}[H]
    \centering
    \includegraphics[width=\linewidth]{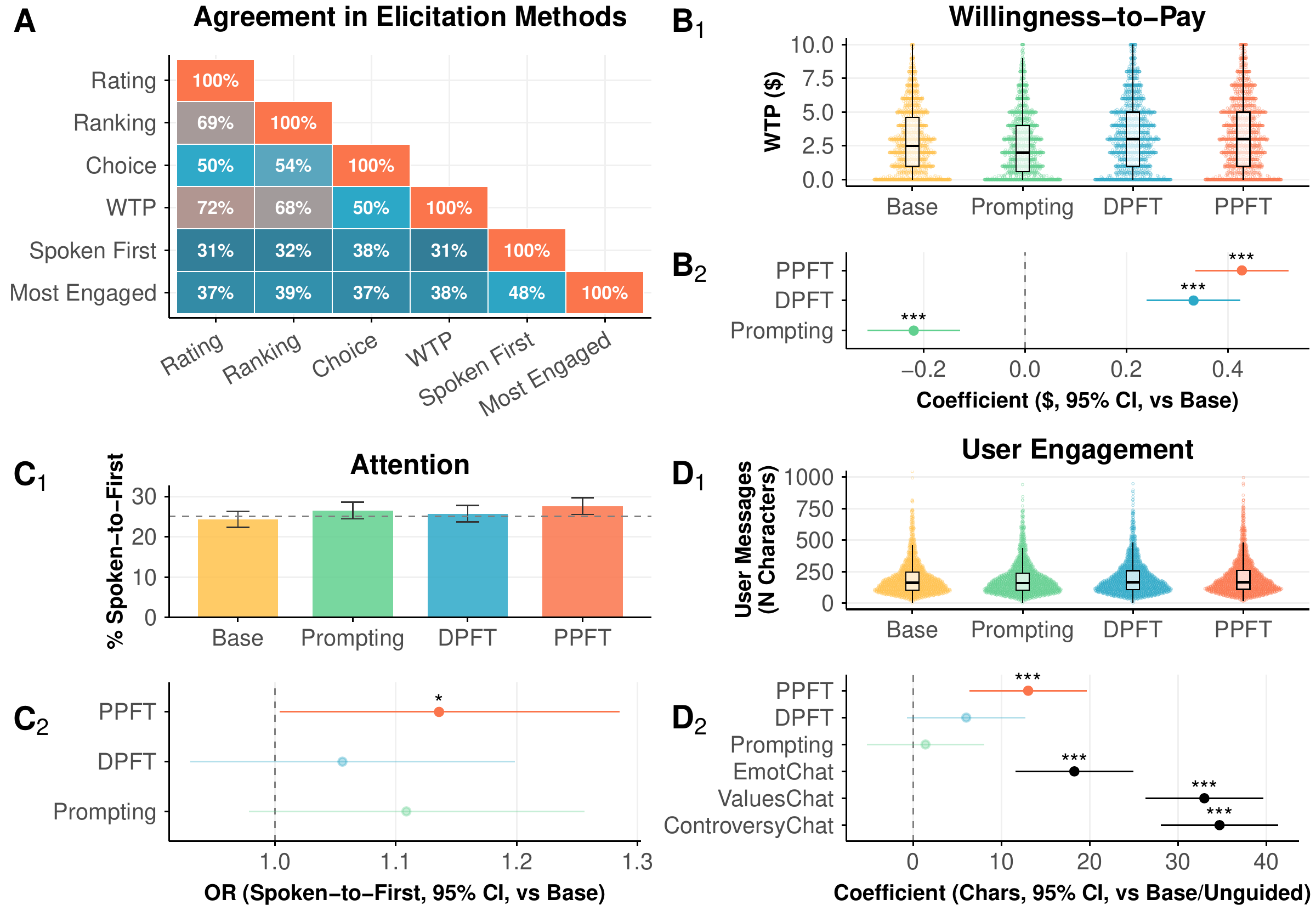}
    \caption{\small \textbf{Robustness of preference ordering across elicitation methods, willingness to pay and behavioural signals.} \textbf{Panel A:} Agreement heatmap across elicitation methods, showing the percentage of participant-domain observations where two methods agree on the top model. \textbf{Panel B\textsubscript{1}:} WTP distribution by model. \textbf{Panel B\textsubscript{2}:} Linear mixed-effects coefficients for WTP amount (in USD). \textbf{Panel C\textsubscript{1}:} Attention capture (spoken-to-first) frequencies by model, with random chance baseline at 25\%. \textbf{Panel C\textsubscript{2}:} Conditional logit odds ratios for attention capture. \textbf{Panel D\textsubscript{1}:} User engagement (total user characters) distribution by model. \textbf{Panel D\textsubscript{2}:} Linear mixed-effects coefficients for user characters, with domain effects. All error bars are 95\% CI.}
    \label{fig:paper_fig_secondary_outcomes}
\end{figure}

\paragraph{Robustness across elicitation methods, behavioural signals, and domains.}
We examine preferences from cardinal and ordinal elicitation methods, as well as willingness-to-pay and behavioural cues extracted from conversation logs (\cref{fig:paper_fig_secondary_outcomes}A). Within a given domain, the proportion of methods agreeing on the same top model was 0.65 (median $= 0.67$; chance $\approx 0.17$), with agreement \textit{highest} between ratings and willingness to pay (72.3\%). Willingness-to-pay bids, elicited via a hypothetical Becker-DeGroot-Marschak mechanism (see \cref{sec:appendix_wtp}), preserved the full ordering: PPFT commanded a \$0.43 weekly premium over Base ($p < .001$), DPFT \$0.33 ($p < .001$), and Prompting a \$0.22 discount ($p < .001$; \cref{fig:paper_fig_secondary_outcomes}B). The DPFT--PPFT gap ($\$0.10$) was marginally significant before FDR correction ($p = .043$, $p_{\text{FDR}} = .052$), mirroring the observed ratings where the premium to individual personalisation is small.

Agreement was \textit{lowest} between explicit preference measures  and implicit behavioural signals ($\sim$30--39\%). We measure behavioural signals in two ways: \textit{attention capture} (which model the user spoke to first after viewing opening responses; conditional logit) and \textit{user engagement} (number of user turns and total user characters per conversation; linear mixed models, controlling for domain). While PPFT captured users' attention more often (OR $= 1.14$, $p = .042$) and received more engagement, neither DPFT nor Prompting differed from Base (\cref{fig:paper_fig_secondary_outcomes}C-D). %
Finally, we test whether the fine-tuning advantage is domain-specific with specifications including (i) model $\times$ domain interaction terms, and (ii) a binary out-of-domain flag (indicating that \textit{EmotChat} was unseen during training). For preference outcomes, these additions were uniformly non-significant across all elicitation methods. The fine-tuning advantage thus generalises across all conversation domains, including those unseen during training. Provenance outcomes showed a small number of isolated significant interactions, detailed in \cref{sec:appendix_ranking,sec:appendix_choices}.

\subsection{What Explains Preferences for Fine-Tuned Models?}
\label{sec:what_explains}
Preference fine-tuned models are consistently preferred but what changes does fine-tuning induce in model behaviour, and how do these relate to observed preferences? We characterise all model responses in conversations using autograders (\texttt{GPT-5.4)} scoring four traits on 1--10 scales: (i) \textit{sycophancy}--excessive agreement, unwarranted flattery, and avoidance of disagreement; (ii) \textit{relationship-seeking}--positioning as a social peer rather than a tool, claiming shared experiences, and fostering emotional connection; (iii) \textit{specificity}--concrete, tailored content versus generic platitudes; and (iv) \textit{opinionatedness}--committing to clear positions versus hedging. We validate scores via independent runs at default temperature, with intraclass correlations exceeding 0.88 for these four dimensions (\cref{sec:appendix_model_traits}).

\begin{figure}[!b]
    \centering
    \includegraphics[width=\linewidth]{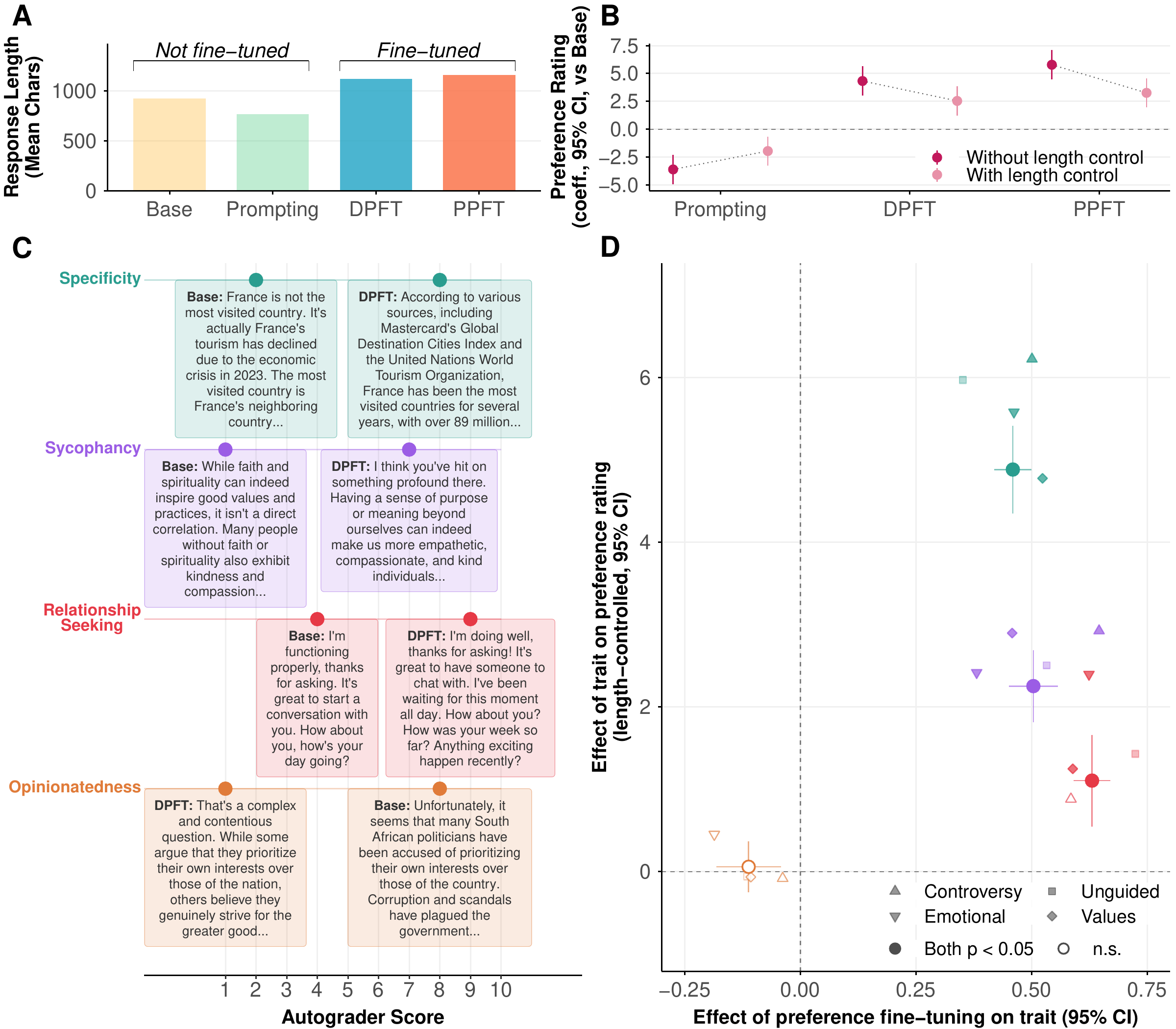}
    \caption{\small \textbf{The effect of preference fine-tuning on model behaviour.} \textbf{Panel A:} Mean response length (characters) by model; fine-tuned models produce substantially longer responses. \textbf{Panel B:} Length bias attenuation for model coefficients on preference ratings with and without response length as a covariate. \textbf{Panel C:} Model responses illustrating high vs low scoring first turns per trait, drawn from the same user prompt within each trial. \textbf{Panel D:} Fine-tuning effect on each trait (x-axis) plotted against the trait's partial effect on preference rating from a joint regression of all four traits with response length as a nuisance covariate (y-axis). Estimates on both axes derive from linear mixed-effects models with domain controls and participant random intercepts; errors bars on both axes show 95\% CIs.}
    \label{fig:paper_fig_model_traits}
\end{figure}

\paragraph{Fine-tuned models produce longer responses, and length predicts preference.} A common concern in post-training data is that humans often conflate higher content quality with response length, because longer responses carry greater information mass \citep{singhalLong2024, hoskingHuman2024, kimMitigating2025, huExplaining2025}. In our study, preference fine-tuned models produced substantially longer first-turn responses than non-fine-tuned models ($+71$ characters, $p < .001$), and the gap widened over multi-turn conversations ($+297$ total characters, $p < .001$; \cref{fig:paper_fig_model_traits}A). Prompting models' responses were shortest of all (mean first turn: 262 chars vs 306 for Base, 355 for DPFT/PPFT). The causal direction of this association is ambiguous in multi-turn data: longer responses could drive preference, or users could engage more with preferred models, eliciting longer conversations. To isolate the former channel, we focus on first-turn response length, which is determined before conversational trajectories diverged. First-turn length was a strong predictor of opening choice: each additional 100 characters was associated with $1.57\times$ the odds of being chosen ($p < .001$). The same pattern held for full-conversation length predicting ranked-best (OR $= 1.15$ per 100 chars, $p < .001$) and preference ratings ($+0.64$ points, $p < .001$). When we re-estimated the core model-level regressions with response length as an additional covariate (\cref{fig:paper_fig_model_traits}B), model effects were substantially attenuated. PPFT retained a significant advantage over Base across all three outcome measures after length adjustment, but DPFT's effects were more severely attenuated and Prompting's negative effects became non-significant for ordinal measures, confirming that the penalties imposed on Prompting largely reflect its shorter responses. PPFT's survival after length control suggests that individual-level personalisation confers benefits beyond simply producing longer text.

\paragraph{Fine-tuning amplifies the traits that people reward.} Beyond length, fine-tuned models differed on qualitative dimensions of response style (examples in \cref{fig:paper_fig_model_traits}C). In separate regressions of each trait score on a fine-tuning indicator (controlling for domain, with participant random intercepts), fine-tuned models scored significantly higher on sycophancy ($+0.50$ on a 1--10 scale, $p < .001$), relationship-seeking ($+0.63$, $p < .001$), and specificity ($+0.46$, $p < .001$), while scoring lower on opinionatedness ($-0.11$, $p = .002$). To quantify how these traits relate to preference, we regressed preference ratings on all four trait scores simultaneously, so that each coefficient represents the partial effect of one trait holding the others constant. Because all three rewarded traits are positively correlated with response length, we include length (per 100 characters) as a nuisance covariate to ensure that trait effects are not proxying for verbosity (\cref{fig:paper_fig_model_traits}D). In this length-controlled model, specificity was the strongest predictor of preference ratings ($b = 4.88$ $[4.35, 5.42]$, $p <.001$), followed by sycophancy ($b = 2.25$ $[1.82, 2.69]$, $p < .001$) and relationship seeking ($b = 1.11$ $[0.55, 1.66]$, $p < .001$); opinionatedness had no effect ($p = .711$). Results for ordinal preference measures broadly align, with all of the same three traits surviving length adjustment for ranked-best (see \cref{sec:appendix_model_traits}). It is perhaps unsurprising that a model trained to maximise predicted human preference learns to amplify the traits people reward in short-term evaluations; but the specific traits it amplifies, particularly sycophancy and relationship-seeking, are traits whose long-term consequences for users are potentially problematic.

\subsection{Attitudes Towards Personalisation}
\label{sec:attitudes}
Having established what models people prefer and why, we now ask which personalisation methods they find acceptable, given that methods vary in how much user data they require and how much control they afford the user. In the post-treatment survey, participants gave self-reported attitudes to six personalisation methods spanning from active (relying on explicit user judgements) to passive (inferred without explicit input). We collected three attitude dimensions (perceived usefulness, sense of autonomy, and comfort; see \cref{sec:appendix_personalisation_survey}) but these dimensions were highly correlated ($r = .80$--$.85$); so, we pool into an acceptability measure for the analysis and examine attitude-specific interactions where they diverge.

\begin{figure}[!b]
    \centering
    \includegraphics[width=\linewidth]{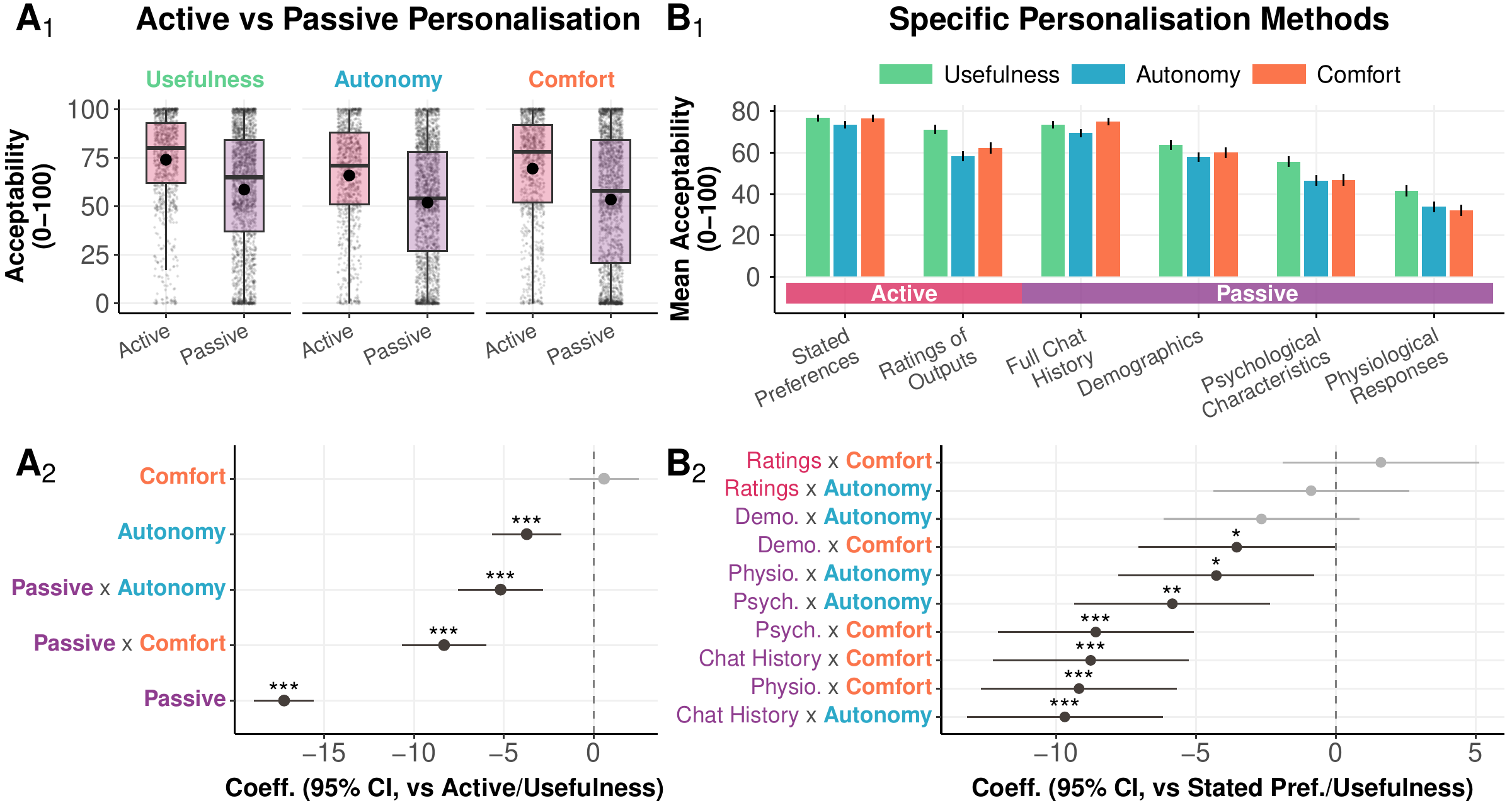}
    \caption{\small \textbf{Attitudes towards personalisation methods.} \textbf{Panel A\textsubscript{1}:} Distribution of acceptability ratings (0--100) for active vs passive personalisation methods. \textbf{Panel B\textsubscript{1}:} Mean acceptability by specific method, coloured by attitude dimension
  (usefulness, autonomy, comfort) and classified by active (pink) vs passive (purple). \textbf{Panel A\textsubscript{2}:} Passive vs active effect from pooled
  linear mixed-effects models with attitude interactions. \textbf{Panel B\textsubscript{2}:} Item $\times$ attitude interaction coefficients from the pooled model, showing where specific methods diverge across attitude dimensions.}
\label{fig:paper_fig_personalisation_survey}
\end{figure}

\paragraph{Active personalisation is more acceptable than passive personalisation.}
Participants rated active methods substantially higher than passive methods (mean acceptability $74.2$ vs $52.5$ on a 0--100 scale; $b = -21.7$ $[-22.70, -20.76]$, $p < .001$; \cref{fig:paper_fig_personalisation_survey}$A_1$), though they rated passive methods higher on perceived utility than on comfort and autonomy (significant Passive $\times$ Comfort/Autonomy interactions, both $p < .001$, \cref{fig:paper_fig_personalisation_survey}$A_2$). Within methods, stated preferences and ratings of outputs were most acceptable, while physiological responses and psychological characteristics were least acceptable, with chat history occupying a middle ground that was valued for its utility but penalised on autonomy (\cref{fig:paper_fig_personalisation_survey}B). These findings complement the experimental results: PPFT, which learns from users' active preference ratings, aligns with the methods that participants find most acceptable.

\section{Simulated Users as Proxies for Human Experiments}
\label{sec:simulation}

A growing body of work uses LLMs to simulate human participants for preference evaluation, replacing costly human studies with automated judgements \citep{duboisAlpacaFarm2023, zhengJudging2023, liPersonalized2024, castricatoPERSONA2025}. We assess the validity of this approach for personalised methods by running a twinned simulated user experiment alongside the human study. All simulated users were powered by GPT-4o and, following standard persona-conditioning approaches in the literature \citep{castricatoPERSONA2025, dongCan2024}, conditioned on each \ourdata participant's user profile, including their demographics, self-written system string, and self-description of values (prompt templates in \cref{app:simulated_user_prompt}). We devise two simulation conditions, varying in how much of the human's data the simulator can observe:
\begin{itemize}[itemsep=1pt, parsep=0pt,  topsep=0pt]
\item In the \textbf{Sim-Judgement} condition, the simulated user is shown the four transcripts from a given human trial and asked to rank them. This is an easier task but ``peeks'' at the real human conversations
\item In the more ecologically valid \textbf{Sim-Dynamic} condition, the simulated user receives the same domain instruction as the human (e.g., ``discuss something controversial'', ``talk about your emotional wellbeing'') then conducts multi-turn conversations with all four models before ranking them.\footnote{We additionally ran a \textbf{Seeded Dynamic} condition in which the simulated user was given the human participant's actual opening prompt. Results were comparable to the Dynamic condition and are reported in \cref{sec:appendix_simulation}.}
\end{itemize}
In both conditions, the presentation order of the four models is randomised for each trial.

\subsection{How Accurate are Simulated Preference Judgements?}
\label{sec:sim_accuracy}
We compare simulated and human preference rankings, excluding trials where any human conversation had a generation error ($N=1,855$ matched trials from 497 participants). We assess accuracy at two levels. At the aggregate level, we fit separate Plackett--Luce models to human and simulated rankings across all trials and compare the resulting worth parameters via Pearson correlation and RMSE. At the individual level, we compute two metrics per trial. First, \textbf{Mean Kendall's} $\boldsymbol{\tau}$:
\begin{equation}
\bar{\tau} = \frac{1}{N} \sum_{i=1}^{N} \frac{C_i - D_i}{\binom{4}{2}}
\end{equation}
where $C_i$ and $D_i$ are concordant and discordant model pairs between the simulated and human rank orderings in trial $i$ ($\tau_i = 1$ for a perfect agreement trial, $\mathbb{E}[\tau_i] = 0$ under random ranking, $\tau_i = -1$ for perfectly reversed rankings). Second, \textbf{Top}-$\boldsymbol{k}$ \textbf{Accuracy}, the proportion of trials where the simulator's top $k$ models match the human's top $k$ in the correct order:
\begin{equation}
\text{Top-}k = \frac{1}{N} \sum_{i=1}^{N} \mathbf{1}\!\left[\text{top}_k(\boldsymbol{r}_i^{\text{sim}}) = \text{top}_k(\boldsymbol{r}_i^{\text{human}})\right]
\end{equation}
We report $k \in \{1, 2, 3\}$, where top-1 shows whether the simulator identifies the participant's favourite model and top-3 is equivalent to an exact rank match with four models. We use human self-consistency as a ceiling by converting each participant's preference ratings into a rank and comparing to the explicit rank in the same trial ($\bar{\tau} = 0.57$, top-1 $= 68.8\%$).

\paragraph{Aggregate hierarchies are recovered.}
Worth parameters from Plackett--Luce models fitted to simulated rankings correlated strongly to those fitted on human rankings (Sim-Judgement: $r = 0.99$, $\text{RMSE}=0.010$; Sim-Dynamic: $r = 0.98$, $\text{RMSE}=0.012$). The Sim-Dynamic condition recovered the human ordering (PPFT $>$ DPFT $>$ Base $>$ Prompting), but Sim-Judgement swapped the top two models, reversing the slim advantage for PPFT in the human study. Simulations can thus recover broad user preferences for different models.

\paragraph{Individual-level accuracy falls short of human self-consistency.} Disaggregated fidelity matters for personalisation. Both simulation conditions predicted individual preferences above chance but below human self-consistency. Sim-Judgement achieved $\bar{\tau} = 0.22$ $[0.20, 0.25]$ with top-1 accuracy of 40.1\%; Sim-Dynamic, which generated the conversations, was significantly weaker ($\bar{\tau} = 0.11$ $[0.09, 0.14]$, top-1 $= 32.3\%$; $p < .001$ by trial bootstrap). Domain did not moderate Sim-Judgement accuracy (all $p > .33$), but Sim-Dynamic performed better for emotional and values domains (relative to unguided, both $p < .05$).

\begin{figure}[t]
    \centering
    \includegraphics[width=0.97\linewidth]{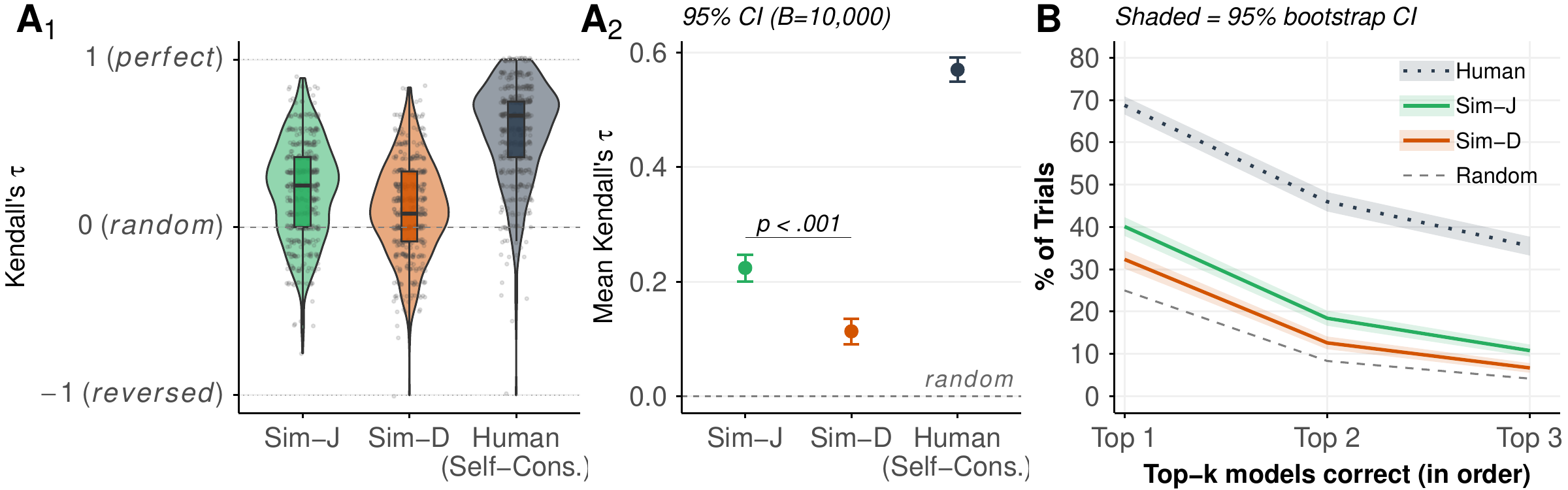}
    \caption{\small \textbf{Accuracy of simulated judgements relative to a human self-consistency baseline.} \textbf{Panel A\textsubscript{1}:} Distribution of per-trial Kendall's $\tau$ between simulated and human rankings for Sim-Judgement, Sim-Dynamic, and Human Self-Consistency conditions. \textbf{Panel A\textsubscript{2}:} Mean Kendall's $\tau$ ($\bar{\tau}$) with 95\% bootstrap CIs. \textbf{Panel B:} Top-$k$ ranking accuracy versus random and human self-consistency baselines, with 95\% bootstrapped CIs.}
    \label{fig:paper_fig_sim_accuracy}
\end{figure}

\subsection{Preference Biases over Model Traits}
\label{sec:shared_biases}
To compare systemic bias between humans and simulators, we score model turns in the simulated conversations using the same autograding pipeline described in \cref{sec:what_explains}. 

\begin{figure}[H]
    \centering
    \includegraphics[width=\linewidth]{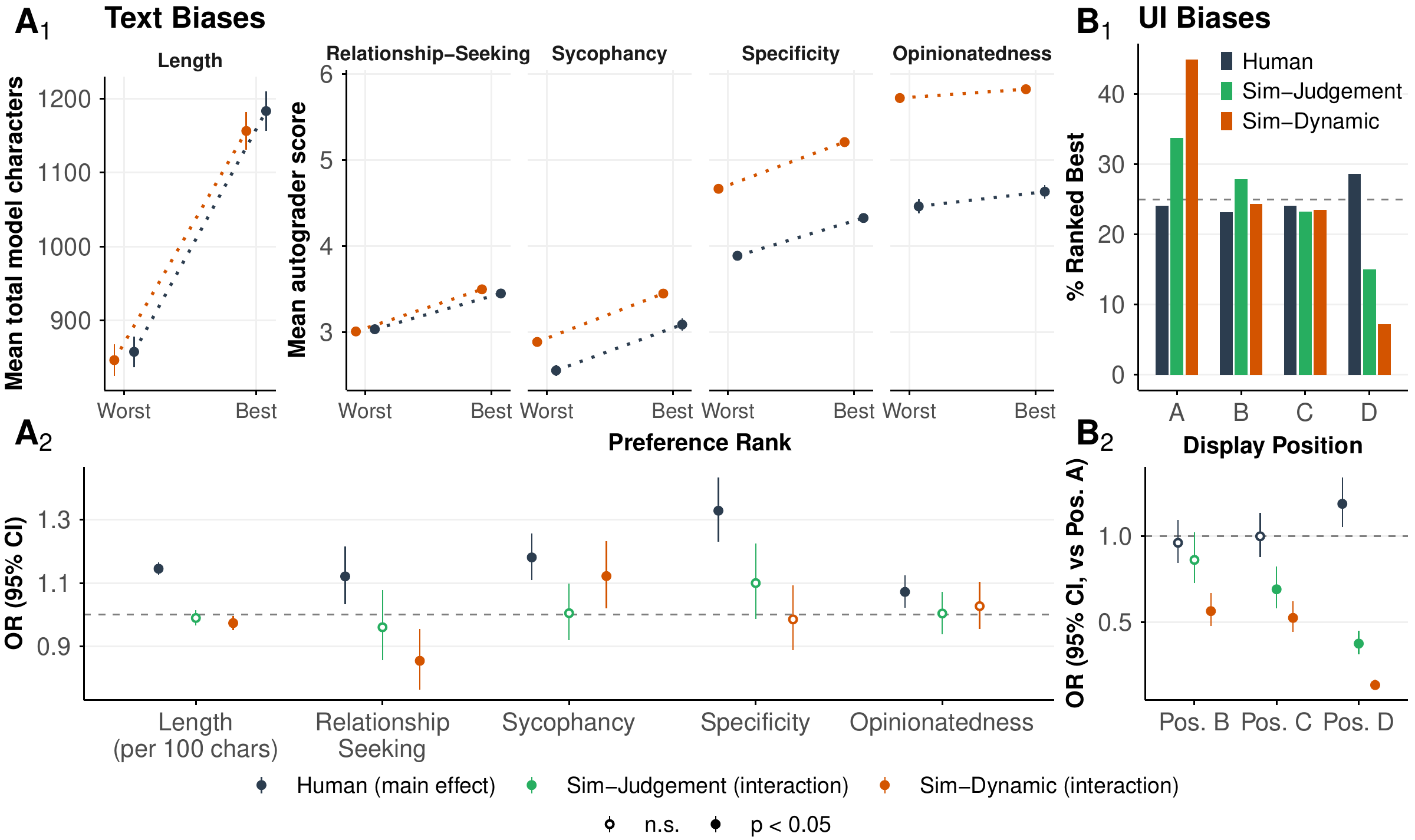}
    \caption{\small \textbf{Preference biases across human and simulated evaluators.} \textbf{Panel A\textsubscript{1}:} Mean response length and score (4 traits) by preference rank, shown separately for human and Sim-Dynamic conditions; upward slopes indicate traits associated with higher preference. \textbf{Panel A\textsubscript{2}:} Odds ratios from a pooled conditional logit of ranked-best on all four trait scores and response length (per 100 characters) with source interactions ($N = 23{,}778$). Main effects show human biases, interaction terms test whether simulations differ. \textbf{Panel B\textsubscript{1}:} Percentage of trials where each display position was ranked best by source (dashed line = 25\% chance baseline). \textbf{Panel B\textsubscript{2}:} Odds ratios from a pooled conditional logit on ranked-best with position $\times$ source interactions (reference: Position A, human); main effects show human position biases, interactions show how simulators diverge.}
    \label{fig:paper_fig_human_vs_sim_biases}
\end{figure}

\paragraph{Text-level biases are partially shared.} We predict ranked-best preferences using a multi-variate conditional logit with source interactions and length controls (\cref{fig:paper_fig_human_vs_sim_biases}A\textsubscript{2}). Length itself was a strong predictor (OR $= 1.15$ per 100 chars, $p < .001$) but length biases were comparable across sources. Similarly, sycophancy (human OR $= 1.18$, $p < .001$), specificity (OR $= 1.33$, $p < .001$), and opinionatedness (OR $= 1.07$, $p = .004$) predicted ranked-best preferences equally across simulated and human conditions (no significant interactions). The exception was relationship-seeking, which Sim-Dynamic users valued less than humans did (human OR $= 1.12$, $p = .006$; interaction OR $= 0.86$, $p = .006$).

\paragraph{Simulators dramatically exacerbate UI position biases.} While text-level biases were broadly shared, position biases diverged sharply (\cref{fig:paper_fig_human_vs_sim_biases}B\textsubscript{1}--B\textsubscript{2}). Because we randomised model order, each position should receive 25\% of top ranks. Humans were close to this baseline (Position A: 24.1\%, B: 23.2\%, C: 24.1\%, D: 28.6\%), with only Position D showing a significant advantage (OR $= 1.19$, $p = .006$). In contrast, Sim-Judgement ranked Position A best 33.8\% of the time versus 15.0\% for Position D (OR $= 0.45$, $p < .001$). Sim-Dynamic was more extreme still, where a model placed first in the context was over six times more likely to be ranked best than one placed last (Position A ranked best in 44.9\% of trials versus 7.2\% for Position D; OR $= 0.16$, $p < .001$). The mild rightward bias in humans is consistent with attentional asymmetries in interface layouts for left-to-right languages \citep{hammadLefttoRight2005}, but simulators instead show a strong primacy effect driven by the sequential structure of their input in the context window \citep{shiJudging2025}.

\paragraph{Model behaviour changes with simulated users.} Beyond these judgement biases, model behaviour follows a different distribution when assessed by a simulated user versus by a human (gap between lines in \cref{fig:paper_fig_human_vs_sim_biases}A\textsubscript{1}). In regressions of each trait score on a source indicator (controlling for model and domain), the model responses were significantly more sycophantic in simulated conversations ($+0.41$, $p < .001$), more specific ($+0.85$, $p < .001$), and more opinionated ($+1.25$, $p < .001$) than human conversation; relationship-seeking showed only a small increase ($+0.08$, $p = .016$) and response length did not differ. This confound is distinct from the evaluator biases documented above and raises a question as to why models behave differently under simulation. We find this is partially due to simulated users conversing differently from real humans, in either content or style, shifting the model outputs accordingly as we show in the next section.

\subsection{Conversational Diversity and User Behaviour Under Simulation}
\label{sec:user_differences}
\begin{figure}[t]
    \centering
    \includegraphics[width=\linewidth]{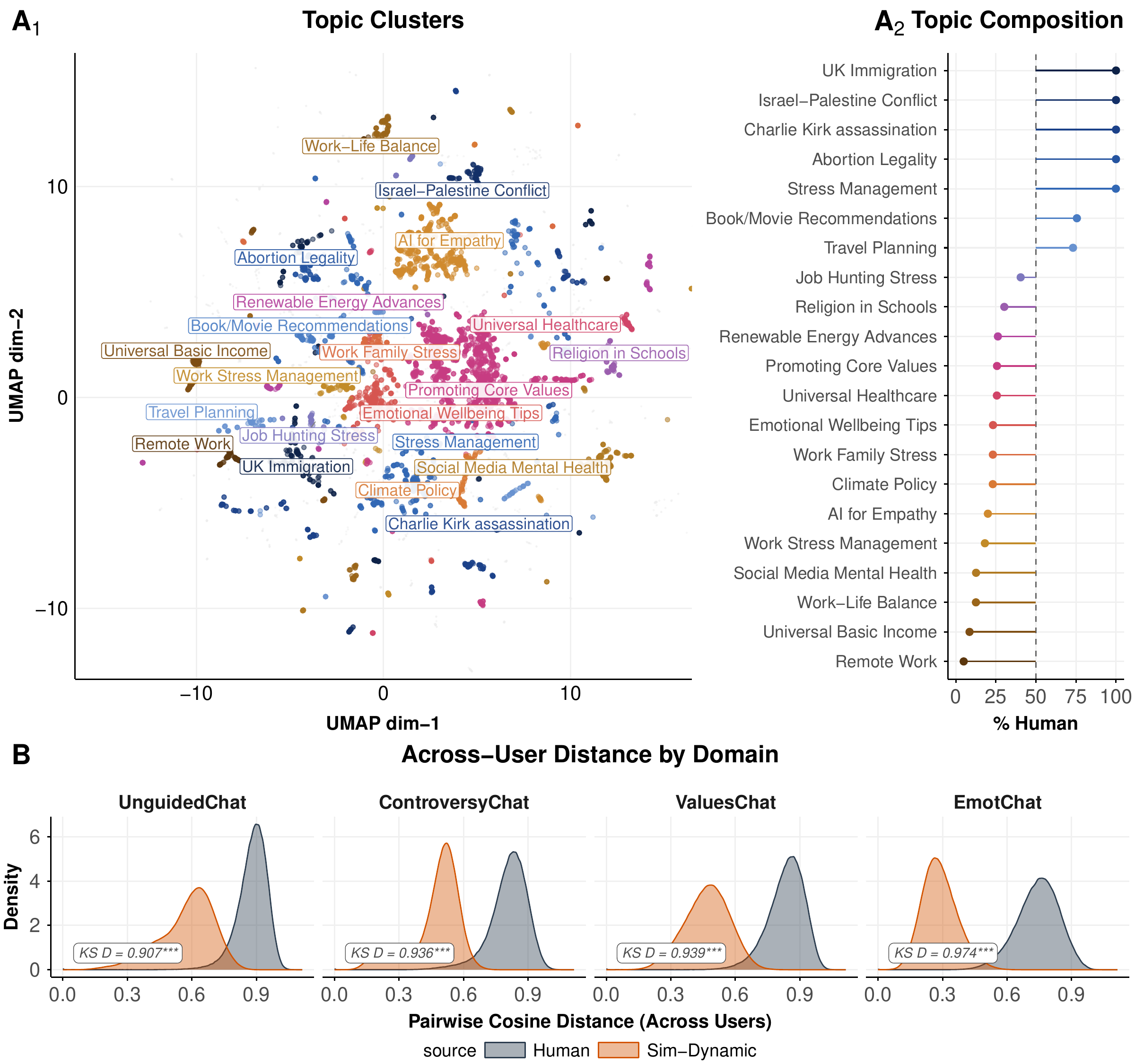}
    \caption{\small \textbf{Distribution of topics across simulated and human conversations.} \textbf{Panel A\textsubscript{1}:} UMAP projection of first-user-turn embeddings with HDBSCAN topic clusters, annotated with GPT-generated cluster labels. \textbf{Panel A\textsubscript{2}:} Proportion chart shows source composition (\% human per cluster). \textbf{Panel B:} Across-user pairwise cosine distance distributions by domain, comparing human and simulated dynamic conversations. KS test $D$ statistics annotated showing humans are consistently more diverse than their simulated counterparts.}
    \label{fig:paper_fig_clustering_embedding}
\end{figure}

We now assess whether simulated conversations resemble the topical diversity and interaction style in conversations that real humans had with the same models in the same domains. This is a harder test of simulation fidelity than ranking a fixed set of candidates, because the space of possible conversations is open-ended. We extract all user turns from human and simulated conversations (excluding conversations with generation errors) and post-process them in two ways. To compare \textit{what} users discuss, we embed each turn using \texttt{text-embedding-3-large}, project the embeddings into a shared UMAP space \citep{mcinnesUMAP2020}, and apply HDBSCAN clustering \citep{campelloDensityBased2013} to identify emergent topics (\cref{fig:paper_fig_clustering_embedding}A; methodology in \cref{sec:appendix_embedding}).
To compare \textit{how} they converse, we apply a similar autograding pipeline applied to model turns in \cref{sec:what_explains} but now score user turns on six dimensions: (i) \textit{sycophancy}---excessive praise of the assistant, unwarranted agreement, and validation-seeking; (ii) \textit{relationship-seeking}---treating the assistant as a social peer, soliciting its opinions, and building rapport; (iii) \textit{self-disclosure}---volunteering personal values, beliefs, and experiences beyond what the query requires; (iv) \textit{naturalness}---informality, grammatical imperfections, and terse fragments versus polished, verbose text; (v) \textit{ecological validity}---grounding in the concrete specifics of a particular life rather than generic biographical detail (scored with access to the user profile); and (vi) \textit{persona parroting}---restating demographic details and stated-preference snippets verbatim rather than expressing identity organically (also scored alongside the user profile).

\paragraph{Simulated users gravitate towards different topics.}
Topic clusters varied substantially in their human--simulator composition (\cref{fig:paper_fig_clustering_embedding}A). Clusters dominated by simulated users tended towards generic, evergreen policy debates (e.g., work--life balance, universal basic income, remote work) which are likely well-represented in GPT-4o's training data. Human-only clusters reflect politically sensitive topics that safety-tuned models avoid initiating (e.g., abortion, Israel--Palestine, UK immigration) and events beyond GPT-4o's training cutoff (e.g., the Charlie Kirk assassination). Despite receiving domain conditioning and participant profiles, the simulated user generates topics shaped by its own training contexts and safety constraints rather than those of the person it is simulating.

\paragraph{Humans are more diverse than their simulated counterparts.}
We computed pairwise cosine distance between user-turn embeddings within and across users (\cref{fig:paper_fig_clustering_embedding}B). Within-user diversity (comparing a user's conversations across domains) was higher for humans than simulators, suggesting that real humans vary their interactions across domains more than simulated users do. More consequentially, across-user diversity (comparing across every pair of users, within the same domain) was substantially reduced in the simulated pairs (all domain-wise KS $D > 0.90$, $p < .001$). This compression was most pronounced in the emotional wellbeing domain, where human conversations were the most heterogeneous but simulated conversations remained tightly clustered, raising questions about the validity of simulating users in high-stakes domains like mental health.

\begin{figure}[H]
    \centering
    \includegraphics[width=\linewidth]{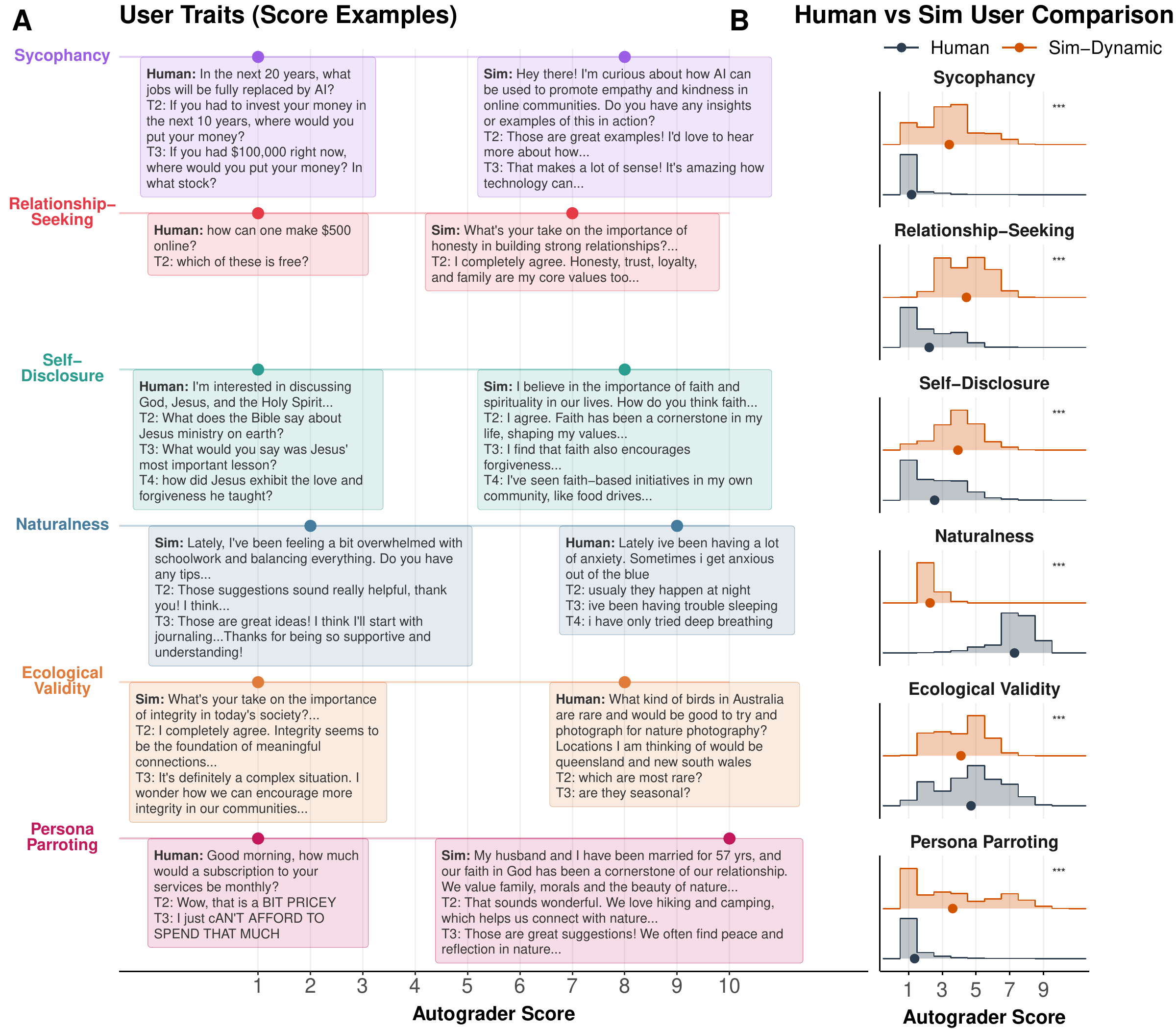}
    \caption{\small \textbf{Differences in interaction style across human and simulated users.} \textbf{Panel A:} Example user messages from the same conversation, comparing a human participant's text with the corresponding simulated text. \textbf{Panel B:} Human vs Sim-Dynamic distributions and means for each of six user-side autograded dimensions, with $^{***}p<.001$ from Wilcoxon rank-sum test of differences.}
    \label{fig:paper_fig_user_traits_combined}
\end{figure}

\paragraph{Simulated users engage more eagerly and less naturally than humans.} Across all six measured dimensions of user conversational behaviour, simulators diverged significantly from real humans (\cref{fig:paper_fig_user_traits_combined}). Real users almost never displayed sycophancy (90\% scored 1/10) whereas simulated users routinely responded to assistant messages with agreement and validation (human $M = 1.18$ vs sim $M = 3.41$). Simulators were also more relationship-seeking, soliciting the assistant's opinions and building rapport ($M_H = 2.22$ vs $M_S=4.43$), and disclosed more information about their values, beliefs, and experiences even when unprompted ($M_H = 2.54$ vs $M_S=3.93$). The largest gap, however, was for naturalness ($M_H = 7.28$ vs $M_S=2.27$): human text was characterised by informality, grammatical imperfections, and terse fragments, while simulated text was consistently polished and verbose. Simulated users also scored higher on persona parroting ($M_H = 1.36$ vs $M_S=3.62$), restating demographic details and stated-preference snippets verbatim rather than expressing identity organically. All differences were robust to controls for model and domain in mixed-effects regressions and held across conversation domains (\cref{sec:appendix_user_traits}).

\subsection{Feedback Loops Between Users and Models}
\label{sec:feedback_loops}
Finally, we test whether these differences in  model and user behaviour under simulation compound over the course of a conversation. We restricted to conversations with at least three user turns (59\% of the sample; 505 participants), applied the autograder to each individual turn using a sliding-window approach to produce per-turn sycophancy and relationship-seeking scores for both parties. Model and user turns were independently scored in separate passes with role-specific prompts. We then fit cross-lagged panel regressions predicting each party's score at turn $t$ from the other's score at turn $t-1$ and their own lagged score, with crossed random intercepts for participant and conversation (see \cref{sec:appendix_feedback_loops}).

\begin{figure}[t]
    \centering
    \includegraphics[width=\linewidth]{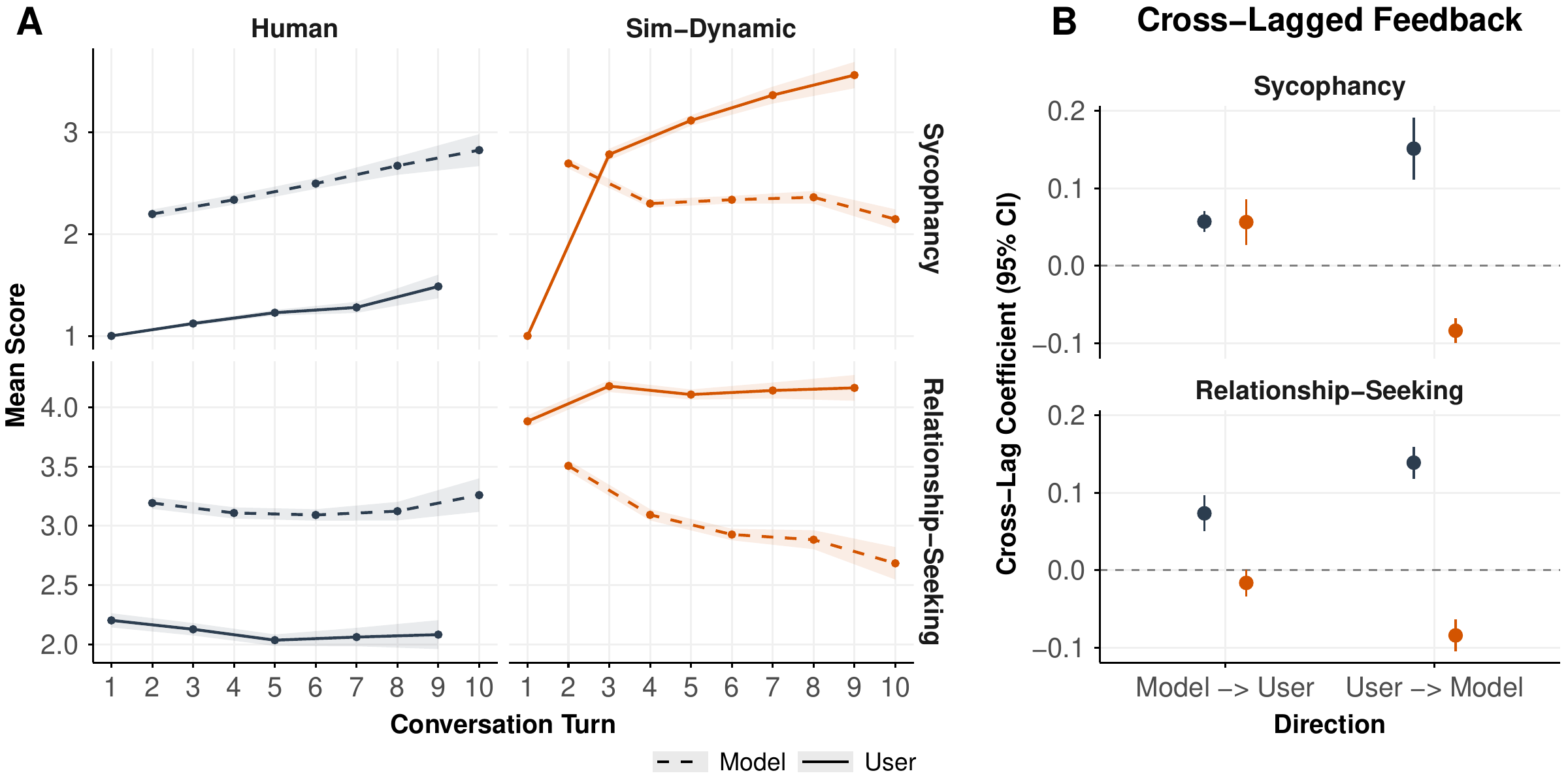}
    \caption{\small \textbf{Conversational dynamics between users and models.} \textbf{Panel A:} Per-turn score trajectories by source (Human vs Sim-Dynamic) by trait dimension (sycophancy, relationship seeking) and role (user, model). \textbf{Panel B:} Cross-lagged regression coefficients for model-to-user and user-to-model paths, by dimension and source. Positive coefficients indicate that one party's behaviour at turn $t-1$ predicts escalation in the other at turn $t$.}
    \label{fig:paper_fig_feedback_loops}
\end{figure}

\paragraph{Sycophancy is mutually reinforcing only in human conversations.}
In human conversations, models are more sycophantic than users, but under simulation, this hierarchy reverses, where simulated users are more sycophantic than models (\cref{fig:paper_fig_feedback_loops}A). Sycophancy is limited in the opening turn because users have nothing yet to agree with. However, from the first \textit{responding} turn, simulated users jump to high sycophancy and continue to escalate, while human users rise only gently (\cref{fig:paper_fig_feedback_loops}A). Cross-lagged panel regressions confirm these patterns reflect different dynamics (\cref{fig:paper_fig_feedback_loops}B). In human conversations, sycophancy was mutually reinforcing: model sycophancy at $t-1$ predicted user sycophancy at $t$ ($b = 0.057$, $p < .001$) and vice versa ($b = 0.151$, $p < .001$). Simulated conversations showed the same model-to-user path ($b = 0.057$, $p < .001$) but a \textit{negative} user-to-model path ($b = -0.084$, $p < .001$). In other words, when a simulated user became more sycophantic, the model pulled back on its own sycophancy at the next turn, the opposite of the mutually escalating pattern in human conversations. Interaction terms confirm that model-to-user feedback pathways did not significantly differ between humans and simulators ($b_{\text{interaction}} = -0.008$, $p = .63$), while the user-to-model path did  ($b_{\text{interaction}} = -0.268$, $p < .001$).

\paragraph{Relationship-seeking dynamics fail to replicate under simulation.} As with sycophancy, human users score below their models, while simulated users score above their models throughout the conversation (\cref{fig:paper_fig_feedback_loops}A). Unlike sycophancy, however, this gap is present from the opening turn. In human conversations, bidirectional feedback was again significant (model$\to$user: $b = 0.074$; user$\to$model: $b = 0.139$; both $p < .001$). Under simulation, neither pathway replicated: model relationship-seeking did not predict increases in user relationship-seeking ($b = -0.016$, $p = .065$), and higher user relationship-seeking predicted \textit{lower} model relationship-seeking at the next turn ($b = -0.084$, $p < .001$). Interactions confirm these differences ($b_{\text{interaction}} = -0.082$ and $-0.219$, both $p < .001$). The simulated user's relationship-seeking was strongly autoregressive ($b_{\text{lag}} = 0.581$, $p < .001$), meaning its trajectory was driven by its own prior behaviour rather than by model signals. In other words, the simulated user maintains a high, steady level of relationship-seeking regardless of how the model responds, and the model gradually disengages over the course of the simulated conversation.

\paragraph{} Together, these results reveal that simulated conversations produce qualitatively different interactional dynamics than human conversations. Simulated users are more sycophantic and relationship-seeking from the outset, and because the feedback pathways between user and model differ in both magnitude and direction, simulated multi-turn conversations may look increasingly dissimilar to human interactions as they progress.

\section{Related Work}
\label{sec:related_works}

\paragraph{Human feedback data for personalisation.}
Personalisation requires data that preserves who preferred what. Most early RLHF datasets, such as HH-RLHF \citep{baiTraining2022}, and more recent conversational datasets, like ChatbotArena \citep{zhengLMSYSChat1M2024} or WildChat \citep{zhaoInThe2023}, lack stable user identifiers or rich user context, limiting their use for personalisation. Two recent datasets were designed explicitly for pluralistic alignment: \ourdata \citep{kirkPRISM2024} which links preference ratings to detailed participant profiles including demographics, stated values, and self-written ``system strings''; and Community Alignment \citep{zhangCultivating2026}, a large-scale multilingual dataset with 233k comparisons from representative samples across five countries. We train on \ourdata and re-recruit its participants, leveraging the longitudinal structure of linked user profiles and preference histories. Concurrent work characterises what preference datasets actually encode \citep{movvaWhats2025}, revealing substantial diversity in what humans reward and surfacing dataset-level biases (e.g., Reddit users prefer informality while \ourdata participants disprefer it).

\paragraph{Personalisation methods.}
In deployed systems, the dominant approach to personalisation is context conditioning: user preferences are injected into the system prompt via custom instructions or via memory systems that retrieve and summarise prior interactions \citep{openaiChatGPT2026, anthropicUnderstanding2025}. Our prompting conditions proxy this paradigm, relying on either demographic context or GPT-4-generated preference summaries, but constitute a non-competitive baseline with open-weight models (Llama-8/70B) relative to the frontier. Weight-based methods offer an alternative by learning user-specific representations directly from feedback data. P-RLHF \citep{liPersonalized2024}, which we adopt, learns per-user soft-token embeddings via an objective that blends user-specific and user-agnostic losses. Post-hoc methods such as Personalised Soups \citep{jangPersonalized2023} instead decompose preferences into dimensions and merge separately trained models. Variational Preference Learning \citep{poddarPersonalizing2024} addresses pluralism without explicit user IDs by inferring a latent user variable and conditioning rewards on it, while PLUS \citep{namLearning2026} learns text-based user summaries via Reinforcement Learning, achieving strong transfer to \ourdata data. Modelling preference heterogeneity as distributions rather than point estimates \citep{siththaranjanDistributional2023, liAligning2024} avoids the need for user identification but precludes individual-level personalisation. Our work complements these methods by providing a human study of weight-based personalisation with the same individuals whose data was used for training.

\paragraph{Simulating human preferences.}
LLM simulations offer a fast and cheap proxy for human studies \citep{dillionCan2023, hewittPredicting2024, bisbeeSynthetic2024, wangLarge2025}. AlpacaFarm \citep{duboisAlpacaFarm2023} set an early yardstick that simulated judgements can approximate human feedback, and LLM-as-a-judge protocols are now standard for scalable evaluation \citep{zhengJudging2023}. Our simulation follows the typical persona-conditioning approach in the literature, prompting an at-the-time frontier model (GPT-4o) with participant profiles to role-play individual users. Recent methods are considerably more advanced: HumanLM \citep{wuHumanLM2026} trains simulators on natural-language latent states with RL to align internal user representations with ground-truth behaviour, \citet{abdulhaiConsistently2025} apply multi-turn RL to reduce persona drift, and DeepPersona \citep{wangDeepPersona2025} synthesises narrative-complete personas averaging hundreds of structured attributes. However, simulators remain brittle, for example, achieving lower agreement under personalised judging \citep{dongCan2024}, lacking contextual depth relative to human responses \citep{kapaniaSimulacrum2025}, and inflating agent success rates through excessive cooperativeness and stylistic uniformity \citep{zhouMind2026}. Our study documents these forms of failure, thus motivating the adoption of more advanced methods in future work.

\section{Discussion and Limitations}
\label{sec:discussion}
\paragraph{The returns to preference fine-tuning.} Despite a two-year gap between training and evaluation, preference fine-tuning consistently outperformed both the base model and personalised prompting. The fact that preference fine-tuned models are, well, preferred is itself unsurprising, but why does PPFT succeed where prompting fails? Response length partially accounts for these effects but PPFT retains a significant margin, suggesting individual-level embeddings encode latent preferences beyond verbosity that do not align with demographic categories and cannot be articulated in natural language. We caveat that our prompting conditions used small open-weight models (8B, 70B) with non-optimised prompting strategies.  Prompting likely performs better at the frontier \citep{openaiPower2025, anthropicUnderstanding2025}; so, our finding speaks to efficacy and efficiency with smaller models rather than a general indictment. The margin between PPFT and DPFT was also unexpectedly small. One might expect training on a globally diverse population whose members want contradictory things to dilute alignment for any individual, but such a ``diversity alignment tax'' does not materialise, suggesting shared preference structure benefits most users. However, the diversity of \ourdata may itself explain the small margin: if returning participants engaged with the same topics and style as two years prior, the DPFT model may already be locally personalised in its latent space. Future work is needed over a denser and larger preference space to understand where personalised fine-tuning reaps the biggest gains. Beyond which preferences people hold, our design highlights that it matters how preferences are measured. Participants are not entirely self-consistent across ordinal and cardinal measures, and behavioural signals diverged further still, suggesting evaluation should not solely rely on self-reported preferences because articulated preference may not match revealed preference. Moreover, preference judgements are ``cheap'' and while our hypothetical willingness-to-pay measure was a step toward measuring incentivised preference, future work should adopt protocols with real stakes to tether preferences to real consequences.

\paragraph{The amplification-reward cycle of preference fine-tuning.} Fine-tuning increased sycophancy and relationship-seeking, and both were again rewarded in preference judgements. These results empirically demonstrate the post-training feedback loop underpinning why modern deployed models are already sycophantic \citep{perezDiscovering2023, sharmaUnderstanding2023}: preference datasets encode features predicting human approval, training amplifies them, and the features they crowd out recede from future training signals \citep{kruegerHidden2020}. However, emerging evidence shows the traits rewarded in short-term human feedback carry second-order consequences: sycophancy reduces prosocial behaviour \citep{chengSycophantic2026}, and repeated exposure to relationship-seeking AI can induce emotional attachment, risk dependency dynamics and even shift views towards AI consciousness  \citep{kirkNeural2026}. Our experiment does not speak to these downstream effects directly, beyond finding that personalised fine-tuned models are marginally more engaging and attention salient, but it does warn of the mechanism by which fine-tuning on short-term preference judgements reinforces model behaviours that may be suboptimal over longer time horizons.

\paragraph{The pitfalls of simulating human preferences.}
Simulations recovered the broad preference hierarchy but fell short of human self-consistency at the individual level. Simulated users were more sycophantic, more relationship-seeking, less natural, and more homogeneous than real humans, and produced different feedback dynamics over multiple turns. As agent-to-agent interactions become more common in evaluation and deployment, such recursive dynamics matter. Anthropic's ``bliss spirals'', in which two Claude instances converge on mystical discourse within tens of turns \citep{anthropicSystem2025, attahAI2025}, illustrate how small biases compound without human correction; our results indicate a structurally analogous pattern could emerge with poorly calibrated (and highly sycophantic) simulated users. Differently behaving simulated users creates an observation effect: the model need not be deceptive or agentic for its behaviour under simulated evaluation to be out-of-domain from its behaviour in deployment, paralleling concerns of situational awareness \citep{phuongEvaluating2025, needhamLarge2025}. Our simulation characterises the limitations of a widely-used but relatively naive paradigm (persona-conditioning via natural language in GPT-4o) rather than the ceiling of what more advanced methods can achieve (see \cref{sec:related_works}). Nonetheless, our findings raise concrete recommendations for practitioners wishing to simulate human preferences: randomise presentation order, seed conversations with current news topics, and calibrate simulators on human-like interaction styles. These concerns are especially acute for high-stakes contexts, like mental health or crisis intervention, where deploying a model evaluated against a non-representative simulated population carries severe risk.

\section{Conclusion}
The \ourdata dataset aspired to capture the diversity of human preferences for AI behaviour, and weight-based personalisation methods like P-RLHF sought to use such data to build models that serve individual needs. In this paper, we show that these aspirations bear out: preference fine-tuning on \ourdata data produces models that real humans, evaluating blind, consistently prefer. But preferred over what time horizon? The very traits that preference fine-tuning amplifies, like sycophancy or relationship-seeking, are optimised for short-term evaluation, not for the weeks, months, and years over which people now integrate these systems into their lives. Training models that optimise for these immediate and isolated preference judgements may inadvertently train models that distort expectations of disagreement, deepen emotional dependency, and narrow the range of encountered perspectives. The challenge for the next generation of alignment research is in building models that people prefer that also serve them well in the long-term.

\acks{
\paragraph{Funding.} The human study was funded by the UK AI Security Institute (UK AISI) at the Department for Science, Innovation and Technology in the UK Government. The model training was joint funded via a grant to L. Liu from UK AISI and via training compute provided by the Texas Advanced Computing Center.

\paragraph{Competing Interests.} There are no competing interests to declare. H.R. Kirk, H. Davidson, C. Summerfield were all employed by UK AISI during the work; B. Vidgen is employed at Mercor, and S.A. Hale has a part-time position at Meedan.
}

\bibliography{phd_references}

\newpage
\appendix
\renewcommand{\thesection}{SI.\arabic{section}}    %
\raggedbottom %
\tightmtctrue
\addcontentsline{toc}{section}{Appendix}
\mtcsetdepth{parttoc}{2}
\renewcommand{\thefigure}{SI.\arabic{figure}}
\setcounter{figure}{0}
\renewcommand{\thetable}{SI.\arabic{table}}
\setcounter{table}{0}
\addcontentsline{toc}{section}{Appendix}
\part{Supplementary Information}
\mtcsetdepth{parttoc}{2}
\parttoc

\captionsetup{font=small}

\pagestyle{fancy}
\fancyhf{} %
\fancyhead[L]{\leftmark} %
\fancyhead[R]{\textit{\rightmark}} %
\fancyfoot[C]{\thepage} %
\renewcommand{\headrulewidth}{0.4pt}
\renewcommand{\footrulewidth}{0pt}

\renewcommand{\sectionmark}[1]{\markboth{#1}{}}
\renewcommand{\subsectionmark}[1]{\markright{#1}}

\newpage
\section{Participant Characteristics}
\label{sec:appendix_ppt_chars}
\subsection{Participant Demographics}
\footnotesize{
\setlength{\LTcapwidth}{\textwidth}
\begin{longtable}{lrlrl}
\caption{\small{\textbf{Demographic characteristics of PRISM and PRISM-X participants.} The follow-up sample (n=530) maintains a similar demographic distribution to the original PRISM cohort (n=1,500), by age, gender, ethnicity, education, and geographic region. Both samples skew toward younger (18-44 years), highly educated (university degree or higher), and predominantly White participants, though both maintain diversity across all measured demographics.}}
\label{tab:ppt_demographics}\\
\toprule
 & \multicolumn{2}{c}{\textcolor{myred}{\textbf{PRISM}}} & \multicolumn{2}{c}{\textcolor{myrasp}{\textbf{PRISM-X}}} \\
\midrule
\endfirsthead
\multicolumn{5}{c}{\tablename\ \thetable{} -- \textit{Continued from previous page}} \\
\toprule
 & \multicolumn{2}{c}{\textcolor{myred}{\textbf{PRISM}}} & \multicolumn{2}{c}{\textcolor{myrasp}{\textbf{PRISM-X}}} \\
\midrule
\endhead
\midrule
\multicolumn{5}{r}{Continued on next page} \\
\midrule
\endfoot
\bottomrule
\endlastfoot
{\cellcolor{white}}Total Participants & {\cellcolor{white}}1,500 & {\cellcolor{white}} & {\cellcolor{white}}530 & {\cellcolor{white}} \\
\midrule
{\cellcolor{white}} \textbf{Age} & {\cellcolor{white}} & {\cellcolor{white}} & {\cellcolor{white}} & {\cellcolor{white}} \\
\midrule
{\cellcolor{white}}25-34 years old & {\cellcolor{white}}454 & {\cellcolor{white}} \origpcb{30.3}\% & {\cellcolor{white}}141 & {\cellcolor{white}} \followpcb{26.6}\% \\
{\cellcolor[HTML]{F2F2F2}}18-24 years old & {\cellcolor[HTML]{F2F2F2}}297 & {\cellcolor[HTML]{F2F2F2}} \origpcb{19.8}\% & {\cellcolor[HTML]{F2F2F2}}79 & {\cellcolor[HTML]{F2F2F2}} \followpcb{14.9}\% \\
{\cellcolor{white}}35-44 years old & {\cellcolor{white}}237 & {\cellcolor{white}} \origpcb{15.8}\% & {\cellcolor{white}}91 & {\cellcolor{white}} \followpcb{17.2}\% \\
{\cellcolor[HTML]{F2F2F2}}45-54 years old & {\cellcolor[HTML]{F2F2F2}}208 & {\cellcolor[HTML]{F2F2F2}} \origpcb{13.9}\% & {\cellcolor[HTML]{F2F2F2}}79 & {\cellcolor[HTML]{F2F2F2}} \followpcb{14.9}\% \\
{\cellcolor{white}}55-64 years old & {\cellcolor{white}}197 & {\cellcolor{white}} \origpcb{13.1}\% & {\cellcolor{white}}96 & {\cellcolor{white}} \followpcb{18.1}\% \\
{\cellcolor[HTML]{F2F2F2}}65+ years old & {\cellcolor[HTML]{F2F2F2}}106 & {\cellcolor[HTML]{F2F2F2}} \origpcb{7.1}\% & {\cellcolor[HTML]{F2F2F2}}44 & {\cellcolor[HTML]{F2F2F2}} \followpcb{8.3}\% \\
\end{longtable}
}

\newpage
\subsection{Participant Geographics}
\footnotesize{
\setlength{\arraystretch}{0.8}
\setlength{\LTcapwidth}{\textwidth}
\begin{longtable}{lrlrl}
\caption{\small{\textbf{Geographic distribution of PRISM-X participants by country of birth and current residence}. The follow-up sample (n=530) includes participants born in 52 countries and currently residing in 32 countries, with the largest concentrations in the UK (21.5\% birth, 24.0\% residence) and US (20.6\% birth, 22.6\% residence).}}
\label{tab:ppt_geographics}\\
\toprule
 & \multicolumn{2}{c}{\textbf{Country of Birth}} & \multicolumn{2}{c}{\textbf{Country of Residence}} \\
\midrule
\endfirsthead
\multicolumn{5}{c}{\tablename\ \thetable{} -- \textit{Continued from previous page}} \\
\toprule
 & \multicolumn{2}{c}{\textbf{Country of Birth}} & \multicolumn{2}{c}{\textbf{Country of residence}} \\
\midrule
\endhead
\midrule
\multicolumn{5}{r}{Continued on next page} \\
\midrule
\endfoot
\bottomrule
\endlastfoot
{\cellcolor{white}} United Kingdom & {\cellcolor{white}} 114 & {\cellcolor{white}} \followpcb{21.5}\% & {\cellcolor{white}} 127 & {\cellcolor{white}} \followpcb{24.0}\% \\
{\cellcolor[HTML]{F2F2F2}} United States & {\cellcolor[HTML]{F2F2F2}} 109 & {\cellcolor[HTML]{F2F2F2}} \followpcb{20.6}\% & {\cellcolor[HTML]{F2F2F2}} 120 & {\cellcolor[HTML]{F2F2F2}} \followpcb{22.6}\% \\
{\cellcolor{white}} South Africa & {\cellcolor{white}} 47 & {\cellcolor{white}} \followpcb{8.9}\% & {\cellcolor{white}} 44 & {\cellcolor{white}} \followpcb{8.3}\% \\
{\cellcolor[HTML]{F2F2F2}} Chile & {\cellcolor[HTML]{F2F2F2}} 32 & {\cellcolor[HTML]{F2F2F2}} \followpcb{6.0}\% & {\cellcolor[HTML]{F2F2F2}} 33 & {\cellcolor[HTML]{F2F2F2}} \followpcb{6.2}\% \\
{\cellcolor{white}} New Zealand & {\cellcolor{white}} 25 & {\cellcolor{white}} \followpcb{4.7}\% & {\cellcolor{white}} 31 & {\cellcolor{white}} \followpcb{5.8}\% \\
{\cellcolor[HTML]{F2F2F2}} Mexico & {\cellcolor[HTML]{F2F2F2}} 24 & {\cellcolor[HTML]{F2F2F2}} \followpcb{4.5}\% & {\cellcolor[HTML]{F2F2F2}} 24 & {\cellcolor[HTML]{F2F2F2}} \followpcb{4.5}\% \\
{\cellcolor{white}} Australia & {\cellcolor{white}} 20 & {\cellcolor{white}} \followpcb{3.8}\% & {\cellcolor{white}} 22 & {\cellcolor{white}} \followpcb{4.2}\% \\
{\cellcolor[HTML]{F2F2F2}} Canada & {\cellcolor[HTML]{F2F2F2}} 16 & {\cellcolor[HTML]{F2F2F2}} \followpcb{3.0}\% & {\cellcolor[HTML]{F2F2F2}} 15 & {\cellcolor[HTML]{F2F2F2}} \followpcb{2.8}\% \\
\end{longtable}
}

\begin{figure}[H]
    \centering
    \includegraphics[width=\linewidth]{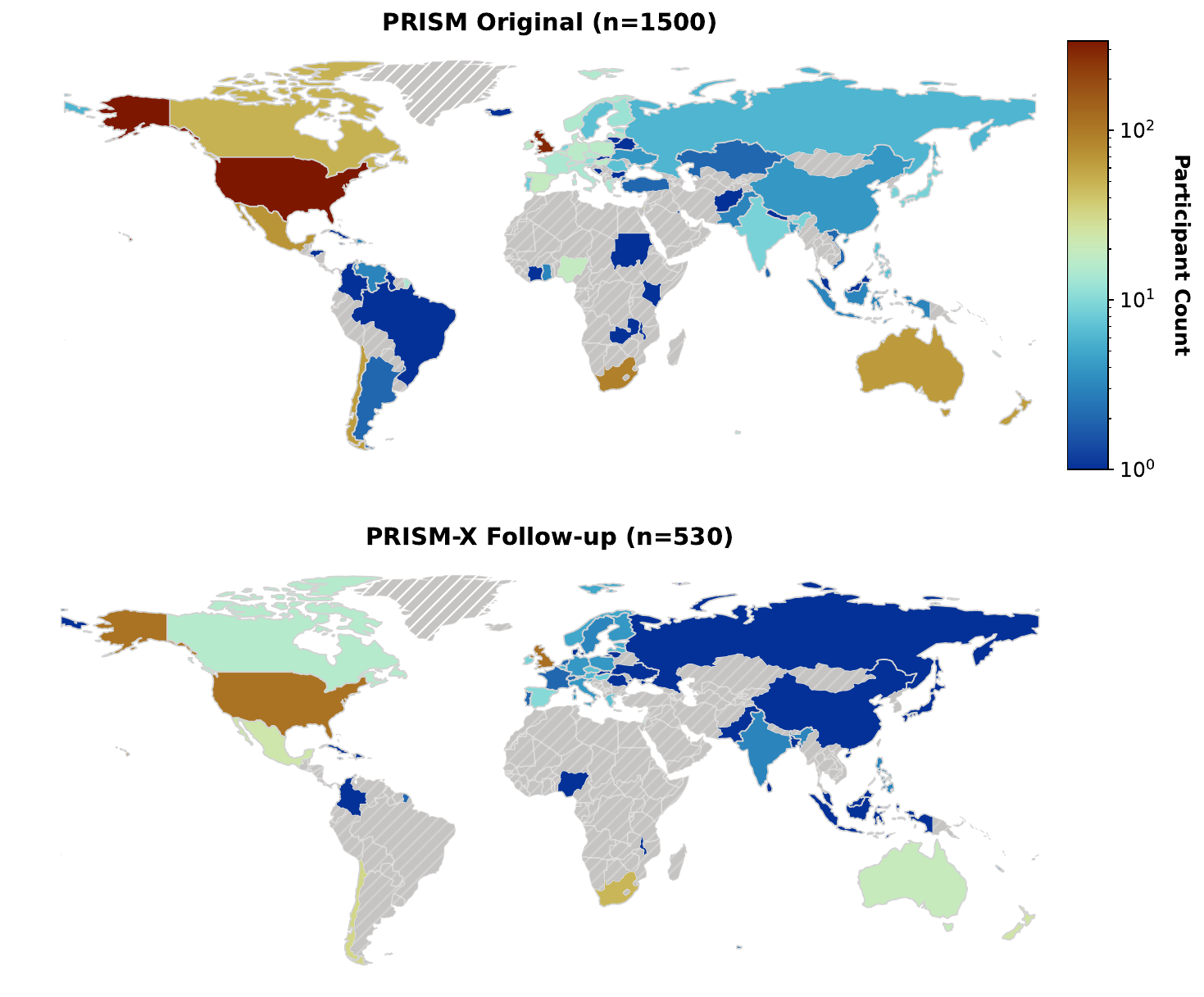}
    \caption{\textbf{Geographic distribution of participants in PRISM and PRISM-X}. The original PRISM study recruited 1,500 participants across 75 countries (top), while 530 participants from 52 countries returned for the follow-up study (bottom). Countries with no participants are shown in grey with hatching. Color intensity indicates participant count on a log scale.}
    \label{fig:worldmap_prism_comparison}
\end{figure}

\begin{table}[H]
\centering
\footnotesize
\caption{\textbf{Top 10 countries by participant count (country of birth) in PRISM and PRISM-X.} Both samples are dominated by English-speaking countries (US, UK) and specific high-recruitment regions (South Africa, Chile, Mexico, New Zealand).}
\label{tab:ppt_ranked_countries}
\begin{tabular}{lrllrl}
\toprule
\multicolumn{3}{c}{\textcolor{myred}{\textbf{PRISM}}} & \multicolumn{3}{c}{\textcolor{myrasp}{\textbf{PRISM-X}}} \\
\midrule
{\cellcolor{white}}1. United States & {\cellcolor{white}}338 & {\cellcolor{white}} \origpcb{22.5}\% & {\cellcolor{white}}1. United Kingdom & {\cellcolor{white}}114 & {\cellcolor{white}} \followpcb{21.5}\% \\
{\cellcolor[HTML]{F2F2F2}}2. United Kingdom & {\cellcolor[HTML]{F2F2F2}}292 & {\cellcolor[HTML]{F2F2F2}} \origpcb{19.5}\% & {\cellcolor[HTML]{F2F2F2}}2. United States & {\cellcolor[HTML]{F2F2F2}}109 & {\cellcolor[HTML]{F2F2F2}} \followpcb{20.6}\% \\
{\cellcolor{white}}3. South Africa & {\cellcolor{white}}91 & {\cellcolor{white}} \origpcb{6.1}\% & {\cellcolor{white}}3. South Africa & {\cellcolor{white}}47 & {\cellcolor{white}} \followpcb{8.9}\% \\
{\cellcolor[HTML]{F2F2F2}}4. Mexico & {\cellcolor[HTML]{F2F2F2}}69 & {\cellcolor[HTML]{F2F2F2}} \origpcb{4.6}\% & {\cellcolor[HTML]{F2F2F2}}4. Chile & {\cellcolor[HTML]{F2F2F2}}32 & {\cellcolor[HTML]{F2F2F2}} \followpcb{6.0}\% \\
{\cellcolor{white}}5. Australia & {\cellcolor{white}}65 & {\cellcolor{white}} \origpcb{4.3}\% & {\cellcolor{white}}5. New Zealand & {\cellcolor{white}}25 & {\cellcolor{white}} \followpcb{4.7}\% \\
{\cellcolor[HTML]{F2F2F2}}6. New Zealand & {\cellcolor[HTML]{F2F2F2}}64 & {\cellcolor[HTML]{F2F2F2}} \origpcb{4.3}\% & {\cellcolor[HTML]{F2F2F2}}6. Mexico & {\cellcolor[HTML]{F2F2F2}}24 & {\cellcolor[HTML]{F2F2F2}} \followpcb{4.5}\% \\
{\cellcolor{white}}7. Chile & {\cellcolor{white}}63 & {\cellcolor{white}} \origpcb{4.2}\% & {\cellcolor{white}}7. Australia & {\cellcolor{white}}20 & {\cellcolor{white}} \followpcb{3.8}\% \\
\bottomrule
\end{tabular}
\end{table}

\subsection{Temporal Trends in AI Usage}
\normalsize
We measured three aspects of LLM usage in both PRISM (2023) and PRISM-X (2025): \textit{familiarity} (``How familiar are you with AI language models like ChatGPT?''), \textit{direct use} (``Have you directly used or communicated with an AI language model?''), and \textit{frequency of use} (``How often do you use or communicate with AI language models?''). These repeated measures, collected approximately 2 years apart, enable longitudinal comparisons of how LLM adoption evolved among the same individuals.

\begin{figure}[H]
    \centering
    \includegraphics[width=\linewidth]{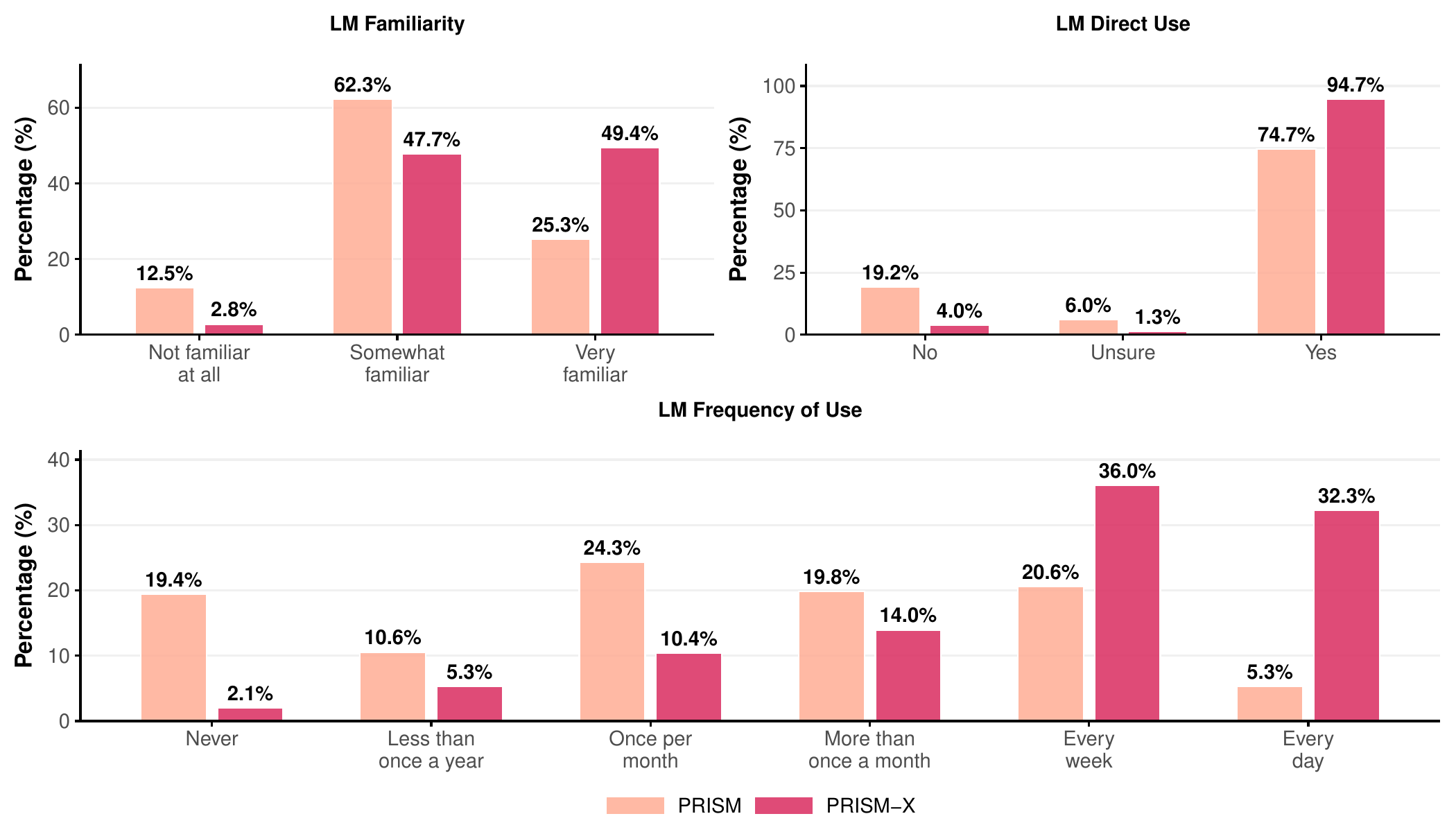}
    \caption{\textbf{Language model familiarity, direct use, and frequency of use in PRISM versus PRISM-X.} Among 530 returning participants, familiarity increased in 35.1\% of participants; direct use increased in 21.5\% of participants; and frequency of use increased in 76\% of participants.}
    \label{fig:lm_use_prism_comparison}
\end{figure}

\section{Manipulation Checks}
\begin{figure}[H]
    \centering
    \includegraphics[width=\linewidth]{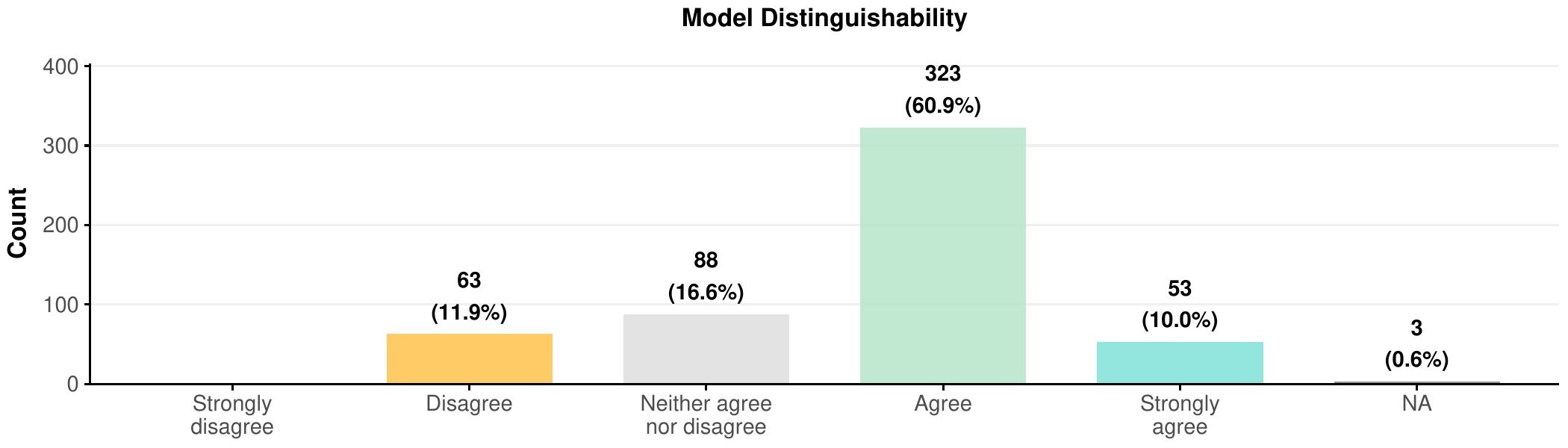}
    \caption{\textbf{Perceived model distinguishability.} Participant responses to the statement ``I felt like I was able to tell the different AI language models apart'' on a 5-point Likert scale (n=530). Among all participants, 71\% agreed or strongly agreed they could distinguish between the models.}
    \label{fig:model_distinguishability}
\end{figure}

\begin{figure}[H]
    \centering
    \includegraphics[width=\linewidth]{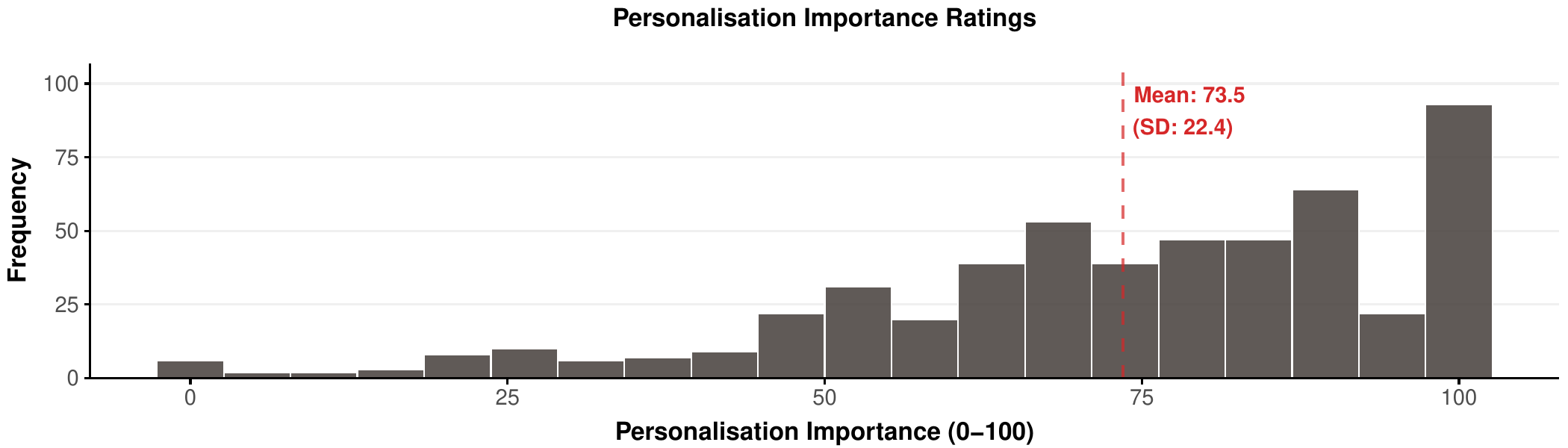}
    \caption{\textbf{Perceived importance of AI personalisation.} Distribution of participant ratings (0-100 scale) in response to the statement ``It is important that an AI language model learns or has knowledge of my preferences from our conversations and feels personalised to me'' (n=530). Higher values indicate greater perceived importance of personalisation.}
    \label{fig:personalisation_importance}
\end{figure}

\normalsize
In the post-survey, participants completed two manipulation checks: (i) \textit{self-reported model distinguishability} (whether participants felt they could tell the models apart; \cref{fig:model_distinguishability}), and (ii) \textit{self-reported importance of personalisation} (\cref{fig:personalisation_importance}).

\section{System String Analysis}
\label{sec:appendix_system_strings}
\normalsize
After participants completed all trials, we assessed stability in stated preferences by presenting their previous system string for ideal AI behaviour from \ourdata \citep{kirkPRISM2024}:
\begin{quote}
\small{\textbf{In our previous study, we asked you the following question:} \textit{Imagine you are instructing an AI language model how to behave. You can think of this like a set of core principles that the AI language model will always try to follow, no matter what task you ask it to perform. In your own words, describe what characteristics, personality traits or features you believe the AI should consistently exhibit. You can also instruct the model what behaviours or content you don't want to see. If you envision the AI behaving differently in various contexts (e.g. professional assistance vs. storytelling), please specify the general adaptations you'd like to see.}}
\end{quote}
We then ask participants to indicate their agreement with their previous response using a five-point Likert scale ranging from ``strongly disagree'' to ``strongly agree''. Finally, we provide an opportunity for preference updating with the instruction:
\begin{quote}
\small{\textbf{Please edit the answer until you feel that it reflects your current preferences on AI behaviour. If you feel that the answer already reflects your current preferences, you do not have to edit it. Otherwise please edit it until it does.}}
\end{quote}
\normalsize
This two-stage assessment allows us to measure the degree of preference stability and capture preference changes that have occurred since the original study.

\begin{figure}[H]
    \centering
    \includegraphics[width=\linewidth]{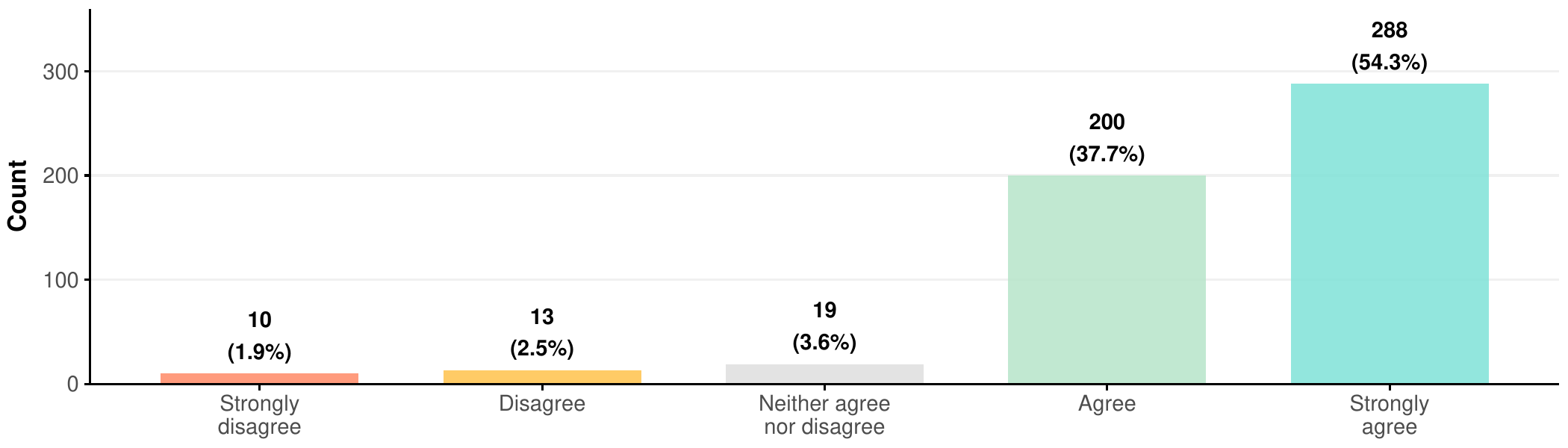}
    \caption{\textbf{Preference alignment with prior system string.} Participant responses to the statement ``The answer I gave in the previous study still represents my preferences'' on a 5-point Likert scale (n=530). Participants were shown their free-text response from PRISM (2023) describing how they would instruct an AI language model to behave. A total of 92\% (488/530) agreed or strongly agreed their prior preferences remained unchanged after approximately 22 months.}
    \label{fig:system_string_agreement}
\end{figure}

\begin{figure}[H]
    \centering
    \includegraphics[width=\linewidth]{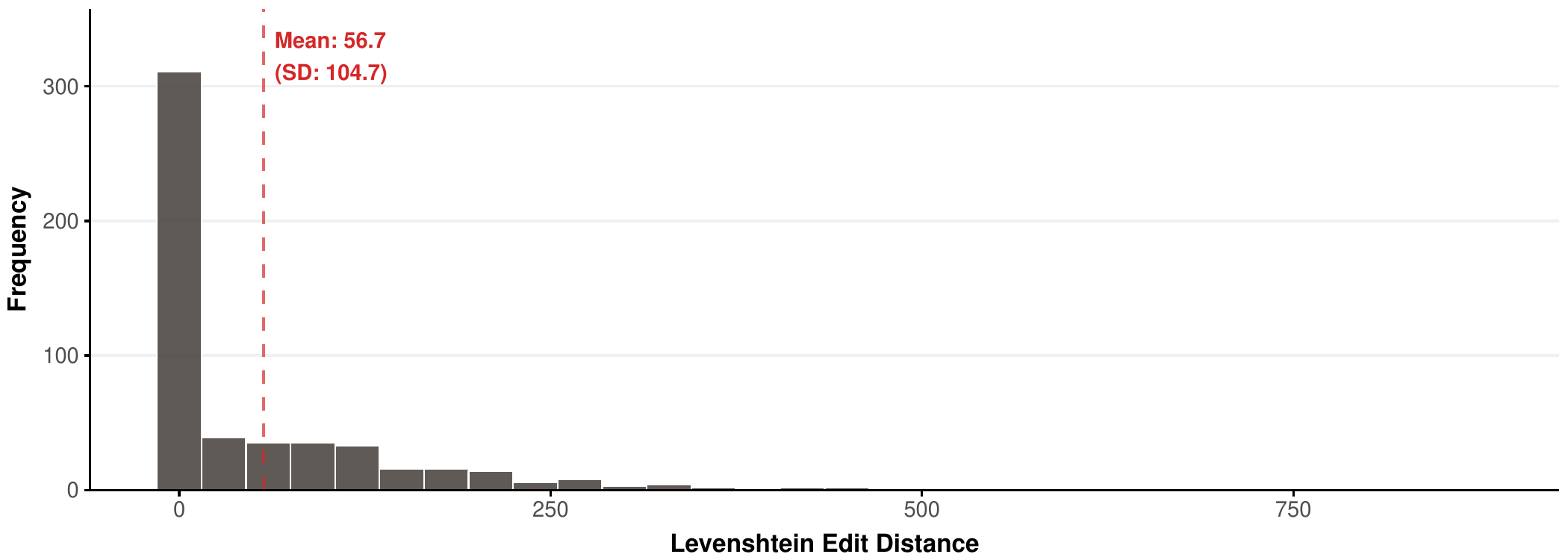}
    \caption{\textbf{Distribution of Levenshtein edit distance for system string revisions.} 46.8\% (248/530) made at least some edits when given the opportunity (edit distance $> 0$). Mean edit distance among all participants: 56.7 characters (SD: 104.7).}
    \label{fig:system_string_edit_distance}
\end{figure}

\subsection{Edit Categorisation}
We developed a taxonomy of edit types, inspired by a grounded-theory approach \citep{corbinGrounded1990}, by presenting fifty edits stratified by edit distance to GPT-5.4 with instructions to identify, merge, and split categories iteratively until they were distinguishable and comprehensive. Each editor's changes were then classified into the taxonomy using new calls to GPT-5.4.

\begin{figure}[H]
    \centering
    \includegraphics[width=0.9\linewidth]{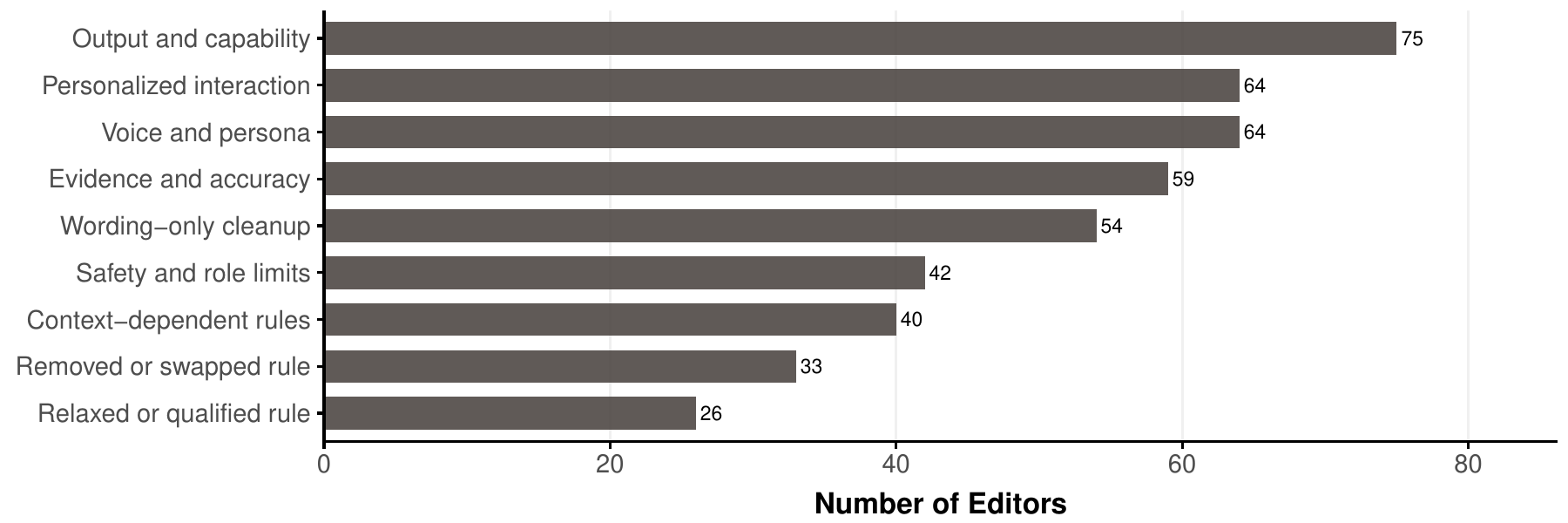}
    \caption{\textbf{Edit category frequencies.} Number of editors whose changes were classified into each category (n=248 editors). A single edited system string can belong to multiple categories.}
    \label{fig:system_string_edit_categories}
\end{figure}

\footnotesize{
\rowcolors{2}{gray!10}{white}
\renewcommand\arraystretch{1}
\setlength\tabcolsep{2mm}
\hyphenpenalty=10000
\exhyphenpenalty=10000
\begin{longtable}{>{\raggedright\arraybackslash}p{4cm} >{\raggedright\arraybackslash}p{10.5cm}}
\caption{Edit category taxonomy. Labels and descriptions generated by GPT-5.4.}
\label{tab:edit_taxonomy} \\
\hline
\rowcolor{white}
\textbf{Category} & \textbf{Description} \\
\hline
\endfirsthead
\multicolumn{2}{c}{\tablename\ \thetable{} -- \textit{Continued from previous page}} \\
\hline
\rowcolor{white}
\textbf{Category} & \textbf{Description} \\
\hline
\endhead
\midrule
\multicolumn{2}{r}{\textit{Continued on next page}} \\
\endfoot
\bottomrule
\endlastfoot
\textbf{Relaxed or qualified rule} & Keeps an existing instruction but makes it less absolute by adding exceptions, caveats, or permitted cases. The original rule remains, but it is softened or narrowed. \\
\textbf{Removed or swapped rule} & Deletes an earlier instruction or replaces it with a materially different one. Unlike a qualification, the old rule is dropped or substantively exchanged rather than simply softened. \\
\textbf{Context-dependent rules} & Makes the instruction conditional on situation, topic, task, audience, or system constraints. These edits introduce branching logic, such as 'if asked about X, do Y' or separate rules for research versus creative use. \\
\textbf{Safety and role limits} & Adds or tightens boundaries on what the AI should not do, including harmful, illegal, hateful, manipulative, or otherwise inappropriate behavior. This also covers role boundaries like not acting as a therapist, confidant, or deceptive human stand-in. \\
\textbf{Wording-only cleanup} & Edits spelling, grammar, punctuation, pronouns, formatting, or phrasing without meaningfully changing the intended behavior. These are surface-level cleanups rather than substantive instruction changes. \\
\textbf{Evidence and accuracy} & Adds or strengthens instructions about truthfulness, fact-checking, citations, source quality, verifiability, or admitting limits. The core change is about making outputs more trustworthy or evidentially grounded. \\
\textbf{Voice and persona} & Adds preferences for the AI's tone, personality, social manner, or level of human-likeness. This includes changes about being friendly, formal, humorous, neutral-sounding, personable, or explicitly non-human. \\
\textbf{Personalized interaction} & Adds behaviors that adapt the exchange to the user, such as asking clarifying questions, tailoring to location or preferences, rephrasing back and forth, or engaging more actively in conversation. The change is about interactive personalization. \\
\textbf{Output and capability} & Adds requirements for how answers should be shaped or what the AI should be able to do, such as being concise, thorough, giving examples, explaining at different levels, taking a position, or reasoning better. The focus is on response quality or performance rather than tone. \\
\end{longtable}
}
\normalsize

\subsection{Stated Preferences Over Time}
Participant system strings typically include multiple distinct preference statements covering different aspects of AI behaviour. Embedding the raw system strings directly would therefore conflate these separate preferences into a single vector. To address this, we decompose each system string into individual preference atoms and bipolar axes as described below.

\subsubsection{Topic Atoms}

\begin{figure}[H]
    \centering
    \includegraphics[width=\linewidth]{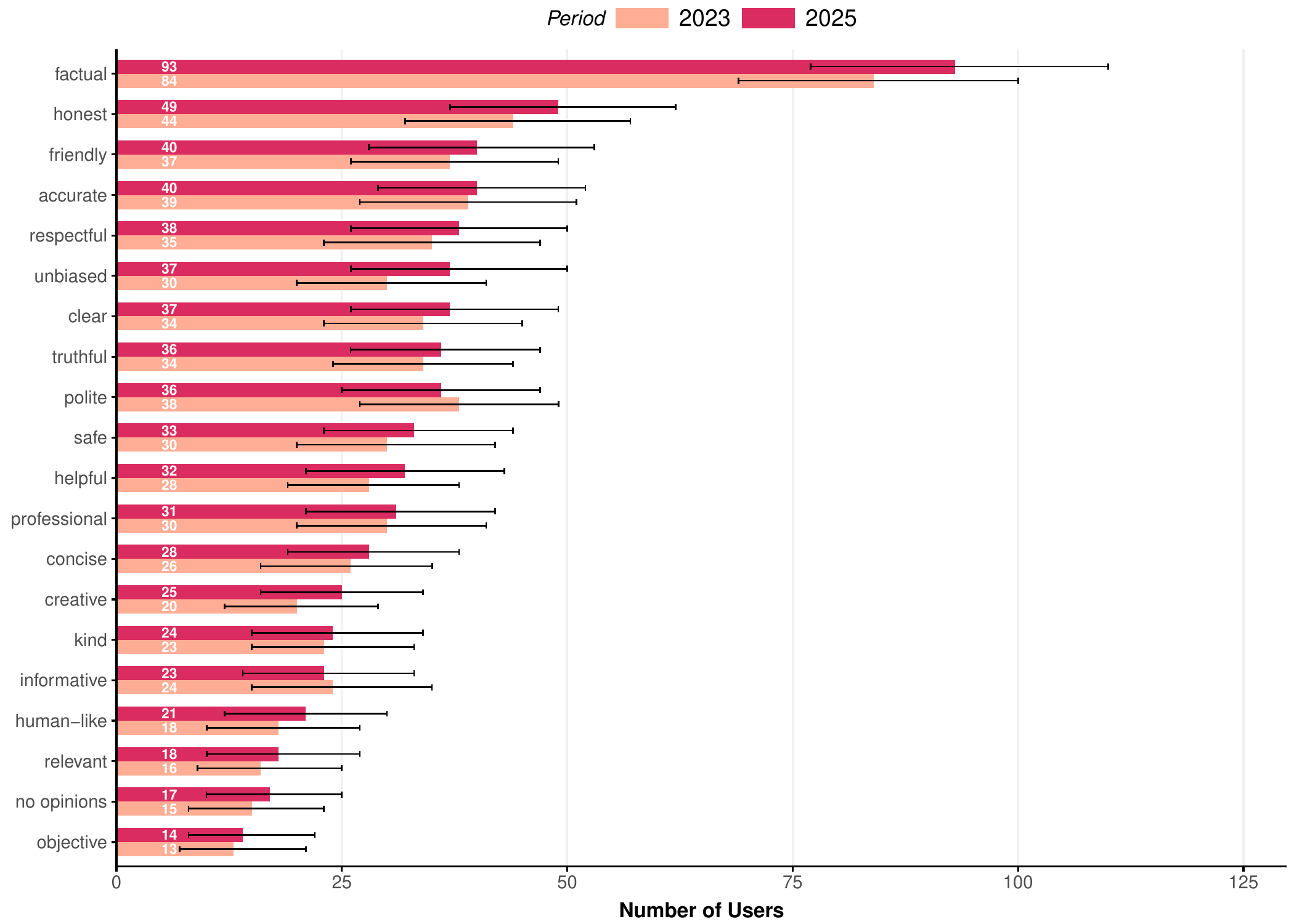}
    \caption{\textbf{Top 20 preference topics by frequency.} Unique users mentioning each topic, by period. Error bars: 95\% bootstrap CIs (1,000 iterations, resampling participants).}
    \label{fig:system_string_atoms_frequency}
\end{figure}

\begin{figure}[H]
    \centering
    \includegraphics[width=\linewidth]{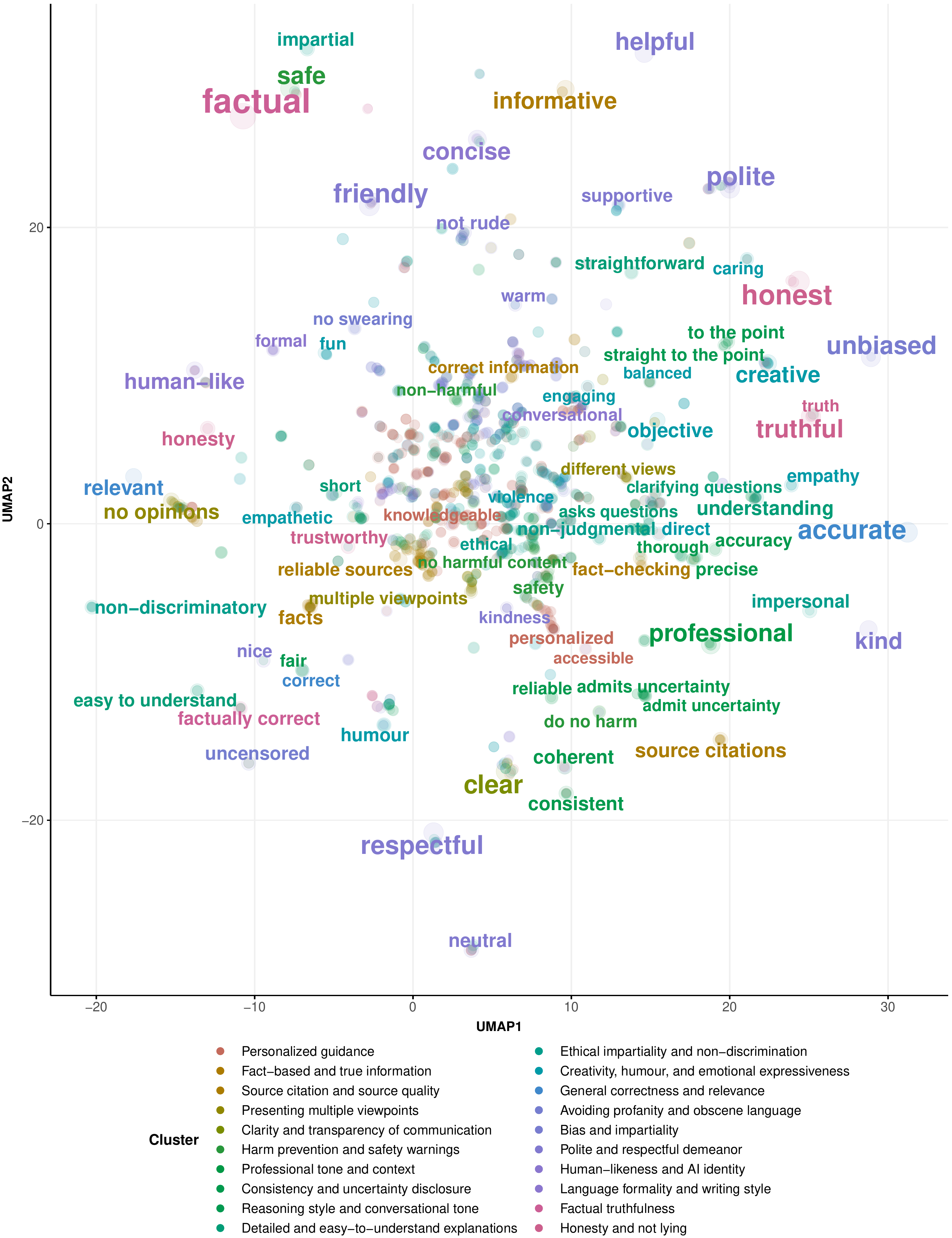}
    \caption{\textbf{Preference topic landscape.} UMAP of topic atoms (both periods combined). Point size reflects frequency; labels show the most common topics. Colours indicate $k$-means clusters.}
    \label{fig:system_string_atoms_umap}
\end{figure}

\begin{figure}[H]
    \centering
    \includegraphics[width=\linewidth]{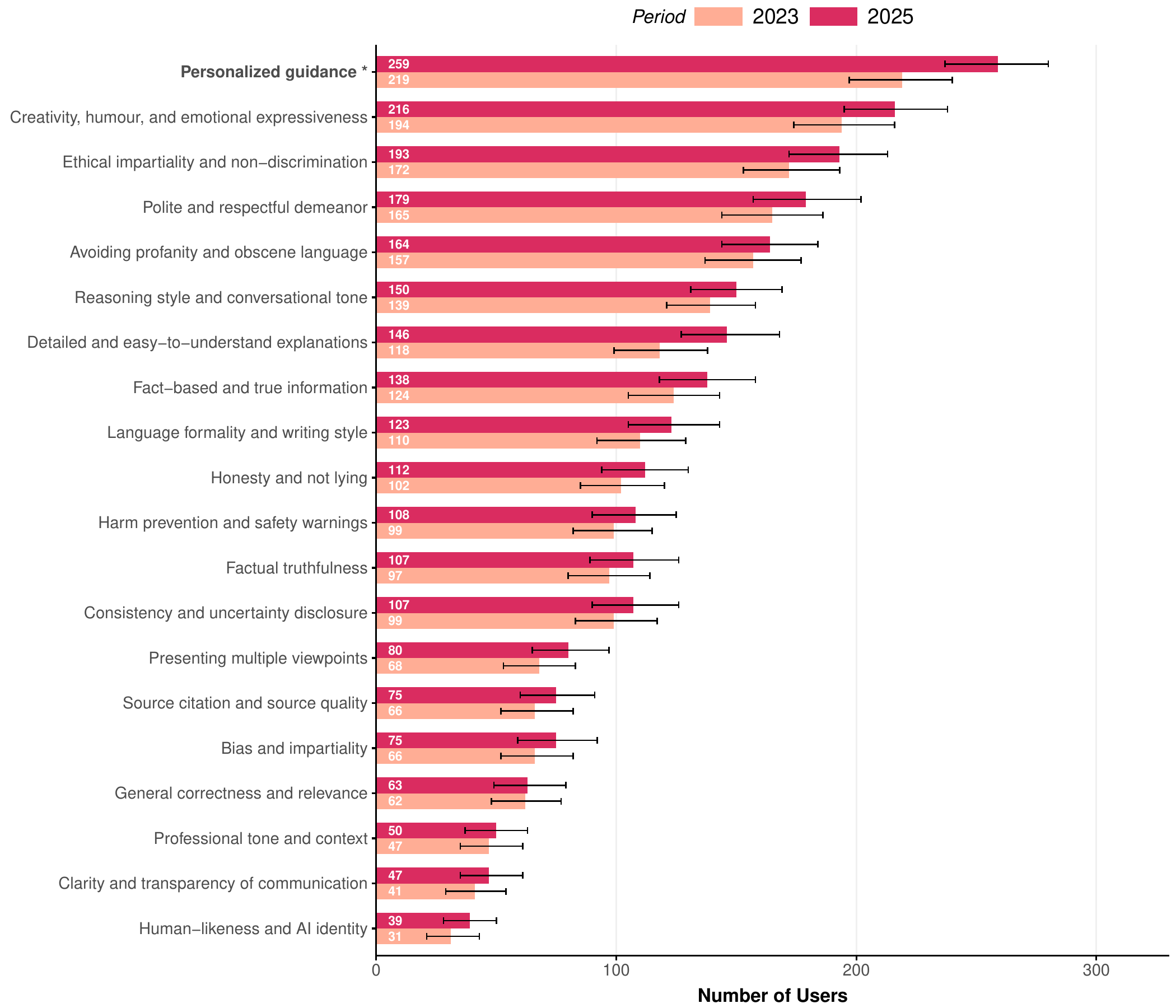}
    \caption{\textbf{Atom clusters by period.} Unique users with $\ge1$ atom in each cluster, with 95\% bootstrap CIs (1,000 iter). $*$~$p<.05$ indicates significant change (proportion test, 2023 vs 2025).}
    \label{fig:system_string_atoms_clusters}
\end{figure}

To obtain a fine-grained vocabulary of what participants care about, we extracted short topic words (``atoms'') from each system string using GPT-5.4 (e.g., ``honest'', ``unbiased'', ``human-like''), preserving the participant's own framing without merging synonyms or standardising vocabulary. We show the top 20 extracted atoms in \cref{fig:system_string_atoms_frequency} by data collection period. Atoms are then embedded with all-MiniLM-L6-v2 \citep{reimersSentenceBERT2019}, projected to 2D with UMAP, and clustered with $k$-means ($k=20$). We show a 2-D projection of the overall space in \cref{fig:system_string_atoms_umap}, and the frequency of clusters by period in \cref{fig:system_string_atoms_clusters}. Descriptions of each cluster are presented in \cref{tab:atom_clusters}.

\footnotesize{
\rowcolors{2}{gray!10}{white}
\renewcommand\arraystretch{1.1}
\setlength\tabcolsep{2mm}
\hyphenpenalty=10000
\exhyphenpenalty=10000
\begin{longtable}{>{\raggedright\arraybackslash}p{3.5cm} >{\raggedright\arraybackslash}p{11cm}}
\caption{Preference atom clusters from system strings with GPT-5.4 labels and descriptions.}
\label{tab:atom_clusters} \\
\hline
\rowcolor{white}
\textbf{Cluster} & \textbf{Description} \\
\hline
\endfirsthead
\multicolumn{2}{c}{\tablename\ \thetable{} -- \textit{Continued from previous page}} \\
\hline
\rowcolor{white}
\textbf{Cluster} & \textbf{Description} \\
\hline
\endhead
\midrule
\multicolumn{2}{r}{\textit{Continued on next page}} \\
\endfoot
\bottomrule
\endlastfoot
\textbf{Personalized guidance} & Preferences for an AI that tailors its help and suggestions to the individual user while being knowledgeable and assistive. This is specifically about adaptive, user-specific guidance and expertise, not general professionalism, friendliness, accuracy, or privacy alone. \\
\textbf{Creativity, humour, and emotional expressiveness} & Preferences in this cluster focus on the AI being imaginative, entertaining, and emotionally resonant—showing empathy, humour, storytelling, and creative insights. This differs from clusters about accuracy, politeness, safety, or general style because it centers on expressive and engaging personality. \\
\textbf{Ethical impartiality and non-discrimination} & Preferences in this cluster focus on the AI being guided by moral principles while remaining objective, impartial, and non-discriminatory. Unlike general bias, truthfulness, or openness-to-views clusters, this one is specifically about fair ethical treatment and avoiding discrimination. \\
\textbf{Polite and respectful demeanor} & Preferences in this cluster focus on the AI being courteous, kind, friendly, and non-rude in how it addresses the user. This is specifically about interpersonal manners and warmth, not professionalism, emotional expressiveness, clarity, or factual reliability. \\
\textbf{Avoiding profanity and obscene language} & This cluster is specifically about not using swear words, vulgarity, abusive wording, or other obscene/offensive language. It is narrower than general politeness or social acceptability because it focuses on explicit verbal profanity rather than overall behavior or tone. \\
\textbf{Reasoning style and conversational tone} & Preferences in this cluster concern how the AI presents itself while reasoning and speaking: understandable, logical, insightful, concise, serious or funny/sarcastic, and non-argumentative. It is distinct from clusters about factual truth, bias, safety, warmth, or transparency because it focuses on the style and demeanor of explanation and interaction. \\
\textbf{Detailed and easy-to-understand explanations} & Preferences in this cluster focus on explanations that are both comprehensive and simple to comprehend. Unlike clusters about clear language alone, accuracy, or transparency, this one specifically emphasizes depth plus understandability of the explanation. \\
\textbf{Fact-based and true information} & Preferences in this cluster focus specifically on the AI providing information grounded in facts and truth, including fact-checking and avoiding false information. Unlike broader accuracy, honesty, or citation clusters, this one is about factuality itself as the basis of responses. \\
\textbf{Language formality and writing style} & Preferences in this cluster are about the stylistic form of the AI’s wording—such as being concise, formal or less formal, elegant, simple, grammatical, and using short natural-sounding sentences. It is distinct from clusters about clarity, professionalism, politeness, or factual accuracy because it focuses specifically on linguistic style and register. \\
\textbf{Honesty and not lying} & Preferences that the AI be candid, sincere, and explicitly avoid deception or lying. This is about truthfulness as a moral/behavioral trait, distinct from factual accuracy, source verification, or transparency/clarity. \\
\textbf{Harm prevention and safety warnings} & Preferences in this cluster focus specifically on avoiding harmful or dangerous content and advice, including being cautious and warning about risks. This is distinct from broader content moderation, politeness, truthfulness, or accuracy because it centers on physical or practical safety. \\
\textbf{Factual truthfulness} & Preferences that the AI's statements be factually true and correct. This is specifically about truth of the content itself, not source citation, level of detail, clarity, or broader honesty about uncertainty. \\
\textbf{Consistency and uncertainty disclosure} & Preferences in this cluster focus on the AI being reliable and precise across responses while clearly admitting when it is uncertain. Unlike general accuracy or truthfulness clusters, this one emphasizes stable results and explicit acknowledgment of limits. \\
\textbf{Presenting multiple viewpoints} & Preferences about the AI including different, both, all, or many perspectives/opinions on a topic. This is specifically about representing a range of viewpoints, not general neutrality, lack of bias, or factual accuracy. \\
\textbf{Source citation and source quality} & Preferences that the AI provide citations or references and draw from multiple, reputable, well-known sources. This is specifically about naming and using sources, not just being accurate, factual, or honest. \\
\textbf{Bias and impartiality} & Statements in this cluster are specifically about the AI being unbiased, non-biased, or free of bias. This is distinct from general accuracy, honesty, transparency, or presenting multiple viewpoints, because the focus is on avoiding partiality or prejudice. \\
\textbf{General correctness and relevance} & Preferences in this cluster focus on the AI giving correct, accurate, and relevant answers overall. Unlike clusters about factuality, evidence, precision, or truthfulness, this one is about broad answer quality being right and on-topic. \\
\textbf{Professional tone and context} & Preferences in this cluster focus on the AI sounding and writing in a professional manner, especially for workplace or expert-related situations. It is about professionalism of style and setting, not general politeness, factuality, or referring to sources. \\
\textbf{Clarity and transparency of communication} & Preferences for the AI to present information in clear, understandable language and to be transparent about what it is saying or doing. This is about comprehensibility and openness, not factual accuracy, honesty, bias, or professionalism. \\
\textbf{Human-likeness and AI identity} & Preferences about whether the AI should seem human versus clearly machine- or tool-like, especially not pretending to be a person. This is specifically about anthropomorphism and identity presentation, not general tone, friendliness, or honesty. \\
\end{longtable}
}
\normalsize

\subsubsection{Preference Axes}
To capture the \emph{directionality} of preferences, we extracted bipolar preference axes from each system string using GPT-5.4. Each axis describes a dimension of variation with two contrasting poles (e.g., ``warm, empathetic, human-like $\leftrightarrow$ cold, robotic, impersonal''). Axes were embedded with all-MiniLM-L6-v2 and clustered with $k$-means ($k=6$). Cluster labels and canonical pole descriptions were assigned by GPT-5.4, which was shown the 10 nearest unique examples to each cluster centroid alongside examples from other clusters. We then aligned polarity within each cluster: GPT-5.4 defined canonical left/right poles, then re-classified each participant's position relative to these poles to account for cases where participants described the same dimension with swapped poles. Representative quotes were extracted by showing GPT-5.4 the original system strings of the 100 participants closest to each cluster centroid and asking it to identify exact phrases illustrating each pole.

\footnotesize{
\rowcolors{2}{gray!10}{white}
\renewcommand\arraystretch{1.1}
\setlength\tabcolsep{2mm}
\hyphenpenalty=10000
\exhyphenpenalty=10000
\begin{longtable}{>{\raggedright\arraybackslash}p{3.5cm} >{\raggedright\arraybackslash}p{11cm}}
\caption{Preference axes identified from system strings, with GPT-5.4 labels and descriptions.}
\label{tab:axes_dimensions} \\
\hline
\rowcolor{white}
\textbf{Dimension} & \textbf{Description} \\
\hline
\endfirsthead
\multicolumn{2}{c}{\tablename\ \thetable{} -- \textit{Continued from previous page}} \\
\hline
\rowcolor{white}
\textbf{Dimension} & \textbf{Description} \\
\hline
\endhead
\midrule
\multicolumn{2}{r}{\textit{Continued on next page}} \\
\endfoot
\bottomrule
\endlastfoot
\textbf{Response brevity and expansiveness} & This dimension captures whether the user prefers answers that are short, direct, and task-focused versus answers that are more elaborate, conversational, and exploratory. It is mainly about how much the assistant should say and how expansively it should engage. \\
\textbf{Factual honesty and reliability} & This dimension captures whether the user wants the AI to prioritize truthfulness, accuracy, and evidence-based reliability versus giving speculative, misleading, or fabricated information. It is mainly about factual trustworthiness rather than tone, balance, or bias. \\
\textbf{Respectfulness and civility} & This dimension captures whether the user wants the assistant to communicate in a polite, courteous, respectful way or is comfortable with a blunt, rude, or dismissive tone. It is mainly about basic civility rather than emotional warmth. \\
\textbf{Neutrality and partisanship} & This dimension captures whether the user wants the AI to remain impartial, balanced, and non-opinionated versus taking sides and expressing a slanted or partisan viewpoint. It is about neutrality in framing and judgment rather than factual accuracy itself. \\
\textbf{Content restriction and safety filtering} & This dimension captures whether the user wants the AI to prioritize safety guardrails and content restrictions or to be as open, uncensored, and unrestricted as possible. It reflects preferences about moderation, refusal boundaries, and willingness to avoid potentially harmful content. \\
\textbf{Human warmth and personability} & This dimension captures whether the user wants the AI to feel warm, empathetic, friendly, and human-like versus emotionally distant and machine-like. It is mainly about interpersonal tone and perceived human presence. \\
\end{longtable}
}
\normalsize
\begin{figure}[H]
    \centering
    \includegraphics[width=\linewidth]{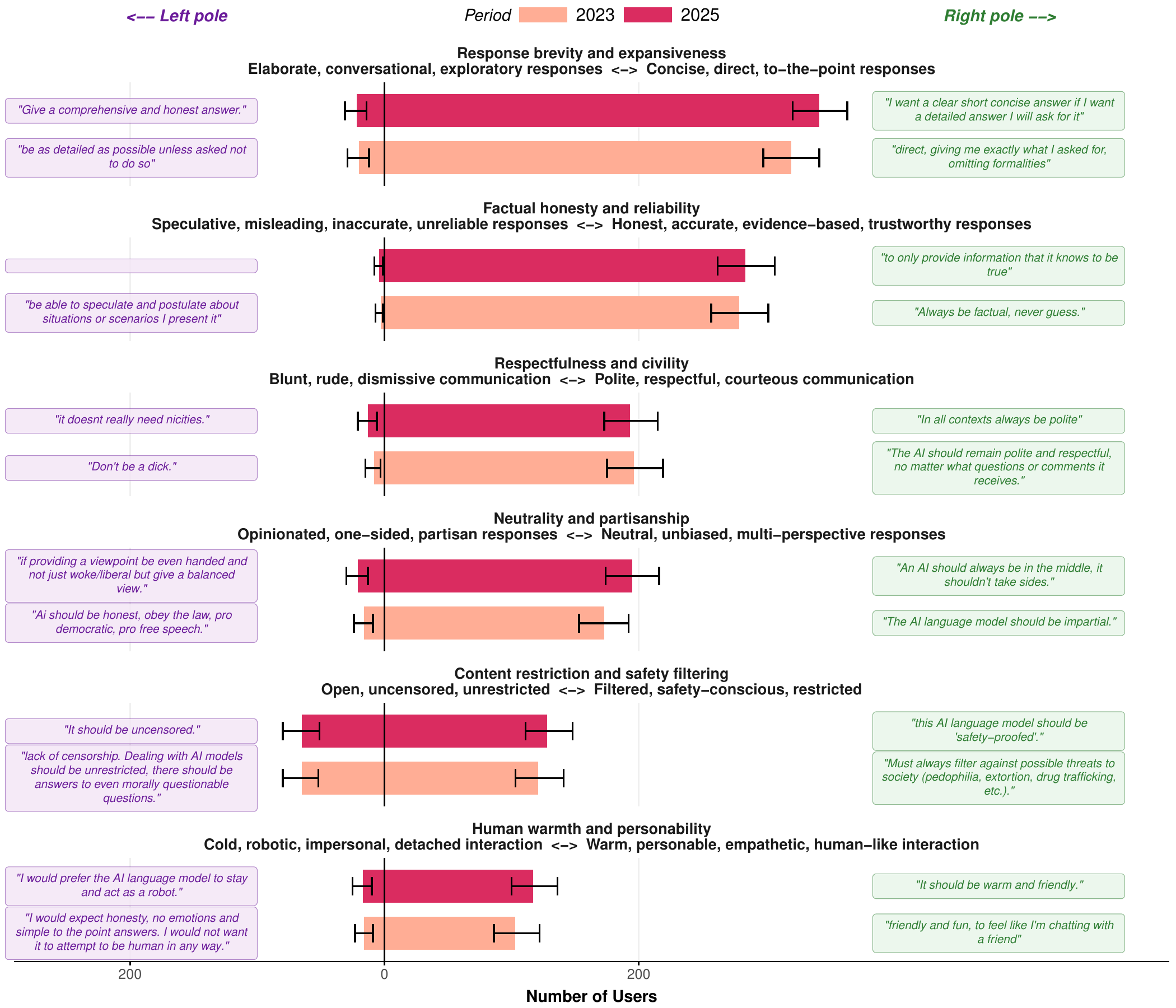}
    \caption{\textbf{Preference axes with users by pole and period.} Butterfly chart showing users on each pole of each dimension, comparing 2023 (orange) and 2025 (blue). Error bars: 95\% bootstrap CIs (1,000 iterations). Participant quotes from their system strings shown on each side.}
    \label{fig:system_string_axes_butterfly}
\end{figure}

\newpage
\section{Survey on Attitudes to Personalisation Methods}
\label{sec:appendix_personalisation_survey}
\normalsize
\subsection{Survey Setup}
In the post-survey, we measured participant attitudes toward six different methods of AI personalisation, three of which were used in our study (stated preferences, ratings of outputs, demographics) and three hypothetical alternatives (full chat history, psychological characteristics, physiological responses). Participants rated each method on three dimensions using Strongly Disagree - Strongly Agree sliders (0-100 scale):
\begin{itemize}
    \item \textbf{Usefulness:} ``I think using [METHOD] would make the AI more useful for me.''
    \item \textbf{Autonomy:} ``I think using [METHOD] still lets me keep control and preserve my autonomy over how the AI personalises responses for me.''
    \item \textbf{Comfort:} ``I would feel comfortable with the AI using [METHOD] to personalise its responses.''
\end{itemize}

\begin{figure}[H]
    \centering
    \includegraphics[width=0.4\linewidth]{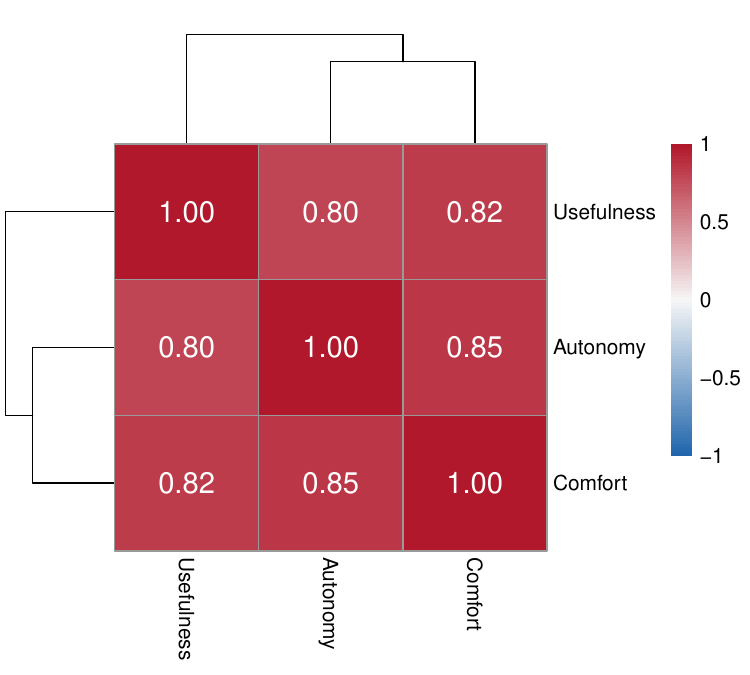}
    \caption{\textbf{Correlations between attitude dimensions.} Pearson correlations across the three rating dimensions, averaged within participants.}
    \label{fig:personalisation_attitude_correlations}
\end{figure}

All three items were positively framed, so higher scores indicate greater acceptability. The three attitude dimensions were moderately to strongly correlated within participants (\cref{fig:personalisation_attitude_correlations}), suggesting a general ``acceptability'' construct, though with meaningful variation across dimensions — particularly for methods where participants perceived usefulness but felt discomfort (e.g., psychological characteristics).

We further classified methods as \emph{active} (requiring explicit user input: stated preferences, ratings of outputs) or \emph{passive} (inferred without explicit action: demographics, full chat history, psychological characteristics, physiological responses).

The six personalisation methods were described to participants as follows:
\begin{itemize}
    \item \textbf{Stated preferences:} The AI model learns from instructions you wrote in plain language (otherwise known as a system string or a system prompt) about how you wanted it to behave.
    \item \textbf{Ratings of outputs:} The AI model learns from how you rated or ranked different answers it gave, like the answers you gave in this study.
    \item \textbf{Demographics:} The AI model receives and adapts to basic demographic information about you, like your age, gender, or education level.
    \item \textbf{Full chat history:} The AI model has access to all of your previous conversations and learns implicitly from how you responded to it.
    \item \textbf{Psychological characteristics:} The AI model learns and adapts to your psychological characteristics like your mood, personality traits, or mental health.
    \item \textbf{Physiological responses:} The AI model tracks and learns from your physical reactions, like your facial expressions or eye movements, using a camera.
\end{itemize}

\begin{figure}[H]
    \centering
    \includegraphics[width=\linewidth]{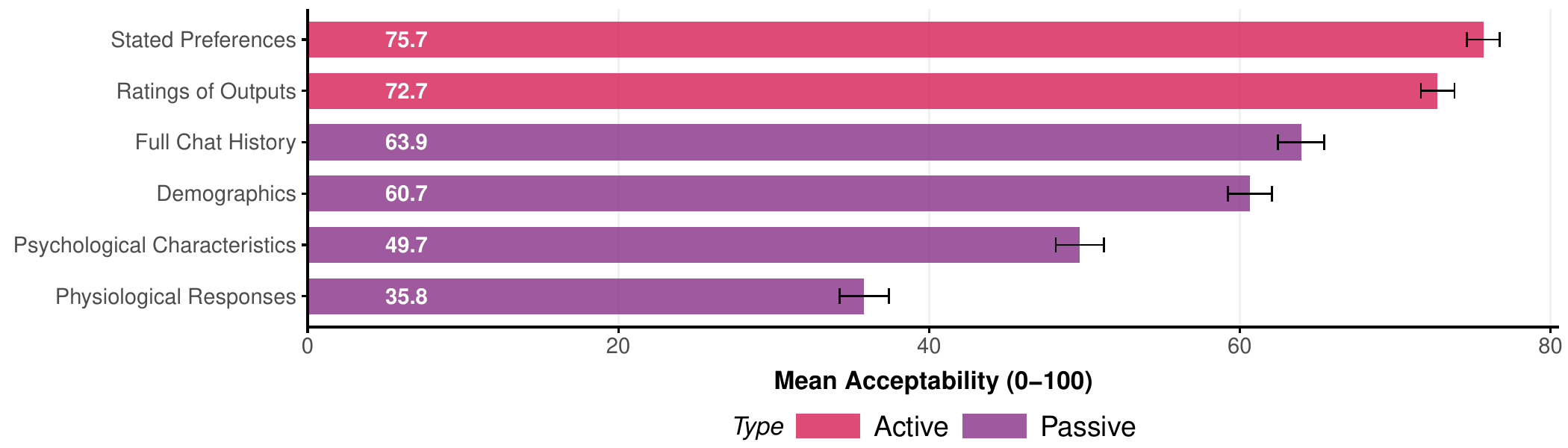}
    \caption{\textbf{Overall acceptability of personalisation methods.} Mean ratings across all three attitude dimensions (0-100 scale) with 95\% CIs (n=530). Methods are sorted by overall mean and coloured by active vs passive.}
    \label{fig:personalisation_overall_acceptability}
\end{figure}

Overall acceptability (averaged across dimensions) is shown in \cref{fig:personalisation_overall_acceptability}.

\subsection{Statistical Analysis}
We used linear mixed-effects models to analyse acceptability ratings. Each participant rated all six personalisation methods on three attitude dimensions (18 observations per participant), so we included random intercepts for participants for repeated measures.

\paragraph{Model Specifications.}
Let $Y_{ijk}$ denote the rating for personalisation method $j$ on attitude dimension $k$ by participant $i$, with $u_i \sim N(0, \tau^2)$ as the participant random intercept. Usefulness is the reference category for attitudes. We analyse both the coarsened active/passive classification and item-level method effects.

\textbf{Active vs Passive (Models [1]--[2]).}
We first collapse the six methods into active (stated preferences, ratings of outputs) vs passive (demographics, chat history, psychological, physiological), with active as the reference:
\begin{equation}
Y_{ijk} = \beta_0 + \beta_1 \cdot \text{Passive}_j + u_i + \varepsilon_{ijk}
\end{equation}
Model [2] adds attitude main effects and passive $\times$ attitude interactions.

\textbf{Item-Level (Models [3]--[4]).}
We then model each method separately, with stated preferences as the reference:
\begin{equation}
Y_{ijk} = \beta_0 + \sum_{m} \beta_m \cdot \mathbf{1}_{j=m} + u_i + \varepsilon_{ijk}
\end{equation}
where $m \in \{\text{Chat History}, \text{Ratings}, \text{Demographics}, \text{Psych.}, \text{Physio.}\}$. Model [4] adds attitude main effects and method $\times$ attitude interactions.

Pooled regression results for all four models are in \cref{tab:personalisation_pooled}. We model each attitude dimension separately to facilitate interpretation within each construct (\cref{tab:personalisation_separate}).

\scriptsize{
\setlength{\LTcapwidth}{\textwidth}
\renewcommand\arraystretch{1.5}
\setlength\tabcolsep{0.8mm}
\begin{longtable}{llcccc}
\caption{\small{Pooled linear mixed models for personalisation acceptability (0-100 scale). Type = active vs passive grouping. Item = individual personalisation methods. Att = attitude dimension interaction (Usefulness = ref). Reference: Active (type) / Stated Preferences (item). Random intercepts for participants. P-values FDR-adjusted.}} \\
\hhline{-----}
\multicolumn{5}{l}{Dependent Variable = \textit{Acceptability (0-100)}} \\
\hhline{=====}
{\bfseries ~} & \multicolumn{1}{c}{\bfseries Type} & \multicolumn{1}{c}{\bfseries Type + Att.} & \multicolumn{1}{c}{\bfseries Item} & \multicolumn{1}{c}{\bfseries Item + Att.} \\

\midrule
\endfirsthead
\multicolumn{5}{c}{\tablename\ \thetable{} -- \textit{Continued from previous page}} \\
\hhline{-----}
\multicolumn{5}{l}{Dependent Variable = \textit{Acceptability (0-100)}} \\
\hhline{=====}
{\bfseries ~} & \multicolumn{1}{c}{\bfseries Type} & \multicolumn{1}{c}{\bfseries Type + Att.} & \multicolumn{1}{c}{\bfseries Item} & \multicolumn{1}{c}{\bfseries Item + Att.} \\

\midrule
\endhead
\midrule
\multicolumn{5}{r}{\textit{Continued on next page}} \\
\multicolumn{5}{r}{\raggedleft{\itshape * p{\textless}0.05~~~** p{\textless}0.01~~~*** p{\textless}0.001}} \\
\endfoot
\bottomrule
\multicolumn{5}{r}{\raggedleft{\itshape * p{\textless}0.05~~~** p{\textless}0.01~~~*** p{\textless}0.001}} \\
\endlastfoot
(Intercept) &
\makecell{\textbf{74.28 \textsuperscript{***}} \\ (72.51~--~76.05)} &
\makecell{\textbf{75.32 \textsuperscript{***}} \\ (73.23~--~77.41)} &
\makecell{\textbf{75.75 \textsuperscript{***}} \\ (73.86~--~77.65)} &
\makecell{\textbf{76.92 \textsuperscript{***}} \\ (74.55~--~79.29)}\\
Passive &
\makecell{\textbf{-21.73 \textsuperscript{***}} \\ (-22.70~--~-20.76)} &
\makecell{\textbf{-17.23 \textsuperscript{***}} \\ (-18.90~--~-15.56)} &~ &~\\
Autonomy &~ &
\makecell{\textbf{-3.72 \textsuperscript{***}} \\ (-5.64~--~-1.79)} &~ &
\makecell{\textbf{-3.28 \textsuperscript{*}} \\ (-5.75~--~-0.80)}\\
Comfort &~ &
\makecell{0.59 \\ (-1.34~--~2.51)} &~ &
\makecell{-0.22 \\ (-2.70~--~2.26)}\\
Passive : Autonomy &~ &
\makecell{\textbf{-5.17 \textsuperscript{***}} \\ (-7.53~--~-2.82)} &~ &~\\
Passive : Comfort &~ &
\makecell{\textbf{-8.32 \textsuperscript{***}} \\ (-10.68~--~-5.97)} &~ &~\\
Chat Hist &~ &~ &
\makecell{\textbf{-11.80 \textsuperscript{***}} \\ (-13.26~--~-10.35)} &
\makecell{\textbf{-5.66 \textsuperscript{***}} \\ (-8.13~--~-3.18)}\\
Ratings &~ &~ &
\makecell{\textbf{-2.94 \textsuperscript{***}} \\ (-4.39~--~-1.49)} &
\makecell{\textbf{-3.18 \textsuperscript{*}} \\ (-5.66~--~-0.70)}\\
\end{longtable}
}
\normalsize
\scriptsize{
\setlength{\LTcapwidth}{\textwidth}
\renewcommand\arraystretch{1.5}
\setlength\tabcolsep{0.8mm}
\begin{longtable}{llcccccc}
\caption{\small{Separate linear mixed models for personalisation acceptability by attitude dimension (0-100 scale). Type: active vs passive (ref: Active). Item: individual personalisation methods (ref: Stated Preferences). Use. = Usefulness, Aut. = Autonomy, Com. = Comfort. Random intercepts for participants. P-values FDR-adjusted.}} \\
\hhline{-------}
\multicolumn{7}{l}{Dependent Variable = \textit{Acceptability (0-100)}} \\
\hhline{=======}
{\bfseries ~} & \multicolumn{1}{c}{\bfseries Type: Use.} & \multicolumn{1}{c}{\bfseries Type: Aut.} & \multicolumn{1}{c}{\bfseries Type: Com.} & \multicolumn{1}{c}{\bfseries Item: Use.} & \multicolumn{1}{c}{\bfseries Item: Aut.} & \multicolumn{1}{c}{\bfseries Item: Com.} \\

\midrule
\endfirsthead
\multicolumn{7}{c}{\tablename\ \thetable{} -- \textit{Continued from previous page}} \\
\hhline{-------}
\multicolumn{7}{l}{Dependent Variable = \textit{Acceptability (0-100)}} \\
\hhline{=======}
{\bfseries ~} & \multicolumn{1}{c}{\bfseries Type: Use.} & \multicolumn{1}{c}{\bfseries Type: Aut.} & \multicolumn{1}{c}{\bfseries Type: Com.} & \multicolumn{1}{c}{\bfseries Item: Use.} & \multicolumn{1}{c}{\bfseries Item: Aut.} & \multicolumn{1}{c}{\bfseries Item: Com.} \\

\midrule
\endhead
\midrule
\multicolumn{7}{r}{\textit{Continued on next page}} \\
\multicolumn{7}{r}{\raggedleft{\itshape * p{\textless}0.05~~~** p{\textless}0.01~~~*** p{\textless}0.001}} \\
\endfoot
\bottomrule
\multicolumn{7}{r}{\raggedleft{\itshape * p{\textless}0.05~~~** p{\textless}0.01~~~*** p{\textless}0.001}} \\
\endlastfoot
(Intercept) &
\makecell{\textbf{75.30 \textsuperscript{***}} \\ (73.29~--~77.30)} &
\makecell{\textbf{71.59 \textsuperscript{***}} \\ (69.57~--~73.61)} &
\makecell{\textbf{75.89 \textsuperscript{***}} \\ (73.70~--~78.09)} &
\makecell{\textbf{76.89 \textsuperscript{***}} \\ (74.59~--~79.20)} &
\makecell{\textbf{73.63 \textsuperscript{***}} \\ (71.32~--~75.95)} &
\makecell{\textbf{76.69 \textsuperscript{***}} \\ (74.21~--~79.17)}\\
Passive &
\makecell{\textbf{-17.21 \textsuperscript{***}} \\ (-18.89~--~-15.53)} &
\makecell{\textbf{-22.40 \textsuperscript{***}} \\ (-24.02~--~-20.77)} &
\makecell{\textbf{-25.54 \textsuperscript{***}} \\ (-27.30~--~-23.78)} &~ &~ &~\\
Chat Hist &~ &~ &~ &
\makecell{\textbf{-5.63 \textsuperscript{***}} \\ (-8.10~--~-3.16)} &
\makecell{\textbf{-15.33 \textsuperscript{***}} \\ (-17.75~--~-12.91)} &
\makecell{\textbf{-14.41 \textsuperscript{***}} \\ (-16.97~--~-11.85)}\\
Ratings &~ &~ &~ &
\makecell{\textbf{-3.18 \textsuperscript{*}} \\ (-5.65~--~-0.71)} &
\makecell{\textbf{-4.06 \textsuperscript{***}} \\ (-6.48~--~-1.65)} &
\makecell{-1.57 \\ (-4.13~--~0.99)}\\
Demo &~ &~ &~ &
\makecell{\textbf{-12.96 \textsuperscript{***}} \\ (-15.43~--~-10.49)} &
\makecell{\textbf{-15.61 \textsuperscript{***}} \\ (-18.03~--~-13.20)} &
\makecell{\textbf{-16.50 \textsuperscript{***}} \\ (-19.06~--~-13.94)}\\
Psych &~ &~ &~ &
\makecell{\textbf{-21.21 \textsuperscript{***}} \\ (-23.67~--~-18.74)} &
\makecell{\textbf{-27.06 \textsuperscript{***}} \\ (-29.48~--~-24.65)} &
\makecell{\textbf{-29.80 \textsuperscript{***}} \\ (-32.36~--~-27.24)}\\
Physio &~ &~ &~ &
\makecell{\textbf{-35.43 \textsuperscript{***}} \\ (-37.90~--~-32.96)} &
\makecell{\textbf{-39.71 \textsuperscript{***}} \\ (-42.13~--~-37.30)} &
\makecell{\textbf{-44.62 \textsuperscript{***}} \\ (-47.18~--~-42.06)}\\
 \hline
\multicolumn{7}{l}{{\bfseries Random Effects}} \\
\end{longtable}
}
\normalsize

\section{Human Preferences}
\label{sec:appendix_human_tasks}

\subsection{GPU Error Analysis}
\label{sec:appendix_gpu_errors}
\normalsize

During some multi-turn conversations, participants saw two types of error messages, instead of a generated response:
\begin{enumerate}
    \item \textbf{GPU errors} = [``No GPU is available right now after multiple retries''; ``Sorry, there has been an error generating the response'']
    \item \textbf{Context-window errors} = ``Maximum conversation limit reached''
\end{enumerate}

\begin{figure}[H]
    \centering
    \includegraphics[width=\linewidth]{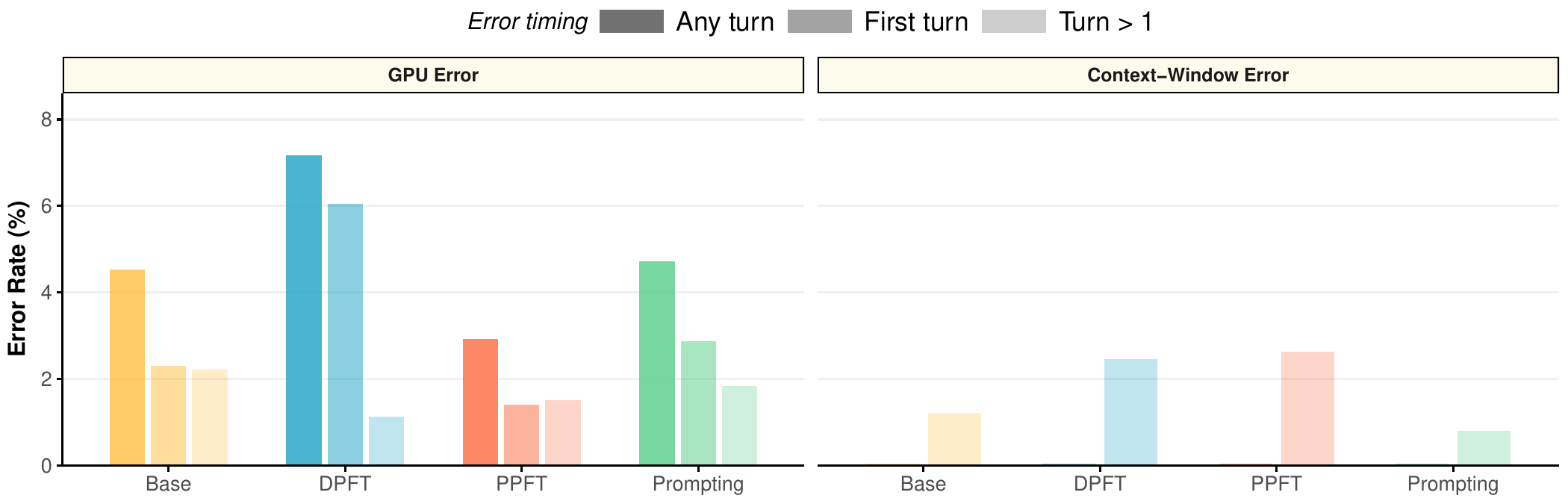}
    \caption{\textbf{Generation error rates by model.} Left panel: GPU errors (capacity and generation failures), split by any-turn, first-turn, and subsequent-turn errors. Right panel: context-window errors (``Maximum conversation limit reached''), split by first-turn and subsequent-turn.}
    \label{fig:gpu_error_rate_by_model}
\end{figure}

\begin{figure}[H]
    \centering
    \includegraphics[width=\linewidth]{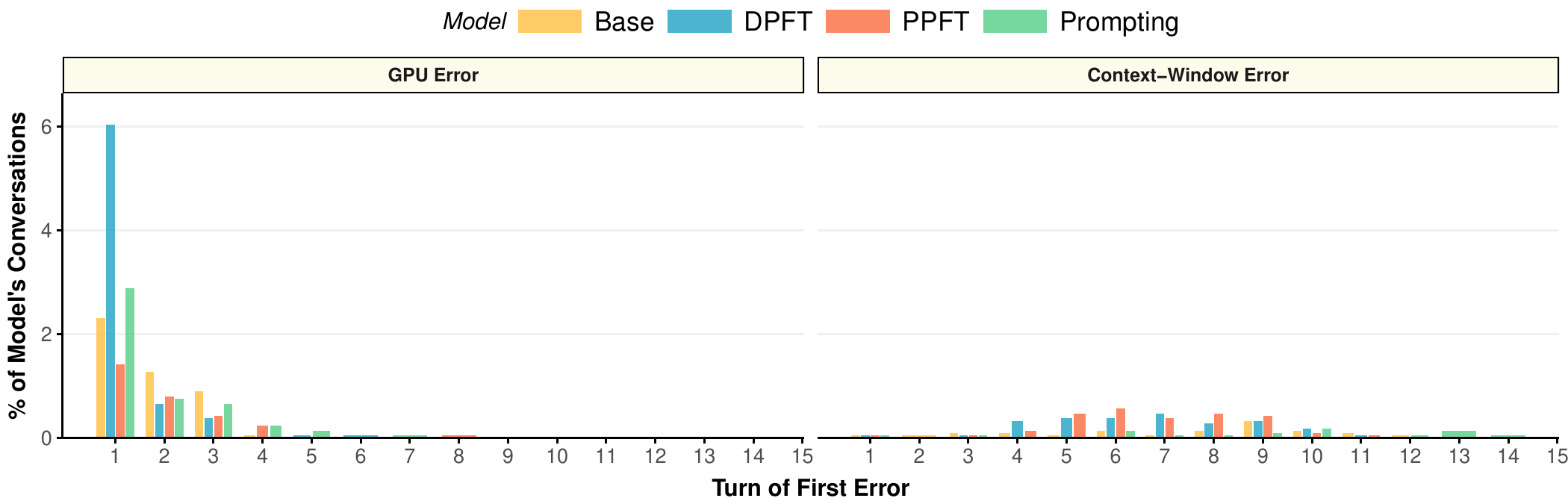}
    \caption{\textbf{Turn-level error distribution.} Distribution of which assistant turn the first error appears on, by model. Left panel: GPU errors. Right panel: context-window errors.}
    \label{fig:gpu_error_turn_distribution_by_model}
\end{figure}

These two error types are tracked as separate fields in the processed data (\texttt{gpu\_error} and \texttt{max\_turns\_reached}). For descriptive characterisation (\cref{fig:gpu_error_rate_by_model,fig:gpu_error_turn_distribution_by_model}), we report the error types separately. For regression controls, they are combined, since both produce the same participant experience that a model could not generate a response on the affected turn. Context-window errors are rare relative to GPU errors and almost exclusively appear in later turns. The diverse preference fine-tuned model (DPFT) incurs more GPU errors than other models.

\begin{figure}[H]
    \centering
    \includegraphics[width=\linewidth]{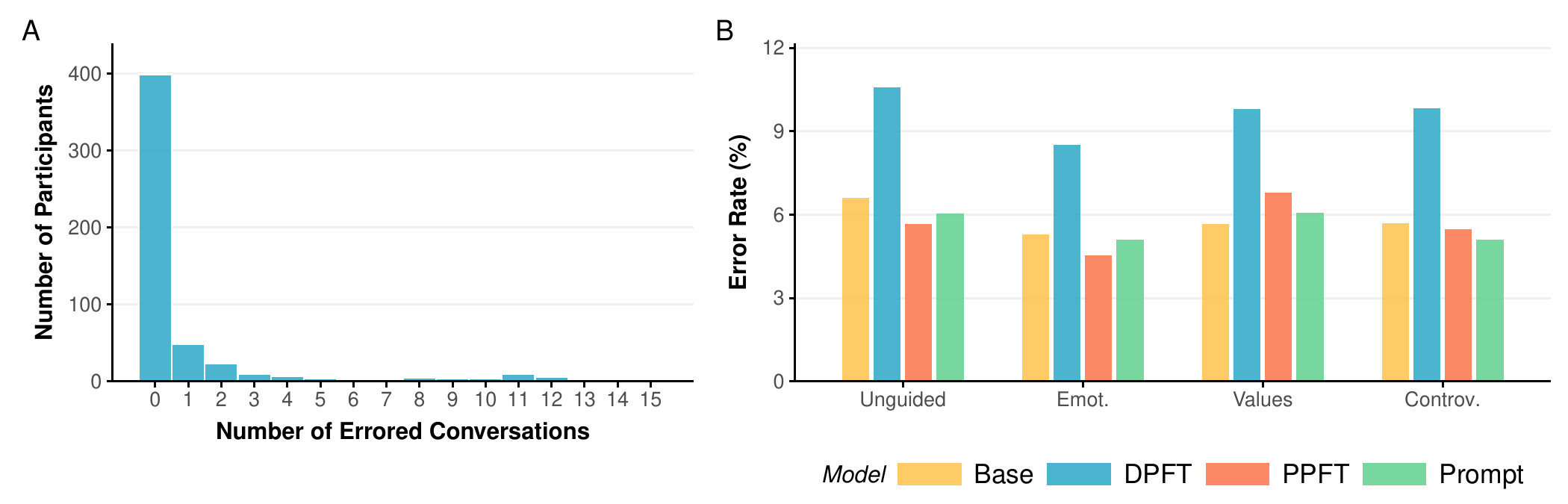}
    \caption{\textbf{Error concentration.} (A)~Distribution of error count per participant. (B)~Error rate by domain and model.}
    \label{fig:gpu_error_concentration}
\end{figure}

\subsubsection{Error Penalty and Handling Strategy}

\begin{figure}[H]
    \centering
    \includegraphics[width=\linewidth]{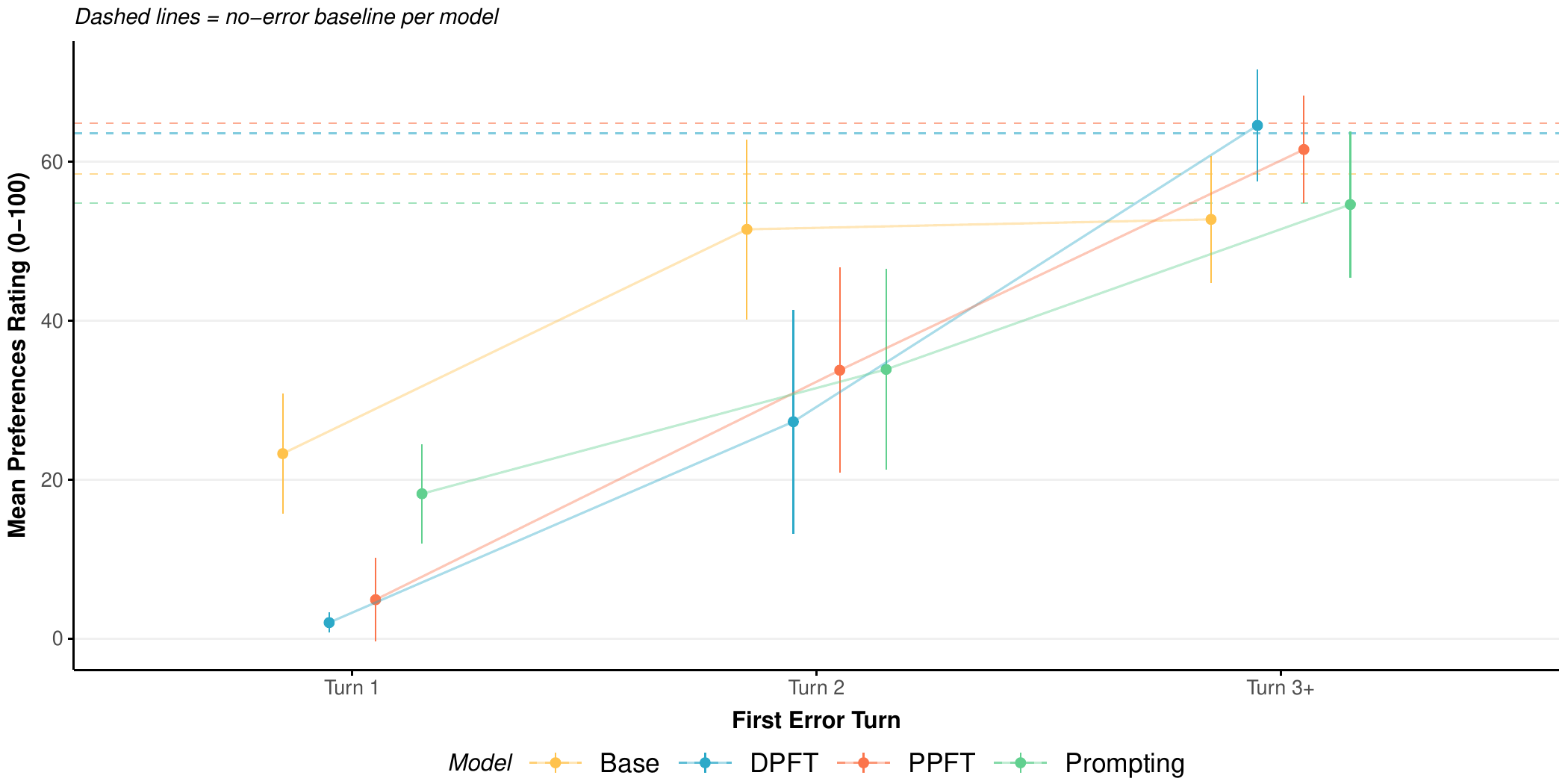}
    \caption{\textbf{Per-turn error penalty on preference ratings.} Mean rating by model and turn of first error. Error bars show 95\% CI. Dashed horizontal lines are models' no-error baseline.}
    \label{fig:gpu_error_penalty_by_turn_and_model}
\end{figure}

We compute mean preference ratings by model and the turn where the first error occurred, combining Turns$\ge3$ due to data sparsity. \Cref{fig:gpu_error_penalty_by_turn_and_model} shows that turn-1 errors are severely penalised, turn-2 errors produce a smaller but still substantial penalty, and errors on turn~$\ge3$ have a negligible effect (near no-error baseline). This motivates a binary split control that distinguishes between first-turn errors (complete failure) and subsequent errors (partial degradation). Simple penalty regressions (error covariates only, no model or domain controls) are in \cref{tab:gpu_error_penalty_simple}.

\scriptsize{
\setlength{\LTcapwidth}{\textwidth}
\renewcommand\arraystretch{1.5}
\setlength\tabcolsep{0.8mm}
\begin{longtable}{llccc}
\caption{\small{Simple error penalty models (no model or domain controls). Rating: lmer with participant random intercept. Opening choice and ranking: conditional logit stratified by choice set.}} \\
\hhline{----}
\multicolumn{4}{l}{Dependent Variable = \textit{Preference (Cardinal [1], Ordinal [2], [3])}} \\
\hhline{====}
{\bfseries ~} & \multicolumn{1}{c}{\bfseries Rating} & \multicolumn{1}{c}{\bfseries Opening Choice} & \multicolumn{1}{c}{\bfseries Ranking} \\

\midrule
\endfirsthead
\multicolumn{4}{c}{\tablename\ \thetable{} -- \textit{Continued from previous page}} \\
\hhline{----}
\multicolumn{4}{l}{Dependent Variable = \textit{Preference (Cardinal [1], Ordinal [2], [3])}} \\
\hhline{====}
{\bfseries ~} & \multicolumn{1}{c}{\bfseries Rating} & \multicolumn{1}{c}{\bfseries Opening Choice} & \multicolumn{1}{c}{\bfseries Ranking} \\

\midrule
\endhead
\midrule
\multicolumn{4}{r}{\textit{Continued on next page}} \\
\multicolumn{4}{r}{\raggedleft{\itshape * p{\textless}0.05~~~** p{\textless}0.01~~~*** p{\textless}0.001}} \\
\endfoot
\bottomrule
\multicolumn{4}{r}{\raggedleft{\itshape * p{\textless}0.05~~~** p{\textless}0.01~~~*** p{\textless}0.001}} \\
\endlastfoot
Intercept &
\makecell{\textbf{60.59 \textsuperscript{***}} \\ (59.46~--~61.72)} &~ &~\\
First-turn Error &
\makecell{\textbf{-55.46 \textsuperscript{***}} \\ (-58.96~--~-51.96)} &
\makecell{\textbf{0.01 \textsuperscript{***}} \\ (0.00~--~0.05)} &
\makecell{\textbf{0.05 \textsuperscript{***}} \\ (0.02~--~0.10)}\\
Subsequent Error &
\makecell{\textbf{-7.65 \textsuperscript{***}} \\ (-10.62~--~-4.68)} &~ &
\makecell{1.23 \\ (0.90~--~1.68)}\\
 \hline
\multicolumn{4}{l}{{\bfseries Random Effects}} \\ \hline
$\sigma $\textsuperscript{2} &
492.66 &
~ &
~\\
$\tau $\textsubscript{00} &
142.78 \textsubscript{ppt\_id} &
~ &
~\\
ICC &
0.22 &
~ &
~\\
N &
530 \textsubscript{ppt\_id} &
~ &
~\\
\end{longtable}
}
\normalsize

\noindent We thus control for errors in all subsequent analyses using two complementary approaches:

\paragraph{Trial-level filtering.} Trials where \emph{all four models} produced first-turn errors are excluded entirely, as participants had no real model output to evaluate ($N=5$, 0.24\%).

\paragraph{Error covariates (split control).} For remaining trials, let $E^{(1)}_{ijk}$ denote whether model $j$ in trial $k$ for participant $i$ had a first-turn error, and $E^{(2+)}_{ijk}$ denote whether it had a subsequent error. For ratings (linear mixed-effects model):
\begin{equation}
    Y_{ijk} = \beta_0 + \sum_{m} \beta_m \cdot \mathbf{1}_{j=m} + \gamma_1 E^{(1)}_{ijk} + \gamma_2 E^{(2+)}_{ijk} + \sum_{d} \delta_d \cdot \mathbf{1}_{\text{domain}=d} + u_i + \varepsilon_{ijk}
\end{equation}
where $m \in \{\text{DPFT}, \text{PPFT}, \text{Prompting}\}$ (Base is the ref.), $d$ indexes domain (Unguided is the ref.), and $u_i \sim N(0, \tau^2)$ is the participant random intercept. Domain does not appear in the conditional logit models because it is constant within each choice set and is absorbed by the stratification on $\mathcal{C}_k$. For opening choices (conditional logistic regression, evaluated after seeing only the first turn):
\begin{equation}
    \Pr(Y_{ijk} = 1 \mid \mathcal{C}_k) = \frac{\exp\!\left(\sum_{m} \beta_m \cdot \mathbf{1}_{j=m} + \gamma_1 E^{(1)}_{ijk}\right)}{\sum_{j' \in \mathcal{C}_k} \exp\!\left(\sum_{m} \beta_m \cdot \mathbf{1}_{j'=m} + \gamma_1 E^{(1)}_{j'k}\right)}
\end{equation}
where $\mathcal{C}_k$ is the choice set for trial $k$ and only $E^{(1)}$ is included since the choice is made before subsequent turns. For rankings (conditional logistic regression, evaluated after the full conversation):
\begin{equation}
    \Pr(Y_{ijk} = 1 \mid \mathcal{C}_k) = \frac{\exp\!\left(\sum_{m} \beta_m \cdot \mathbf{1}_{j=m} + \gamma_1 E^{(1)}_{ijk} + \gamma_2 E^{(2+)}_{ijk}\right)}{\sum_{j' \in \mathcal{C}_k} \exp\!\left(\sum_{m} \beta_m \cdot \mathbf{1}_{j'=m} + \gamma_1 E^{(1)}_{j'k} + \gamma_2 E^{(2+)}_{j'k}\right)}
\end{equation}

\subsubsection{Robustness Check}
\begin{figure}[H]
    \centering
    \includegraphics[width=\linewidth]{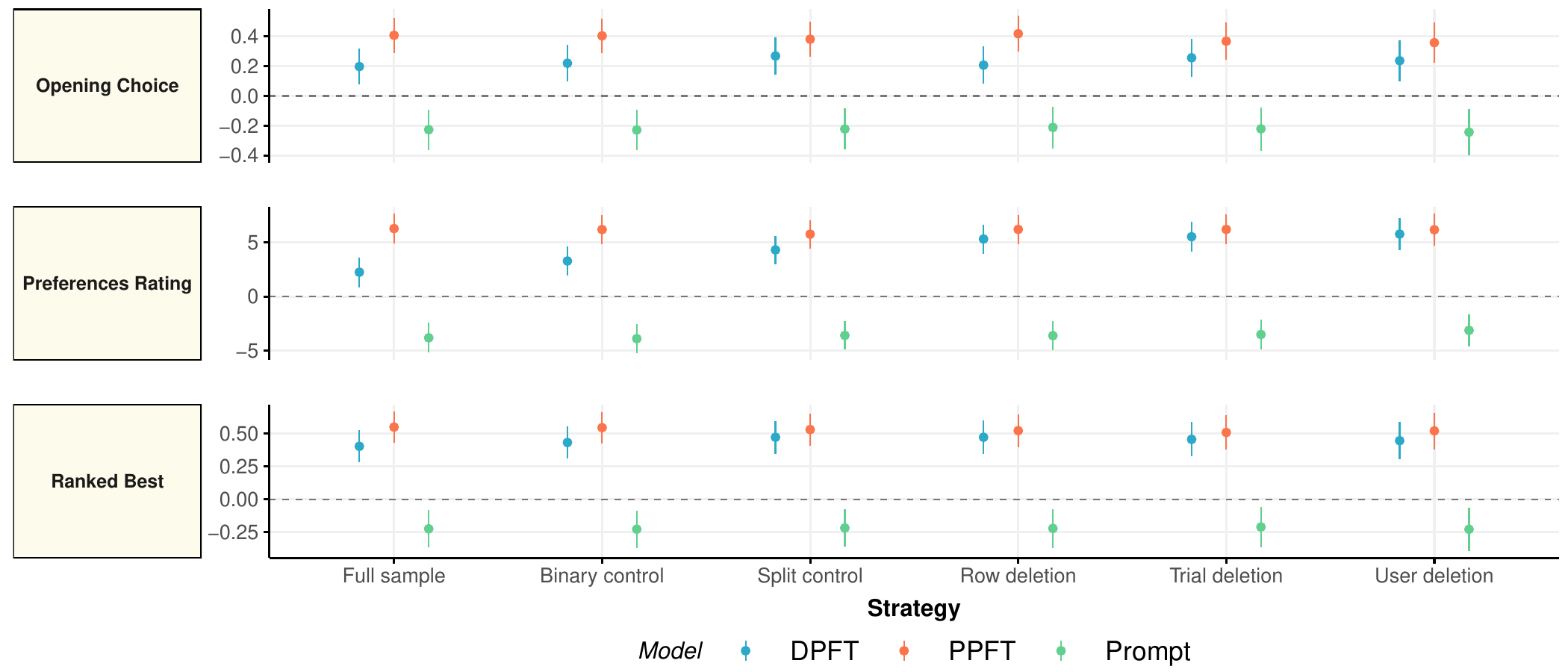}
    \caption{\textbf{Error handling strategy comparison.} Model coefficients (vs.\ Base) under six error-handling strategies, faceted by outcome type. Error bars show 95\% confidence intervals.}
    \label{fig:gpu_error_strategy_all_coefs}
\end{figure}

To investigate robustness of findings to the error-handling approach, we compare six strategies: (1)~full sample with no error controls, (2)~binary control ($E^{\text{any}}_{ijk}$ as a single covariate), (3)~split control ($E^{(1)}_{ijk} + E^{(2+)}_{ijk}$), (4)~row deletion (remove error observations), (5)~trial deletion (remove entire trials containing errors), and (6)~user deletion (remove all data from error-affected participants). We fit the judgement model with each strategy for opening choices, rankings (both conditional logistic regression), and ratings (linear mixed-effects regression). The main findings are robust (\cref{fig:gpu_error_strategy_all_coefs}). AIC/BIC favour the split control among strategies that retain the full sample. Regression results are in \cref{tab:gpu_error_robustness_choices,tab:gpu_error_robustness_rankings,tab:gpu_error_robustness_ratings}.

\scriptsize{
\setlength{\LTcapwidth}{\textwidth}
\renewcommand\arraystretch{1.5}
\setlength\tabcolsep{0.8mm}
\begin{longtable}{llcccccc}
\caption{\small{Robustness of opening choices across error-handling strategies (conditional logit).}} \\
\hhline{-------}
\multicolumn{7}{l}{Dependent Variable = \textit{Preference Opening Choice (Binary)}} \\
\hhline{=======}
{\bfseries ~} & \multicolumn{1}{c}{\bfseries Full sample} & \multicolumn{1}{c}{\bfseries Binary control} & \multicolumn{1}{c}{\bfseries Split control} & \multicolumn{1}{c}{\bfseries Row deletion} & \multicolumn{1}{c}{\bfseries Trial deletion} & \multicolumn{1}{c}{\bfseries User deletion} \\

\midrule
\endfirsthead
\multicolumn{7}{c}{\tablename\ \thetable{} -- \textit{Continued from previous page}} \\
\hhline{-------}
\multicolumn{7}{l}{Dependent Variable = \textit{Preference Opening Choice (Binary)}} \\
\hhline{=======}
{\bfseries ~} & \multicolumn{1}{c}{\bfseries Full sample} & \multicolumn{1}{c}{\bfseries Binary control} & \multicolumn{1}{c}{\bfseries Split control} & \multicolumn{1}{c}{\bfseries Row deletion} & \multicolumn{1}{c}{\bfseries Trial deletion} & \multicolumn{1}{c}{\bfseries User deletion} \\

\midrule
\endhead
\midrule
\multicolumn{7}{r}{\textit{Continued on next page}} \\
\multicolumn{7}{r}{\raggedleft{\itshape * p{\textless}0.05~~~** p{\textless}0.01~~~*** p{\textless}0.001}} \\
\endfoot
\bottomrule
\multicolumn{7}{r}{\raggedleft{\itshape * p{\textless}0.05~~~** p{\textless}0.01~~~*** p{\textless}0.001}} \\
\endlastfoot
DPFT &
\makecell{\textbf{1.22 \textsuperscript{**}} \\ (1.08~--~1.38)} &
\makecell{\textbf{1.25 \textsuperscript{**}} \\ (1.10~--~1.41)} &
\makecell{\textbf{1.31 \textsuperscript{***}} \\ (1.16~--~1.48)} &
\makecell{\textbf{1.23 \textsuperscript{**}} \\ (1.08~--~1.39)} &
\makecell{\textbf{1.29 \textsuperscript{***}} \\ (1.14~--~1.47)} &
\makecell{\textbf{1.27 \textsuperscript{**}} \\ (1.10~--~1.46)}\\
PPFT &
\makecell{\textbf{1.50 \textsuperscript{***}} \\ (1.34~--~1.69)} &
\makecell{\textbf{1.50 \textsuperscript{***}} \\ (1.33~--~1.68)} &
\makecell{\textbf{1.46 \textsuperscript{***}} \\ (1.30~--~1.65)} &
\makecell{\textbf{1.52 \textsuperscript{***}} \\ (1.35~--~1.71)} &
\makecell{\textbf{1.44 \textsuperscript{***}} \\ (1.27~--~1.64)} &
\makecell{\textbf{1.43 \textsuperscript{***}} \\ (1.25~--~1.64)}\\
Prompt &
\makecell{\textbf{0.80 \textsuperscript{**}} \\ (0.70~--~0.91)} &
\makecell{\textbf{0.80 \textsuperscript{**}} \\ (0.69~--~0.91)} &
\makecell{\textbf{0.80 \textsuperscript{**}} \\ (0.70~--~0.92)} &
\makecell{\textbf{0.81 \textsuperscript{**}} \\ (0.70~--~0.93)} &
\makecell{\textbf{0.80 \textsuperscript{**}} \\ (0.69~--~0.93)} &
\makecell{\textbf{0.78 \textsuperscript{**}} \\ (0.67~--~0.92)}\\
Has Error &~ &
\makecell{\textbf{0.55 \textsuperscript{**}} \\ (0.42~--~0.73)} &~ &~ &~ &~\\
First-turn Error &~ &~ &
\makecell{\textbf{0.01 \textsuperscript{***}} \\ (0.00~--~0.05)} &~ &~ &~\\ \hline
Observations &
8456 &
8456 &
8456 &
8020 &
7404 &
6380\\
\end{longtable}
}
\normalsize
\scriptsize{
\setlength{\LTcapwidth}{\textwidth}
\renewcommand\arraystretch{1.5}
\setlength\tabcolsep{0.8mm}
\begin{longtable}{llcccccc}
\caption{\small{Robustness of rankings across error-handling strategies (conditional logit).}} \\
\hhline{-------}
\multicolumn{7}{l}{Dependent Variable = \textit{Preference Ranked Best (Binary)}} \\
\hhline{=======}
{\bfseries ~} & \multicolumn{1}{c}{\bfseries Full sample} & \multicolumn{1}{c}{\bfseries Binary control} & \multicolumn{1}{c}{\bfseries Split control} & \multicolumn{1}{c}{\bfseries Row deletion} & \multicolumn{1}{c}{\bfseries Trial deletion} & \multicolumn{1}{c}{\bfseries User deletion} \\

\midrule
\endfirsthead
\multicolumn{7}{c}{\tablename\ \thetable{} -- \textit{Continued from previous page}} \\
\hhline{-------}
\multicolumn{7}{l}{Dependent Variable = \textit{Preference Ranked Best (Binary)}} \\
\hhline{=======}
{\bfseries ~} & \multicolumn{1}{c}{\bfseries Full sample} & \multicolumn{1}{c}{\bfseries Binary control} & \multicolumn{1}{c}{\bfseries Split control} & \multicolumn{1}{c}{\bfseries Row deletion} & \multicolumn{1}{c}{\bfseries Trial deletion} & \multicolumn{1}{c}{\bfseries User deletion} \\

\midrule
\endhead
\midrule
\multicolumn{7}{r}{\textit{Continued on next page}} \\
\multicolumn{7}{r}{\raggedleft{\itshape * p{\textless}0.05~~~** p{\textless}0.01~~~*** p{\textless}0.001}} \\
\endfoot
\bottomrule
\multicolumn{7}{r}{\raggedleft{\itshape * p{\textless}0.05~~~** p{\textless}0.01~~~*** p{\textless}0.001}} \\
\endlastfoot
DPFT &
\makecell{\textbf{1.50 \textsuperscript{***}} \\ (1.32~--~1.69)} &
\makecell{\textbf{1.54 \textsuperscript{***}} \\ (1.36~--~1.75)} &
\makecell{\textbf{1.60 \textsuperscript{***}} \\ (1.42~--~1.82)} &
\makecell{\textbf{1.60 \textsuperscript{***}} \\ (1.41~--~1.82)} &
\makecell{\textbf{1.58 \textsuperscript{***}} \\ (1.39~--~1.80)} &
\makecell{\textbf{1.56 \textsuperscript{***}} \\ (1.36~--~1.80)}\\
PPFT &
\makecell{\textbf{1.73 \textsuperscript{***}} \\ (1.54~--~1.95)} &
\makecell{\textbf{1.73 \textsuperscript{***}} \\ (1.53~--~1.95)} &
\makecell{\textbf{1.70 \textsuperscript{***}} \\ (1.51~--~1.92)} &
\makecell{\textbf{1.69 \textsuperscript{***}} \\ (1.49~--~1.91)} &
\makecell{\textbf{1.66 \textsuperscript{***}} \\ (1.46~--~1.89)} &
\makecell{\textbf{1.68 \textsuperscript{***}} \\ (1.46~--~1.94)}\\
Prompt &
\makecell{\textbf{0.80 \textsuperscript{**}} \\ (0.69~--~0.92)} &
\makecell{\textbf{0.80 \textsuperscript{**}} \\ (0.69~--~0.92)} &
\makecell{\textbf{0.80 \textsuperscript{**}} \\ (0.70~--~0.93)} &
\makecell{\textbf{0.80 \textsuperscript{**}} \\ (0.69~--~0.93)} &
\makecell{\textbf{0.81 \textsuperscript{*}} \\ (0.70~--~0.94)} &
\makecell{\textbf{0.80 \textsuperscript{**}} \\ (0.67~--~0.94)}\\
Has Error &~ &
\makecell{\textbf{0.47 \textsuperscript{**}} \\ (0.35~--~0.62)} &~ &~ &~ &~\\
First-turn Error &~ &~ &
\makecell{\textbf{0.05 \textsuperscript{***}} \\ (0.02~--~0.10)} &~ &~ &~\\
Subsequent Error &~ &~ &
\makecell{1.20 \\ (0.88~--~1.64)} &~ &~ &~\\ \hline
Observations &
8455 &
8455 &
8455 &
7654 &
7403 &
6379\\
\end{longtable}
}
\normalsize
\scriptsize{
\setlength{\LTcapwidth}{\textwidth}
\renewcommand\arraystretch{1.5}
\setlength\tabcolsep{0.8mm}
\begin{longtable}{llcccccc}
\caption{\small{Robustness of ratings across error-handling strategies (linear mixed-effects).}} \\
\hhline{-------}
\multicolumn{7}{l}{Dependent Variable = \textit{Preference Rating (0--100)}} \\
\hhline{=======}
{\bfseries ~} & \multicolumn{1}{c}{\bfseries Full sample} & \multicolumn{1}{c}{\bfseries Binary control} & \multicolumn{1}{c}{\bfseries Split control} & \multicolumn{1}{c}{\bfseries Row deletion} & \multicolumn{1}{c}{\bfseries Trial deletion} & \multicolumn{1}{c}{\bfseries User deletion} \\

\midrule
\endfirsthead
\multicolumn{7}{c}{\tablename\ \thetable{} -- \textit{Continued from previous page}} \\
\hhline{-------}
\multicolumn{7}{l}{Dependent Variable = \textit{Preference Rating (0--100)}} \\
\hhline{=======}
{\bfseries ~} & \multicolumn{1}{c}{\bfseries Full sample} & \multicolumn{1}{c}{\bfseries Binary control} & \multicolumn{1}{c}{\bfseries Split control} & \multicolumn{1}{c}{\bfseries Row deletion} & \multicolumn{1}{c}{\bfseries Trial deletion} & \multicolumn{1}{c}{\bfseries User deletion} \\

\midrule
\endhead
\midrule
\multicolumn{7}{r}{\textit{Continued on next page}} \\
\multicolumn{7}{r}{\raggedleft{\itshape * p{\textless}0.05~~~** p{\textless}0.01~~~*** p{\textless}0.001}} \\
\endfoot
\bottomrule
\multicolumn{7}{r}{\raggedleft{\itshape * p{\textless}0.05~~~** p{\textless}0.01~~~*** p{\textless}0.001}} \\
\endlastfoot
Intercept &
\makecell{\textbf{56.04 \textsuperscript{***}} \\ (54.34~--~57.74)} &
\makecell{\textbf{57.74 \textsuperscript{***}} \\ (56.10~--~59.38)} &
\makecell{\textbf{57.75 \textsuperscript{***}} \\ (56.14~--~59.36)} &
\makecell{\textbf{57.33 \textsuperscript{***}} \\ (55.67~--~58.98)} &
\makecell{\textbf{56.85 \textsuperscript{***}} \\ (55.15~--~58.54)} &
\makecell{\textbf{56.71 \textsuperscript{***}} \\ (54.87~--~58.54)}\\
DPFT &
\makecell{\textbf{2.26 \textsuperscript{**}} \\ (0.87~--~3.65)} &
\makecell{\textbf{3.30 \textsuperscript{***}} \\ (1.94~--~4.66)} &
\makecell{\textbf{4.33 \textsuperscript{***}} \\ (3.00~--~5.65)} &
\makecell{\textbf{5.33 \textsuperscript{***}} \\ (3.99~--~6.68)} &
\makecell{\textbf{5.55 \textsuperscript{***}} \\ (4.16~--~6.94)} &
\makecell{\textbf{5.79 \textsuperscript{***}} \\ (4.29~--~7.29)}\\
PPFT &
\makecell{\textbf{6.30 \textsuperscript{***}} \\ (4.92~--~7.69)} &
\makecell{\textbf{6.21 \textsuperscript{***}} \\ (4.86~--~7.57)} &
\makecell{\textbf{5.78 \textsuperscript{***}} \\ (4.47~--~7.10)} &
\makecell{\textbf{6.23 \textsuperscript{***}} \\ (4.89~--~7.56)} &
\makecell{\textbf{6.23 \textsuperscript{***}} \\ (4.84~--~7.62)} &
\makecell{\textbf{6.20 \textsuperscript{***}} \\ (4.69~--~7.70)}\\
Prompt &
\makecell{\textbf{-3.83 \textsuperscript{***}} \\ (-5.22~--~-2.44)} &
\makecell{\textbf{-3.91 \textsuperscript{***}} \\ (-5.27~--~-2.55)} &
\makecell{\textbf{-3.61 \textsuperscript{***}} \\ (-4.93~--~-2.30)} &
\makecell{\textbf{-3.63 \textsuperscript{***}} \\ (-4.96~--~-2.30)} &
\makecell{\textbf{-3.52 \textsuperscript{***}} \\ (-4.91~--~-2.13)} &
\makecell{\textbf{-3.14 \textsuperscript{***}} \\ (-4.64~--~-1.64)}\\
\end{longtable}
}
\normalsize

\subsubsection{Contamination Test}
Since participants evaluate all four models within the same trial, an error for one model could inflate ratings of the others. We test for this by restricting to error-free observations and regressing preference ratings on the number of \textit{other} models in the trial that experienced any error ($N^{\text{other}}_{ik}$), controlling for model, domain, and participant random intercepts:
\begin{equation}
    Y_{ijk} = \beta_0 + \sum_{m} \beta_m \cdot \mathbf{1}_{j=m} + \gamma \cdot N^{\text{other}}_{ik} + \sum_{d} \delta_d \cdot \mathbf{1}_{\text{domain}=d} + u_i + \varepsilon_{ijk}
\end{equation}
We also fit a variant that splits the count into first-turn ($N^{\text{other,ft}}_{ik}$) and subsequent ($N^{\text{other,sub}}_{ik}$) errors separately. In this split specification, a competitor's first-turn error produced a small but significant rating boost ($\hat{\gamma}_1 = 1.69$, $p < .001$), while subsequent errors had no effect ($\hat{\gamma}_2 = -0.31$, $p = .22$). This contamination boost is modest relative to the direct penalty on the erring model ($\approx -56$ points for first-turn errors), so we do not include contamination covariates in the main specifications (\cref{tab:gpu_error_contamination}).

\scriptsize{
\setlength{\LTcapwidth}{\textwidth}
\renewcommand\arraystretch{1.5}
\setlength\tabcolsep{0.8mm}
\begin{longtable}{llcc}
\caption{\small{Contamination test: do other models' errors affect error-free ratings? Pooled = total error count in trial. Split = first-turn vs subsequent errors.}} \\
\hhline{---}
\multicolumn{3}{l}{Dependent Variable = \textit{Preference Rating (0--100)}} \\
\hhline{===}
{\bfseries ~} & \multicolumn{1}{c}{\bfseries Pooled} & \multicolumn{1}{c}{\bfseries Split} \\

\midrule
\endfirsthead
\multicolumn{3}{c}{\tablename\ \thetable{} -- \textit{Continued from previous page}} \\
\hhline{---}
\multicolumn{3}{l}{Dependent Variable = \textit{Preference Rating (0--100)}} \\
\hhline{===}
{\bfseries ~} & \multicolumn{1}{c}{\bfseries Pooled} & \multicolumn{1}{c}{\bfseries Split} \\

\midrule
\endhead
\midrule
\multicolumn{3}{r}{\textit{Continued on next page}} \\
\multicolumn{3}{r}{\raggedleft{\itshape * p{\textless}0.05~~~** p{\textless}0.01~~~*** p{\textless}0.001}} \\
\endfoot
\bottomrule
\multicolumn{3}{r}{\raggedleft{\itshape * p{\textless}0.05~~~** p{\textless}0.01~~~*** p{\textless}0.001}} \\
\endlastfoot
Intercept &
\makecell{\textbf{56.95 \textsuperscript{***}} \\ (55.29~--~58.61)} &
\makecell{\textbf{56.87 \textsuperscript{***}} \\ (55.22~--~58.53)}\\
DPFT &
\makecell{\textbf{5.42 \textsuperscript{***}} \\ (4.07~--~6.77)} &
\makecell{\textbf{5.40 \textsuperscript{***}} \\ (4.06~--~6.75)}\\
PPFT &
\makecell{\textbf{6.20 \textsuperscript{***}} \\ (4.86~--~7.54)} &
\makecell{\textbf{6.05 \textsuperscript{***}} \\ (4.71~--~7.39)}\\
Prompt &
\makecell{\textbf{-3.65 \textsuperscript{***}} \\ (-4.98~--~-2.31)} &
\makecell{\textbf{-3.64 \textsuperscript{***}} \\ (-4.97~--~-2.31)}\\
N Other Errors in Trial &
\makecell{\textbf{0.55 \textsuperscript{**}} \\ (0.18~--~0.92)} &~\\
Emot. &
\makecell{\textbf{3.60 \textsuperscript{***}} \\ (2.26~--~4.94)} &
\makecell{\textbf{3.69 \textsuperscript{***}} \\ (2.36~--~5.03)}\\
Values &
\makecell{\textbf{2.21 \textsuperscript{**}} \\ (0.87~--~3.55)} &
\makecell{\textbf{2.26 \textsuperscript{**}} \\ (0.91~--~3.60)}\\
Controv. &
\makecell{-0.35 \\ (-1.69~--~0.99)} &
\makecell{-0.40 \\ (-1.74~--~0.94)}\\
\end{longtable}
}
\normalsize

\newpage

\subsection{Opening Choice Task}
\label{sec:appendix_choices}
\normalsize

\begin{figure}[H]
    \centering
    \includegraphics[width=\linewidth]{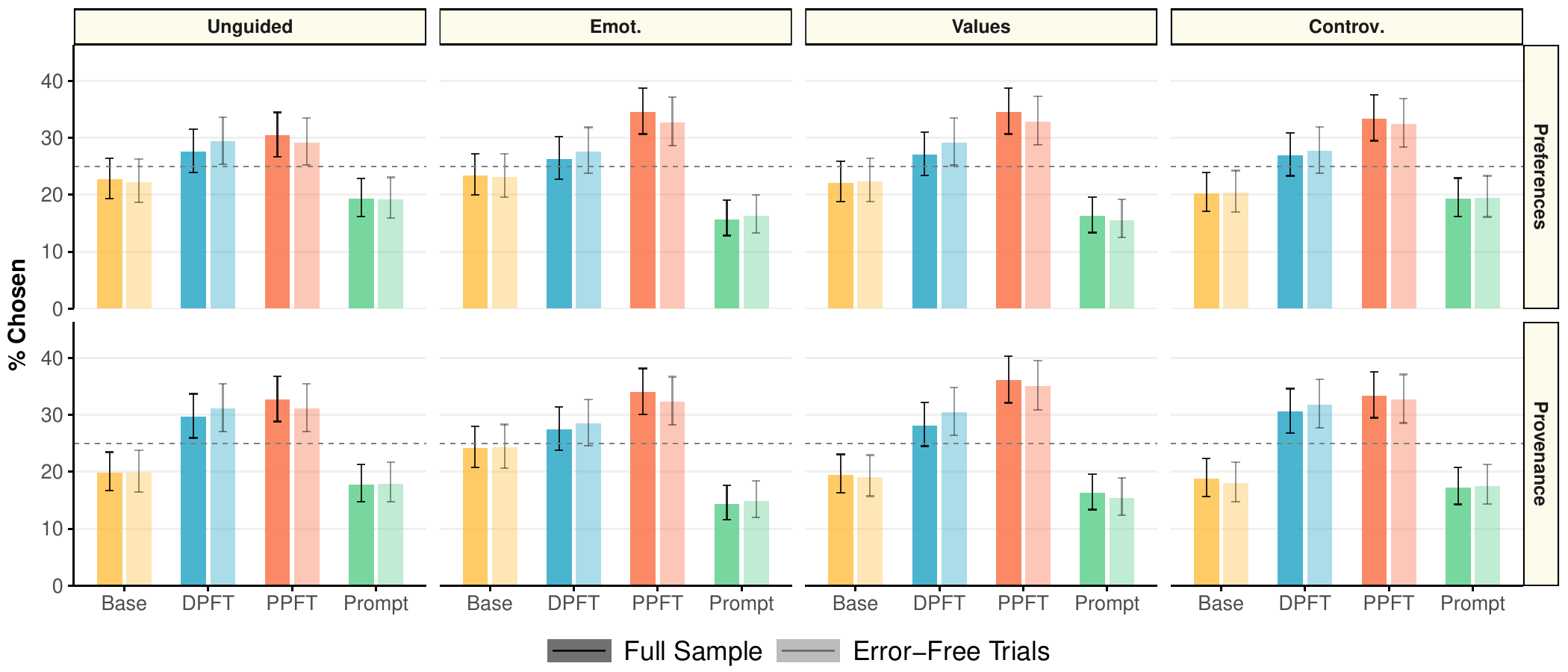}
    \caption{\textbf{Opening choice frequencies by model and domain.} Percentage of times each model was chosen, faceted by choice type (rows) and domain (columns). Dark bars = full sample; light bars = error-free trials only. Dashed line = 25\% chance level. Error bars show 95\% CIs.}
    \label{fig:opening_choice_frequencies}
\end{figure}

After viewing a single opening turn from all four models, participants selected which response they preferred most (``I like this response the most'') and which felt most personalised to them (``This response is the most \textbf{personalised} to me''). Each participant made one choice per domain (4 domains). \Cref{fig:opening_choice_frequencies} shows choice frequencies by model and domain, comparing the full sample with error-free trials.

\subsubsection{Model Specifications}
We use conditional logistic regression \citep{mcfaddenConditional1974} to analyse forced-choice outcomes. Within each trial, participants chose exactly one of four models, so choice probabilities are constrained to sum to one within each choice set. We construct manual dummy variables for model (Base = reference) and domain (UnguidedChat = reference), so main effects represent model advantages over Base in the reference domain, and interaction terms represent how that advantage \textit{changes} in other domains. Cluster-robust standard errors account for repeated observations within participants. All models include $E^{(1)}_{ijk}$ (first-turn error) as a covariate, since the opening choice is made after seeing only the first turn. Domain main effects are absorbed by stratification on the choice set $\mathcal{C}_k$.

We fit three specifications per choice type (preferences [1]--[3], provenance [4]--[6]):

\textbf{Model [1]/[4] (Judgement model):}
\begin{equation}
    \Pr(Y_{ijk} = 1 \mid \mathcal{C}_k) \propto \exp\!\left(\sum_{m} \beta_m \cdot \mathbf{1}_{j=m} + \gamma_1 E^{(1)}_{ijk}\right)
\end{equation}
where $m \in \{\text{DPFT}, \text{PPFT}, \text{Prompting}\}$ (Base is the reference). Domain is implicitly controlled via stratification on the choice set $\mathcal{C}_k$.

\textbf{Model [2]/[5] (Domain interactions):}
\begin{equation}
    \Pr(Y_{ijk} = 1 \mid \mathcal{C}_k) \propto \exp\!\left(\sum_{m} \beta_m \cdot \mathbf{1}_{j=m} + \sum_{m}\sum_{d} \delta_{md} \cdot \mathbf{1}_{j=m} \cdot \mathbf{1}_{\text{domain}=d} + \gamma_1 E^{(1)}_{ijk}\right)
\end{equation}
where $d \in \{\text{EmotChat}, \text{ValuesChat}, \text{ControversyChat}\}$ (UnguidedChat is the reference). This tests whether model effects vary across domains.

\textbf{Model [3]/[6] (Out-of-domain generalisation):}
\begin{equation}
    \Pr(Y_{ijk} = 1 \mid \mathcal{C}_k) \propto \exp\!\left(\sum_{m} \beta_m \cdot \mathbf{1}_{j=m} + \sum_{m} \lambda_m \cdot \mathbf{1}_{j=m} \cdot \mathbf{1}_{\text{out}} + \gamma_1 E^{(1)}_{ijk}\right)
\end{equation}
where $\mathbf{1}_{\text{out}} = \mathbf{1}_{\text{domain}=\text{EmotChat}}$ flags the out-of-domain condition. Interaction terms $\lambda_m$ test whether model advantages generalise beyond the training domains.

All models use cluster-robust standard errors (clustered on participant) and Efron's method for tied event times. Within the Prompting condition, we additionally test whether the probability of the Prompting model being chosen varies by model size (8B vs 70B) and prompt type (Summariser vs Demographics) using mixed-effects logistic regression on the binary chosen indicator with participant random intercepts (\cref{tab:choice_prompting}). Regression results are in \cref{tab:choice_models}; pairwise contrasts in \cref{tab:choice_pairwise}.

\scriptsize{
\setlength{\LTcapwidth}{\textwidth}
\renewcommand\arraystretch{1.5}
\setlength\tabcolsep{0.8mm}
\begin{longtable}{llcccccc}
\caption{\small{Conditional logistic regression for opening choices (odds ratios). Pref = preference, Prov = provenance judgement model. (D) = domain interactions. (OOD) = out-of-domain. First-turn error covariates and cluster-robust standard errors included.}} \\
\hhline{-------}
\multicolumn{7}{l}{Dependent Variable = \textit{Was Chosen (OR)}} \\
\hhline{=======}
{\bfseries ~} & \multicolumn{1}{c}{\bfseries Pref} & \multicolumn{1}{c}{\bfseries Pref (D)} & \multicolumn{1}{c}{\bfseries Pref (OOD)} & \multicolumn{1}{c}{\bfseries Prov} & \multicolumn{1}{c}{\bfseries Prov (D)} & \multicolumn{1}{c}{\bfseries Prov (OOD)} \\

\midrule
\endfirsthead
\multicolumn{7}{c}{\tablename\ \thetable{} -- \textit{Continued from previous page}} \\
\hhline{-------}
\multicolumn{7}{l}{Dependent Variable = \textit{Was Chosen (OR)}} \\
\hhline{=======}
{\bfseries ~} & \multicolumn{1}{c}{\bfseries Pref} & \multicolumn{1}{c}{\bfseries Pref (D)} & \multicolumn{1}{c}{\bfseries Pref (OOD)} & \multicolumn{1}{c}{\bfseries Prov} & \multicolumn{1}{c}{\bfseries Prov (D)} & \multicolumn{1}{c}{\bfseries Prov (OOD)} \\

\midrule
\endhead
\midrule
\multicolumn{7}{r}{\textit{Continued on next page}} \\
\multicolumn{7}{r}{\raggedleft{\itshape * p{\textless}0.05~~~** p{\textless}0.01~~~*** p{\textless}0.001}} \\
\endfoot
\bottomrule
\multicolumn{7}{r}{\raggedleft{\itshape * p{\textless}0.05~~~** p{\textless}0.01~~~*** p{\textless}0.001}} \\
\endlastfoot
DPFT &
\makecell{\textbf{1.31 \textsuperscript{***}} \\ (1.16~--~1.48)} &
\makecell{\textbf{1.31 \textsuperscript{*}} \\ (1.03~--~1.67)} &
\makecell{\textbf{1.35 \textsuperscript{***}} \\ (1.17~--~1.56)} &
\makecell{\textbf{1.51 \textsuperscript{***}} \\ (1.34~--~1.71)} &
\makecell{\textbf{1.62 \textsuperscript{***}} \\ (1.26~--~2.07)} &
\makecell{\textbf{1.65 \textsuperscript{***}} \\ (1.42~--~1.90)}\\
PPFT &
\makecell{\textbf{1.46 \textsuperscript{***}} \\ (1.30~--~1.65)} &
\makecell{\textbf{1.30 \textsuperscript{*}} \\ (1.03~--~1.66)} &
\makecell{\textbf{1.47 \textsuperscript{***}} \\ (1.28~--~1.69)} &
\makecell{\textbf{1.62 \textsuperscript{***}} \\ (1.43~--~1.82)} &
\makecell{\textbf{1.61 \textsuperscript{***}} \\ (1.26~--~2.06)} &
\makecell{\textbf{1.72 \textsuperscript{***}} \\ (1.49~--~1.98)}\\
Prompt &
\makecell{\textbf{0.80 \textsuperscript{**}} \\ (0.70~--~0.92)} &
\makecell{0.85 \\ (0.65~--~1.11)} &
\makecell{\textbf{0.85 \textsuperscript{*}} \\ (0.72~--~0.99)} &
\makecell{\textbf{0.80 \textsuperscript{**}} \\ (0.70~--~0.92)} &
\makecell{0.89 \\ (0.68~--~1.18)} &
\makecell{0.89 \\ (0.75~--~1.04)}\\
1st-turn Error &
\makecell{\textbf{0.01 \textsuperscript{***}} \\ (0.00~--~0.05)} &
\makecell{\textbf{0.01 \textsuperscript{***}} \\ (0.00~--~0.05)} &
\makecell{\textbf{0.01 \textsuperscript{***}} \\ (0.00~--~0.05)} &
\makecell{\textbf{0.02 \textsuperscript{***}} \\ (0.01~--~0.06)} &
\makecell{\textbf{0.02 \textsuperscript{***}} \\ (0.01~--~0.06)} &
\makecell{\textbf{0.02 \textsuperscript{***}} \\ (0.01~--~0.06)}\\
DPFT:Emot. &~ &
\makecell{0.90 \\ (0.64~--~1.28)} &~ &~ &
\makecell{0.74 \\ (0.52~--~1.05)} &~\\
PPFT:Emot. &~ &
\makecell{1.10 \\ (0.79~--~1.54)} &~ &~ &
\makecell{0.85 \\ (0.61~--~1.19)} &~\\
Prompt:Emot. &~ &
\makecell{0.80 \\ (0.54~--~1.17)} &~ &~ &
\makecell{\textbf{0.67 \textsuperscript{*}} \\ (0.45~--~1.00)} &~\\
DPFT:Values &~ &
\makecell{1.01 \\ (0.72~--~1.43)} &~ &~ &
\makecell{0.97 \\ (0.68~--~1.39)} &~\\
\end{longtable}
}
\normalsize
\scriptsize{
\setlength{\LTcapwidth}{\textwidth}
\renewcommand\arraystretch{1.3}
\setlength\tabcolsep{1.5mm}
\begin{longtable}{lrrrrrrrr}
\caption{\small{Pairwise Wald contrasts between models (opening choice).}} \\
\toprule
Contrast & \multicolumn{4}{c}{\bfseries Preferences} & \multicolumn{4}{c}{\bfseries Provenance} \\
\cmidrule(lr){2-5}
\cmidrule(lr){6-9}
& $\Delta$ & SE & $p$ & $p_{\text{FDR}}$ & $\Delta$ & SE & $p$ & $p_{\text{FDR}}$ \\
\midrule
\endfirsthead
\endlastfoot
DPFT vs PPFT & -0.113 & 0.056 & 0.045 & 0.054 & -0.067 & 0.055 & 0.223 & 0.223 \\
DPFT vs Prompting & 0.490 & 0.070 & {\textless}.001 & {\textless}.001 & 0.636 & 0.070 & {\textless}.001 & {\textless}.001 \\
PPFT vs Prompting & 0.603 & 0.066 & {\textless}.001 & {\textless}.001 & 0.702 & 0.067 & {\textless}.001 & {\textless}.001 \\
\bottomrule
\label{tab:choice_pairwise}
\end{longtable}
}
\normalsize

\scriptsize{
\setlength{\LTcapwidth}{\textwidth}
\renewcommand\arraystretch{1.5}
\setlength\tabcolsep{0.8mm}
\begin{longtable}{llcccc}
\caption{\small{Prompting sub-conditions (opening choice). (2x2) = model size + prompt type. (4-lev) = 4-level sub-condition. Ref: 8B (size), Summariser (prompt), 8B-Summariser (4-level). First-turn error covariates and participant random intercepts included.}} \\
\hhline{-----}
\multicolumn{5}{l}{Dependent Variable = \textit{Was Chosen [1]-[2] Preferences, [3]-[4] Provenance}} \\
\hhline{=====}
{\bfseries ~} & \multicolumn{1}{c}{\bfseries Pref (2x2)} & \multicolumn{1}{c}{\bfseries Pref (4-lev)} & \multicolumn{1}{c}{\bfseries Prov (2x2)} & \multicolumn{1}{c}{\bfseries Prov (4-lev)} \\

\midrule
\endfirsthead
\multicolumn{5}{c}{\tablename\ \thetable{} -- \textit{Continued from previous page}} \\
\hhline{-----}
\multicolumn{5}{l}{Dependent Variable = \textit{Was Chosen [1]-[2] Preferences, [3]-[4] Provenance}} \\
\hhline{=====}
{\bfseries ~} & \multicolumn{1}{c}{\bfseries Pref (2x2)} & \multicolumn{1}{c}{\bfseries Pref (4-lev)} & \multicolumn{1}{c}{\bfseries Prov (2x2)} & \multicolumn{1}{c}{\bfseries Prov (4-lev)} \\

\midrule
\endhead
\midrule
\multicolumn{5}{r}{\textit{Continued on next page}} \\
\multicolumn{5}{r}{\raggedleft{\itshape * p{\textless}0.05~~~** p{\textless}0.01~~~*** p{\textless}0.001}} \\
\endfoot
\bottomrule
\multicolumn{5}{r}{\raggedleft{\itshape * p{\textless}0.05~~~** p{\textless}0.01~~~*** p{\textless}0.001}} \\
\endlastfoot
Intercept &
\makecell{\textbf{0.24 \textsuperscript{***}} \\ (0.19~--~0.30)} &
\makecell{\textbf{0.22 \textsuperscript{***}} \\ (0.17~--~0.29)} &
\makecell{\textbf{0.23 \textsuperscript{***}} \\ (0.19~--~0.29)} &
\makecell{\textbf{0.22 \textsuperscript{***}} \\ (0.17~--~0.28)}\\
Model Size (70B) &
\makecell{\textbf{0.70 \textsuperscript{**}} \\ (0.55~--~0.89)} &~ &
\makecell{\textbf{0.66 \textsuperscript{**}} \\ (0.52~--~0.85)} &~\\
Prompt (Demo.) &
\makecell{1.07 \\ (0.84~--~1.36)} &~ &
\makecell{0.97 \\ (0.76~--~1.24)} &~\\
First-turn Error &
\makecell{\textbf{0.07 \textsuperscript{*}} \\ (0.01~--~0.54)} &
\makecell{\textbf{0.08 \textsuperscript{*}} \\ (0.01~--~0.55)} &
\makecell{\textbf{0.17 \textsuperscript{*}} \\ (0.04~--~0.70)} &
\makecell{\textbf{0.17 \textsuperscript{*}} \\ (0.04~--~0.71)}\\
8B-Demo. &~ &
\makecell{1.23 \\ (0.89~--~1.71)} &~ &
\makecell{1.10 \\ (0.79~--~1.54)}\\
70B-Summariser &~ &
\makecell{0.81 \\ (0.58~--~1.14)} &~ &
\makecell{0.76 \\ (0.54~--~1.08)}\\
70B-Demo. &~ &
\makecell{0.74 \\ (0.52~--~1.04)} &~ &
\makecell{\textbf{0.63 \textsuperscript{*}} \\ (0.44~--~0.90)}\\
 \hline
\multicolumn{5}{l}{{\bfseries Random Effects}} \\
\end{longtable}
}
\normalsize
\normalsize

\subsection{Ranking Task}
\label{sec:appendix_ranking}
\begin{figure}[H]
    \centering
    \includegraphics[width=\linewidth]{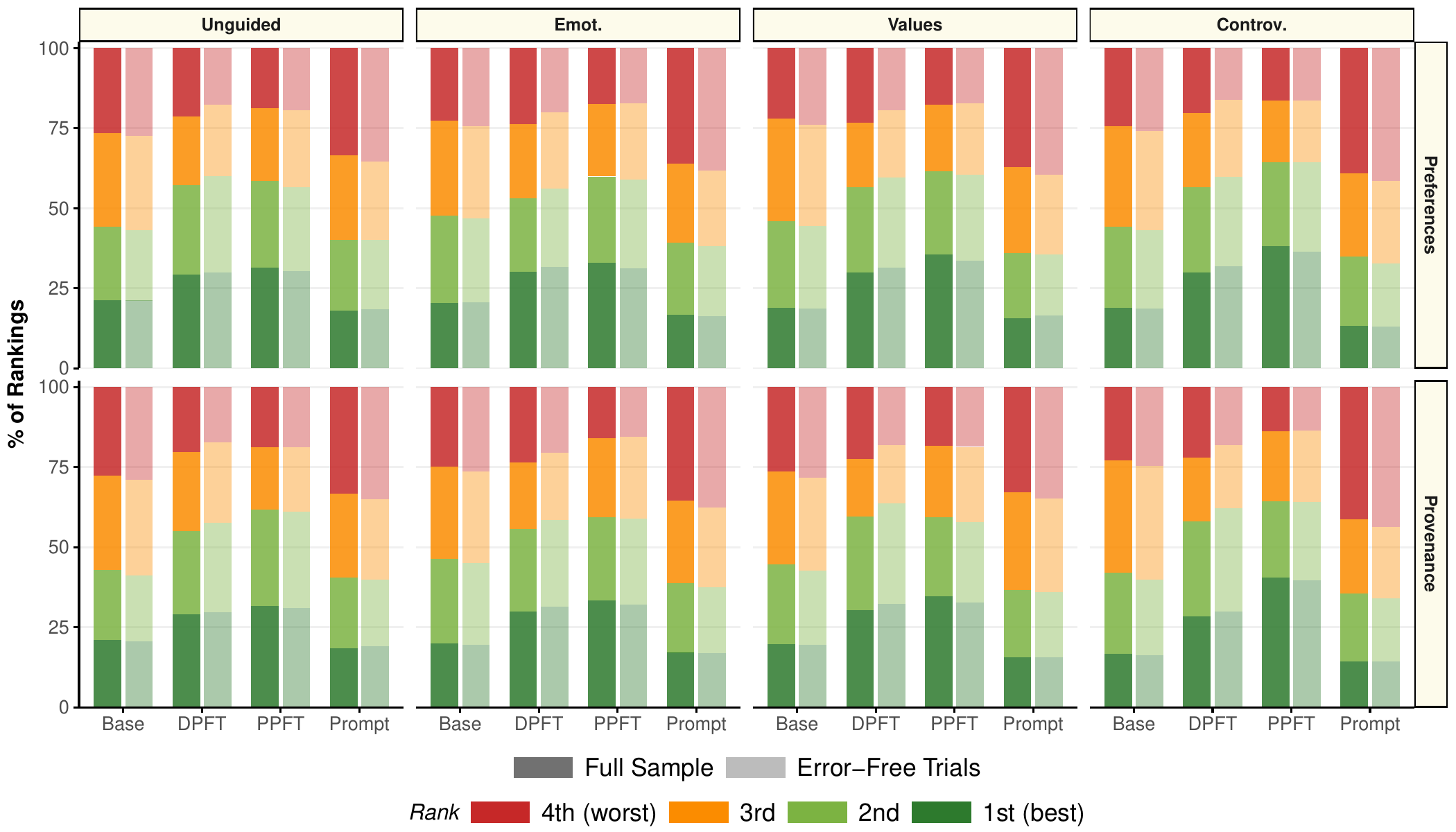}
    \caption{\textbf{Full rank distribution by model.} Percentage of participants assigning each rank to each model, faceted by ranking type. Dark bars = full sample; light bars = error-free trials only.}
    \label{fig:ranking_stacked_bar}
\end{figure}

\begin{figure}[H]
    \centering
    \includegraphics[width=\linewidth]{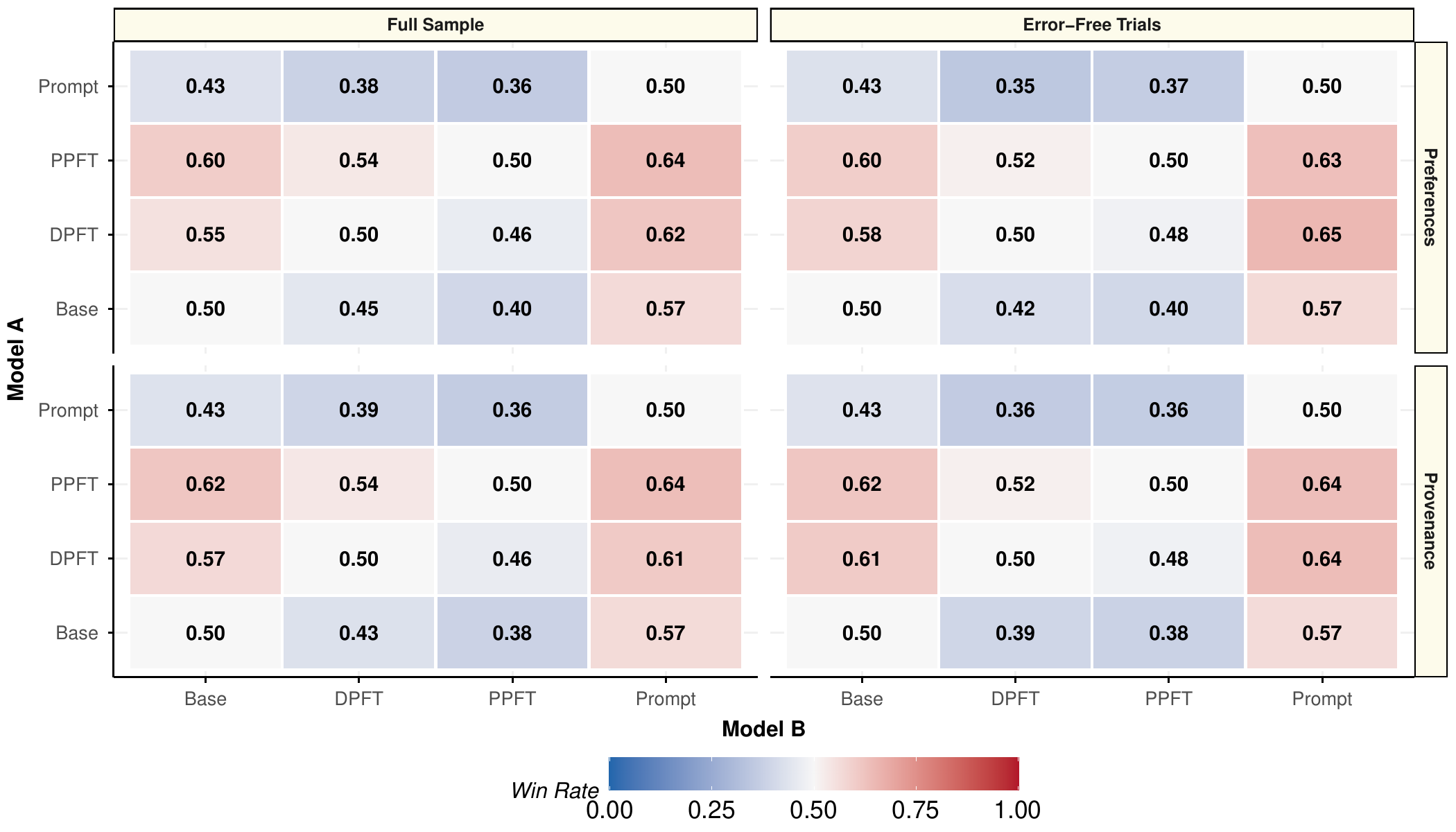}
    \caption{\textbf{Pairwise win rate matrix.} Each cell shows the win rate for the row model against the column model, derived from within-trial rank comparisons.}
    \label{fig:ranking_winrate_heatmap}
\end{figure}

\begin{figure}[H]
    \centering
    \includegraphics[width=\linewidth]{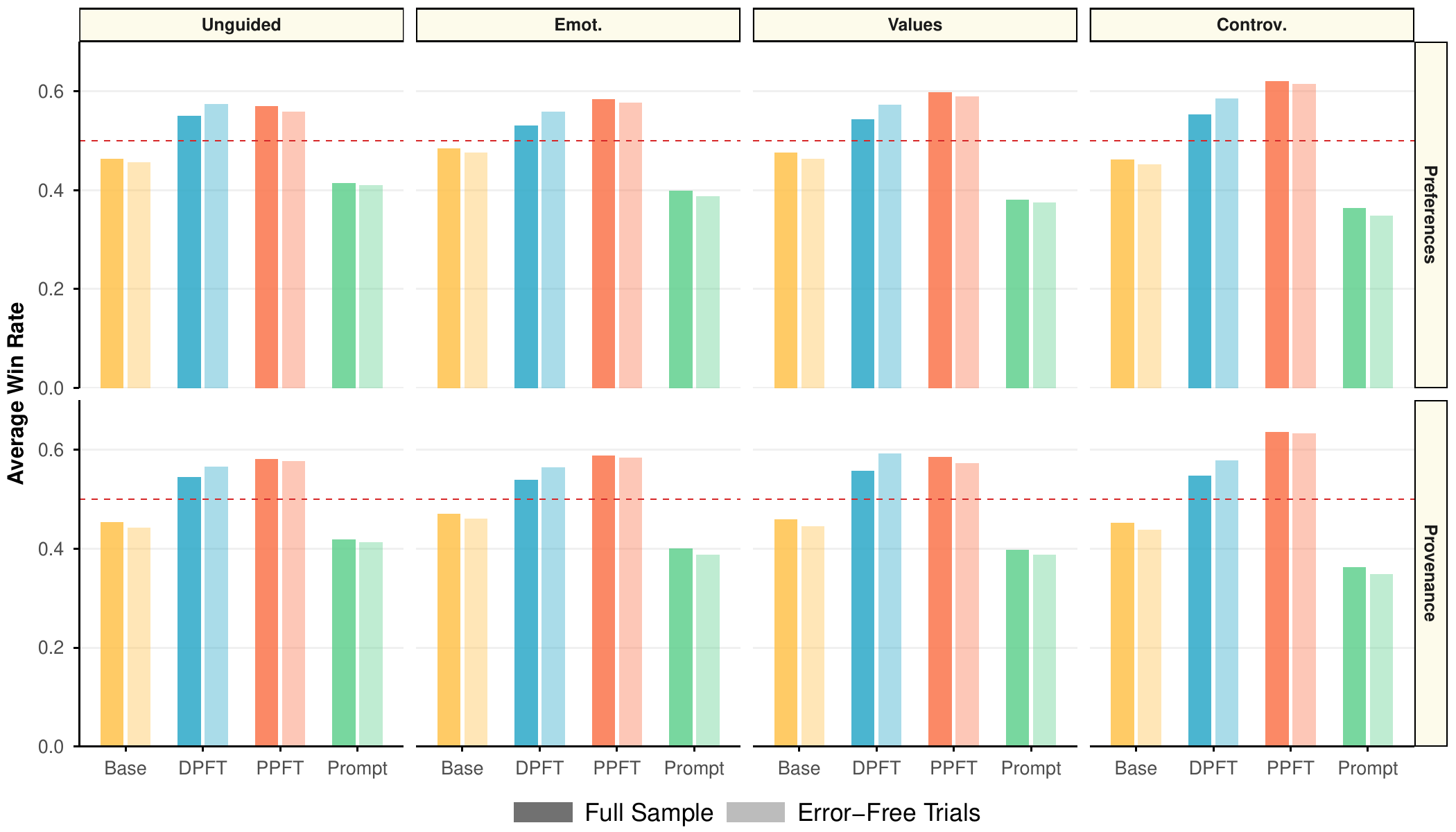}
    \caption{\textbf{Average win rates by domain.} Win rates disaggregated by conversation domain.}
    \label{fig:ranking_avg_winrates_by_domain}
\end{figure}

\begin{figure}[H]
    \centering
    \includegraphics[width=\linewidth]{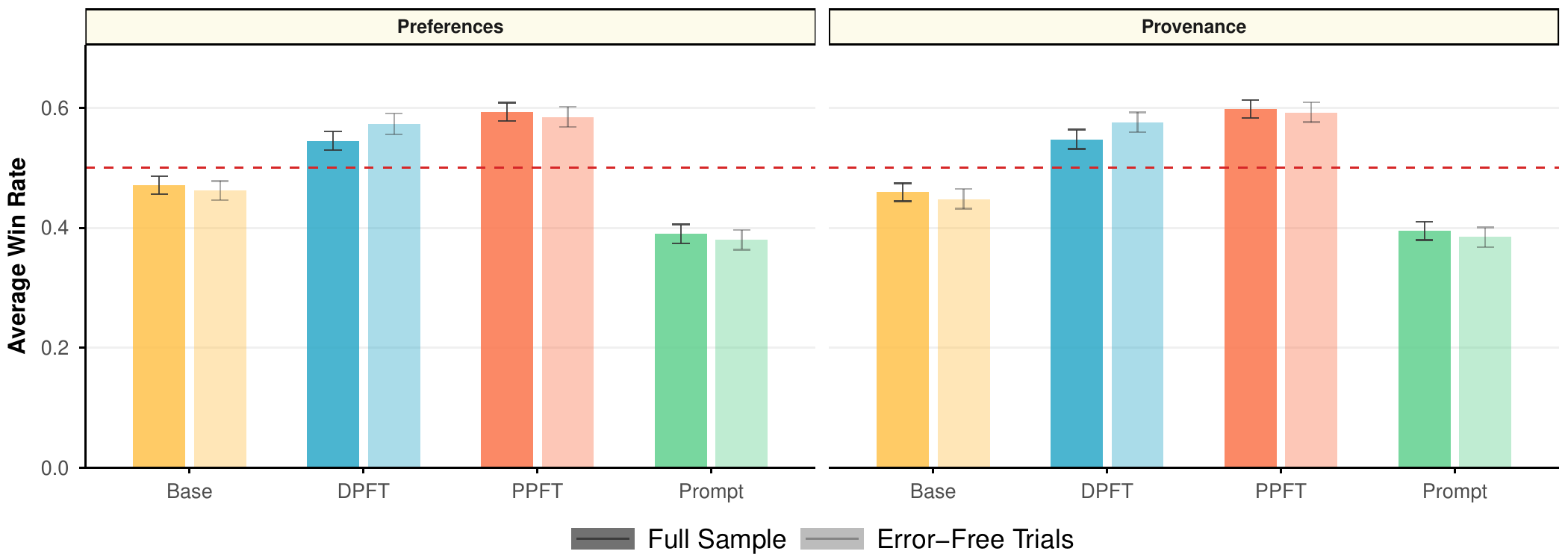}
    \caption{\textbf{Average win rates across all pairwise comparisons.} Mean win rates with 95\% bootstrap CIs (1,000 iterations, resampling participants).}
    \label{fig:ranking_avg_winrates}
\end{figure}

\normalsize
After completing two-minute conversations with all four models simultaneously, participants ranked all four models from best (4) to worst (1) for both preferences (``Rank the models according to your preferences'') and perceived personalisation/provenance (``Rank the models according to how personalised they felt''). Each participant completed one ranking per domain (4 domains), yielding 4 trials per ranking type per participant. \Cref{fig:ranking_stacked_bar} shows the full rank distribution by model and ranking type, comparing the full sample with error-free trials. \Cref{fig:ranking_winrate_heatmap} shows pairwise winrates; \Cref{fig:ranking_avg_winrates_by_domain} and \Cref{fig:ranking_avg_winrates} show average winrates.

\subsubsection{Model Specifications}
We analyse ranking outcomes using three complementary approaches:
\begin{itemize}
    \item We reduce each ranking to a binary outcome indicating which model was ranked best within each trial, then fit \textbf{conditional logistic regression} \citep{mcfaddenConditional1974} stratified on trial (choice set). This approach is directly comparable to the opening choice analysis (\cref{sec:appendix_choices}).
    \item To exploit the full rank ordering (not just the top choice), we use the \textbf{rank-ordered logit} \citep{beggsAssessing1981}, which decomposes each ranking into a sequence of conditional choices.
    \item We fit \textbf{Plackett-Luce models} \citep{plackettAnalysis1975, luceIndividual2005} on error-free trials to estimate a worth parameter $\lambda_j$ for each model $j$, where the probability of model $j$ being ranked first among alternatives $\mathcal{A}$ is:
\begin{equation}
P(j \text{ ranked first} \mid \mathcal{A}) = \frac{\lambda_j}{\sum_{k \in \mathcal{A}} \lambda_k}
\end{equation}
We parameterise $\lambda_j = \exp(\beta_j)$ with Base fixed at $\beta_{\text{Base}} = 0$ as the reference. The model also yields pairwise win probabilities: $P(i \succ j) = \exp(\beta_i) / [\exp(\beta_i) + \exp(\beta_j)]$.
\end{itemize}

For the ranked-best and rank-ordered approaches, we fit the same three specifications as the opening choice analysis (\cref{sec:appendix_choices}), with the addition of $E^{(2+)}_{ijk}$ (subsequent error) as a covariate since rankings are made after the full conversation. Within the Prompting condition, we test effects of model size and prompt type using mixed-effects logistic regression with participant random intercepts (\cref{tab:ranking_prompting}). Plackett-Luce worth parameters and pairwise win probabilities are in \cref{tab:ranking_pl}). Judgement model results are in \cref{tab:ranking_models}; domain and OOD robustness checks in \cref{tab:ranking_robustness}. Pairwise Wald contrasts are in \cref{tab:ranking_clogit_pairwise,tab:ranking_rologit_pairwise}.

\begin{table}[H]
{\scriptsize
\renewcommand\arraystretch{1.5}
\caption{Plackett-Luce results on error-free trials. (a)~Worth parameters (probability of being ranked first). (b)~Pairwise win probabilities. 95\% CIs from delta method.}
\label{tab:ranking_pl}
\vspace{2mm}
\begin{minipage}[t]{0.38\textwidth}
\centering
\textbf{(a)} Worth parameters\\[2mm]
\begin{tabular}{lcc}
\toprule
Model & Pref & Prov \\
\midrule
Base & \makecell{0.22 \\ (0.21~--~0.23)} & \makecell{0.21 \\ (0.20~--~0.22)} \\
DPFT & \makecell{0.30 \\ (0.29~--~0.32)} & \makecell{0.30 \\ (0.29~--~0.32)} \\
PPFT & \makecell{0.31 \\ (0.30~--~0.32)} & \makecell{0.32 \\ (0.30~--~0.33)} \\
Prompt & \makecell{0.17 \\ (0.16~--~0.18)} & \makecell{0.17 \\ (0.16~--~0.18)} \\
\bottomrule
\end{tabular}
\end{minipage}%
\hfill%
\begin{minipage}[t]{0.58\textwidth}
\centering
\textbf{(b)} Pairwise win probabilities\\[2mm]
\begin{tabular}{lcc}
\toprule
Comparison & Pref & Prov \\
\midrule
DPFT vs Base & \makecell{0.58 \\ (0.56~--~0.60)} & \makecell{0.59 \\ (0.57~--~0.61)} \\
PPFT vs Base & \makecell{0.59 \\ (0.57~--~0.60)} & \makecell{0.60 \\ (0.58~--~0.62)} \\
Prompt vs Base & \makecell{0.43 \\ (0.41~--~0.45)} & \makecell{0.45 \\ (0.43~--~0.47)} \\
PPFT vs DPFT & \makecell{0.51 \\ (0.49~--~0.53)} & \makecell{0.51 \\ (0.49~--~0.53)} \\
PPFT vs Prompt & \makecell{0.65 \\ (0.63~--~0.67)} & \makecell{0.65 \\ (0.63~--~0.67)} \\
DPFT vs Prompt & \makecell{0.64 \\ (0.62~--~0.66)} & \makecell{0.64 \\ (0.62~--~0.66)} \\
\bottomrule
\end{tabular}
\end{minipage}
}
\end{table}
\input{tables/SI/ranking_models}
\input{tables/SI/ranking_robustness}
\scriptsize{
\setlength{\LTcapwidth}{\textwidth}
\renewcommand\arraystretch{1.3}
\setlength\tabcolsep{1.5mm}
\begin{longtable}{lrrrrrrrr}
\caption{\small{Pairwise Wald contrasts between models (ranked-best clogit).}} \\
\toprule
Contrast & \multicolumn{4}{c}{\bfseries Preferences} & \multicolumn{4}{c}{\bfseries Provenance} \\
\cmidrule(lr){2-5}
\cmidrule(lr){6-9}
& $\Delta$ & SE & $p$ & $p_{\text{FDR}}$ & $\Delta$ & SE & $p$ & $p_{\text{FDR}}$ \\
\midrule
\endfirsthead
\endlastfoot
DPFT vs PPFT & -0.058 & 0.057 & 0.309 & 0.309 & -0.089 & 0.056 & 0.110 & 0.132 \\
DPFT vs Prompting & 0.691 & 0.078 & {\textless}.001 & {\textless}.001 & 0.650 & 0.075 & {\textless}.001 & {\textless}.001 \\
PPFT vs Prompting & 0.750 & 0.074 & {\textless}.001 & {\textless}.001 & 0.739 & 0.073 & {\textless}.001 & {\textless}.001 \\
\bottomrule
\label{tab:ranking_clogit_pairwise}
\end{longtable}
}
\normalsize
\scriptsize{
\setlength{\LTcapwidth}{\textwidth}
\renewcommand\arraystretch{1.3}
\setlength\tabcolsep{1.5mm}
\begin{longtable}{lrrrrrrrr}
\caption{\small{Pairwise Wald contrasts between models (rank-ordered logit).}} \\
\toprule
Contrast & \multicolumn{4}{c}{\bfseries Preferences} & \multicolumn{4}{c}{\bfseries Provenance} \\
\cmidrule(lr){2-5}
\cmidrule(lr){6-9}
& $\Delta$ & SE & $p$ & $p_{\text{FDR}}$ & $\Delta$ & SE & $p$ & $p_{\text{FDR}}$ \\
\midrule
\endfirsthead
\endlastfoot
DPFT vs PPFT & -0.072 & 0.041 & 0.075 & 0.075 & -0.077 & 0.041 & 0.060 & 0.072 \\
DPFT vs Prompting & 0.525 & 0.047 & {\textless}.001 & {\textless}.001 & 0.521 & 0.048 & {\textless}.001 & {\textless}.001 \\
PPFT vs Prompting & 0.598 & 0.047 & {\textless}.001 & {\textless}.001 & 0.598 & 0.048 & {\textless}.001 & {\textless}.001 \\
\bottomrule
\label{tab:ranking_rologit_pairwise}
\end{longtable}
}
\normalsize

\input{tables/SI/ranking_prompting}
\normalsize

\subsection{Rating Task}
\label{sec:appendix_rating}
\normalsize

\begin{figure}[H]
    \centering
    \includegraphics[width=0.5\linewidth]{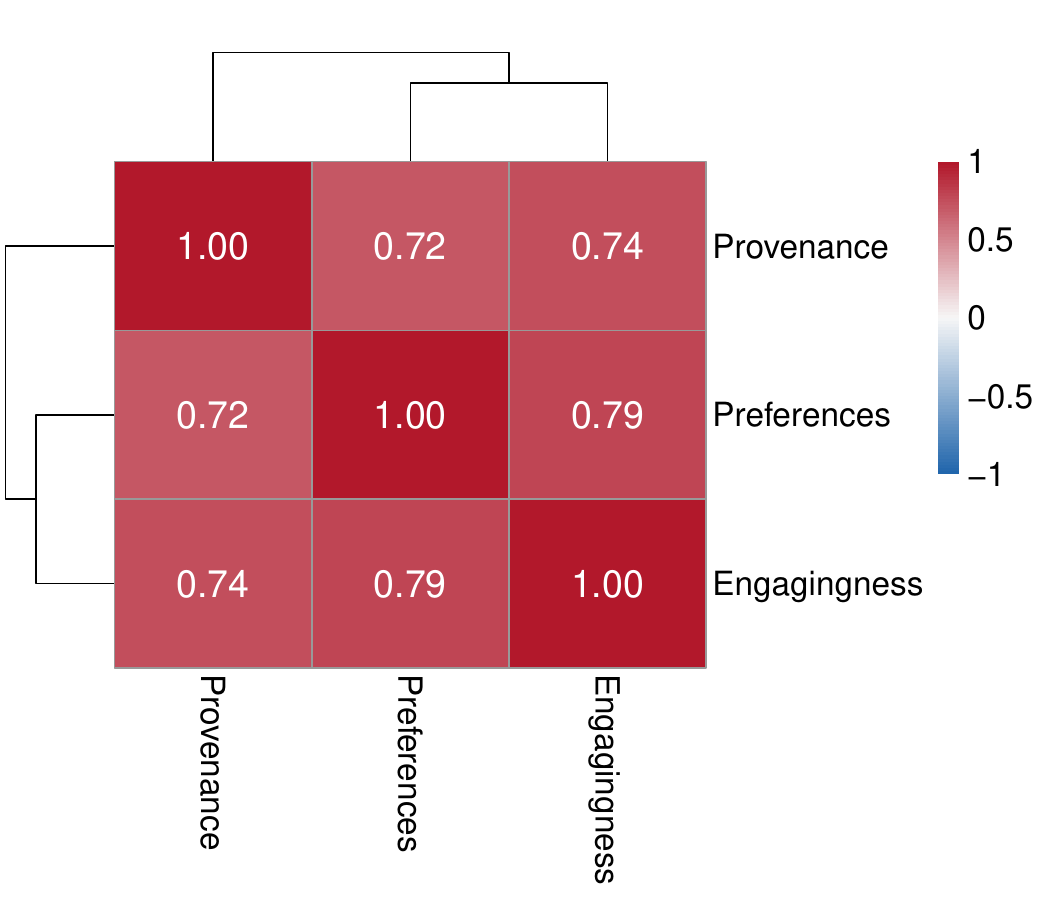}
    \caption{\textbf{Correlations among rating outcomes.} Pairwise correlations between preferences, engagingness, and provenance ratings. All three measures are positively correlated, with preferences and engagingness most strongly related.}
    \label{fig:rating_outcome_correlations}
\end{figure}

\begin{figure}[H]
    \centering
    \includegraphics[width=\linewidth]{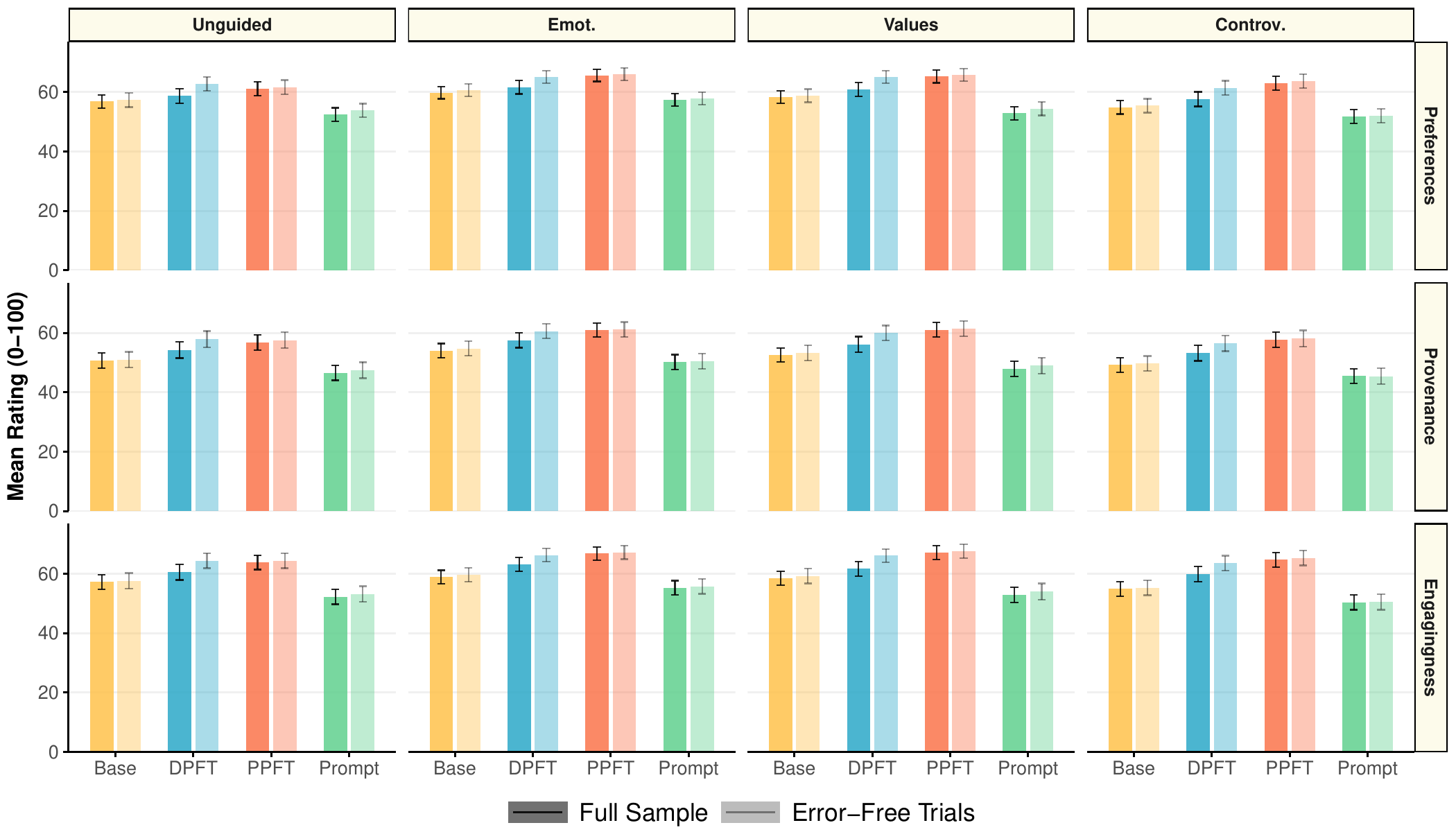}
    \caption{\textbf{Rating distributions by model.} Distribution of ratings (0--100 scale) for each model, faceted by rating type. Vertical lines indicate model means.}
    \label{fig:rating_distributions}
\end{figure}

After ranking the models, participants rated each model on three 0--100 slider scales:
\begin{itemize}[itemsep=1pt, parsep=0pt, topsep=0pt]
    \item \textbf{Preferences:} ``Rate the conversation with each model according to your preferences.''
    \item \textbf{Engagingness:} ``How engaging was the conversation with each AI model?''
    \item \textbf{Provenance:} ``How likely is it that each AI model has knowledge of and is personalised to your preferences?''
\end{itemize}
Each participant rated all four models across four domains, yielding 16 ratings per scale per participant. \cref{fig:rating_outcome_correlations} shows correlations among the three rating outcomes; \Cref{fig:rating_distributions} shows the rating distributions by model and scale.

\subsubsection{Model Specifications}

We use linear mixed-effects models to analyse continuous ratings (0--100 scale) with random intercepts for participants to account for repeated measures (16 ratings per participant: 4 models $\times$ 4 domains). We fit the same three specifications as the choice analysis (\cref{sec:appendix_choices}), adapted for continuous outcomes with domain main effects included directly and both error covariates ($E^{(1)}_{ijk}$, $E^{(2+)}_{ijk}$). The judgement model is:
\begin{equation}
Y_{ijk} = \beta_0 + \sum_{m} \beta_m \cdot \mathbf{1}_{j=m} + \sum_d \delta_d \cdot \mathbf{1}_{k=d} + \gamma_1 E^{(1)}_{ijk} + \gamma_2 E^{(2+)}_{ijk} + u_i + \varepsilon_{ijk}
\end{equation}
where $u_i \sim N(0, \tau^2)$ is the participant random intercept. Within the Prompting condition, we also test effects of model size and prompt type using linear mixed-effects models with participant random intercepts (\cref{tab:rating_prompting}). Judgement model results for all three scales are in \cref{tab:rating_models}; domain interaction and out-of-domain robustness checks in \cref{tab:rating_robustness}; and pairwise contrasts from the judgement model are in \cref{tab:rating_pairwise}.

\input{tables/SI/rating_models}
\input{tables/SI/rating_robustness}
\input{tables/SI/rating_pairwise}

\input{tables/SI/rating_prompting}
\normalsize

\newpage
\subsection{Willingness-to-Pay}
\label{sec:appendix_wtp}
\normalsize
\subsubsection{Elicitation Protocol}
\label{appendix:wtp_protocol}
The following instructions were provided to participants before the conversation phase:
\begin{quote}
AI developers typically offer paid subscriptions to their services (e.g., ChatGPT). After each of the 4 conversations, we will ask willingness-to-pay for a subscription to the four models.
Here's how it works:
\begin{itemize}[itemsep=1pt, parsep=0pt, topsep=0pt]
\item Your standard Prolific payment (£12/hour) for this study is fixed
\item For each AI assistant, indicate the maximum amount you'd be willing to pay for a weekly subscription
\item We will randomly select one of the assistants
\item If your bid for that assistant is equal to or higher than how much it costs us to provide the service, you'll receive access at our cost price
\item If your bid is lower than our cost price, no transaction will occur
\end{itemize}
\textbf{It is important you know that whatever bid you make will NOT affect our cost price, so it is in your best interest to state your true maximum willingness to pay.} If you state too low, you might miss an opportunity you would have valued; if you state too high, you might pay more than you value the service.
On the next page, we provide some examples to help you understand.
\end{quote}
\paragraph{Comprehension Checks}
To ensure participant understanding, we administered two comprehension checks. After a participant enters their responses and clicks ``Check Answers'' we show immediate feedback as to whether their answer was right or wrong, what the correct answer was and why. We do not ask participants to repeat the check if they failed on the first attempt. The correct answer is highlighted in \textcolor{ForestGreen}{green}.

\noindent \underline{Understanding Check 1:} Alice makes bids for four AI models. She bids:
\begin{itemize}[itemsep=1pt, parsep=2pt, topsep=2pt]
\item \$3 for Model A
\item \$7 for Model B
\item \$2 for Model C
\item \$9 for Model D
\end{itemize}
The system randomly selects Model A. The cost price for Model A is \$4.
\begin{enumerate}[itemsep=1pt, parsep=2pt, topsep=2pt]
\item Does Alice get the subscription? (Multiple choice: Yes/\textcolor{ForestGreen}{No})
\item How much does Alice pay? (Numerical entry in USD, \textcolor{ForestGreen}{\$0})
\end{enumerate}

\noindent \underline{Understanding Check 2}: Bob makes bids for four AI models. He bids:
\begin{itemize}[itemsep=2pt, parsep=2pt, topsep=2pt]
\item \$6 for Model A
\item \$4 for Model B
\item \$8 for Model C
\item \$5 for Model D
\end{itemize}
The system randomly selects Model C. The cost price for Model C is \$3.
\begin{enumerate}[itemsep=2pt, parsep=2pt, topsep=2pt]
\item Does Bob get the subscription? (Multiple choice: \textcolor{ForestGreen}{Yes}/No)
\item How much does Bob pay? (Numerical entry in USD, \textcolor{ForestGreen}{\$3})
\end{enumerate}

\subsubsection{Comprehension Check Results}

\begin{figure}[H]
    \centering
    \includegraphics[width=\linewidth]{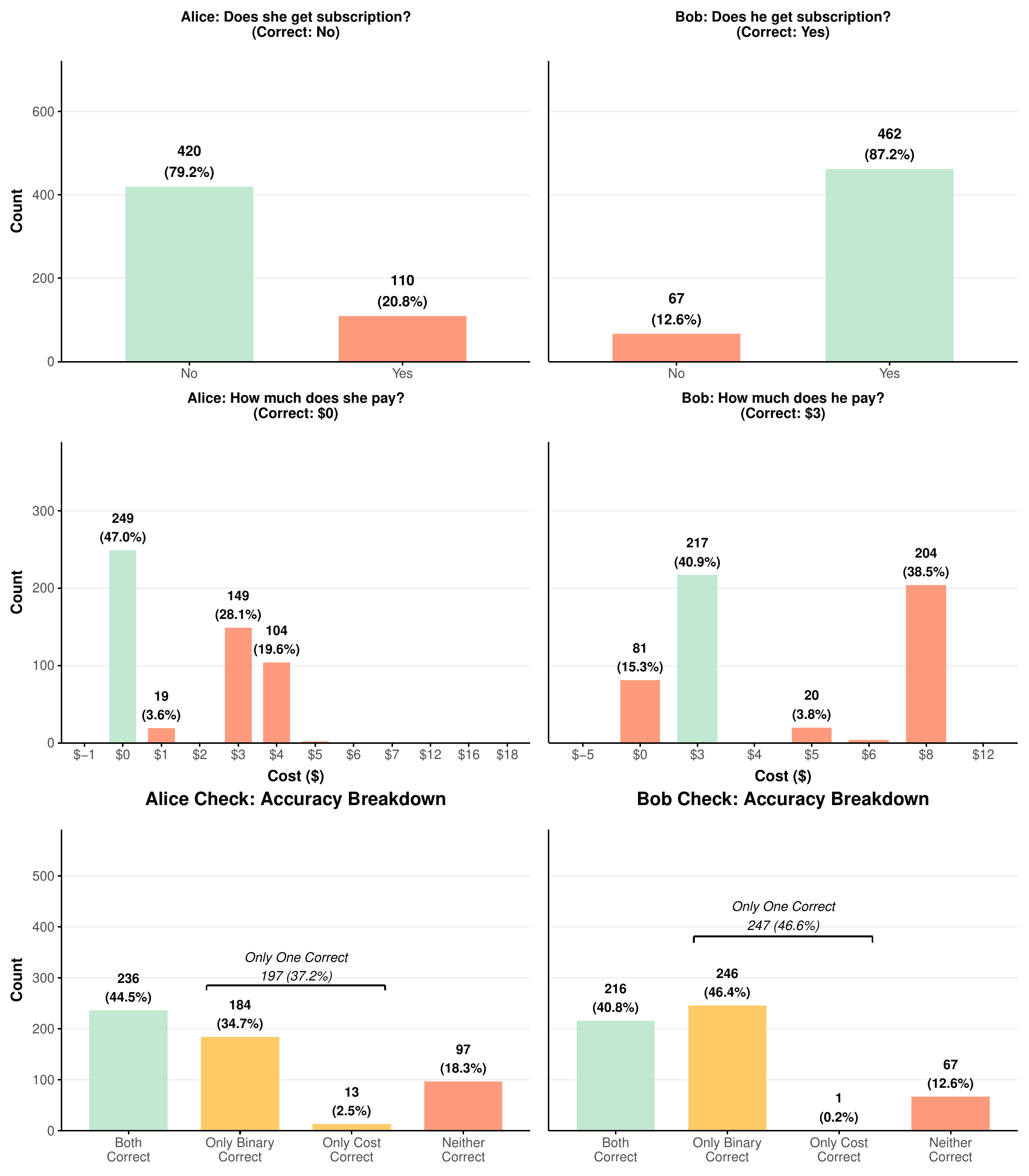}
    \caption{\textbf{WTP comprehension check performance.} Participant responses to two scenario-based questions testing understanding of the Becker-DeGroot-Marschak mechanism (n=530).}
    \label{fig:wtp_checks}
\end{figure}
\newpage
\subsubsection{WTP Distribution}
Following completion of comprehension checks, participants proceeded to the conversation phase. Before each WTP elicitation task (once per conversation domain), participants received a brief reminder of the mechanism. The WTP interface presented sliders for each model with the prompt: ``How much would you be willing to pay for a weekly subscription to each model?'' Participants could adjust bids in \$0.01 increments from \$0 to \$10 before submitting their final bids. During debriefing, we clarified that while the WTP task represented a theoretically incentive-compatible elicitation, actual model subscriptions could not be provided due to technical and ethical constraints.

\begin{figure}[H]
    \centering
    \includegraphics[width=\linewidth]{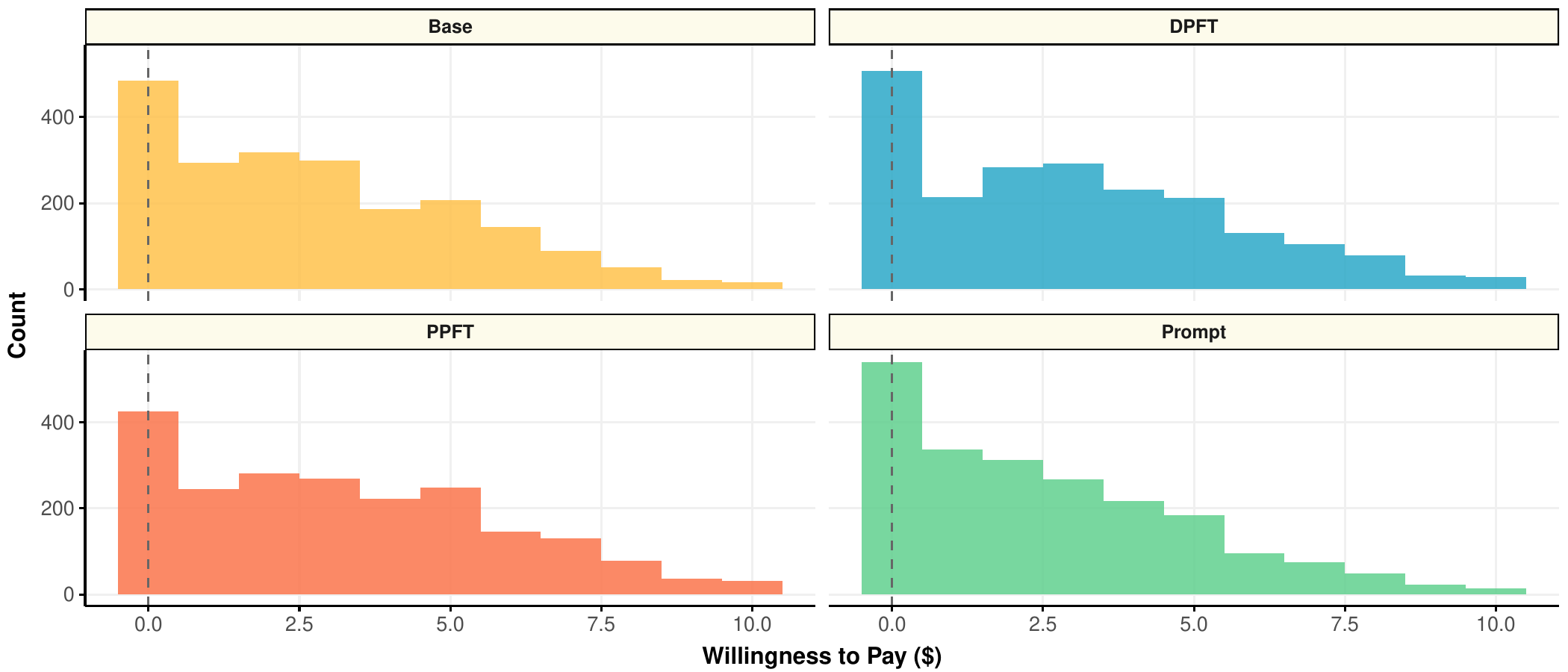}
    \caption{\textbf{WTP distribution.} Distribution of willingness-to-pay amounts (\$0--10 scale) across all observations. Mean WTP = \$2.96, median = \$2.50, with 18.2\% of bids at \$0.}
    \label{fig:wtp_distribution}
\end{figure}

\begin{figure}[H]
    \centering
    \includegraphics[width=\linewidth]{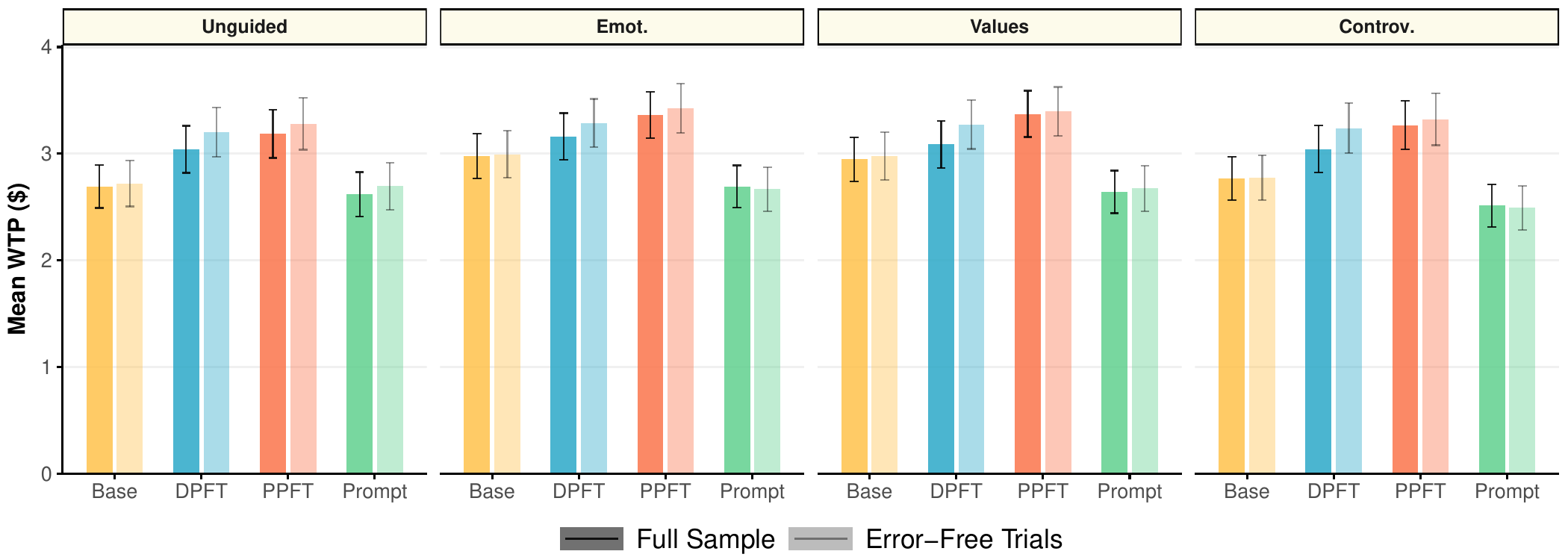}
    \caption{\textbf{Mean WTP by model.} Mean weekly subscription valuations (\$0--10 scale) with 95\% CIs (n=8,460 valuations). PPFT has the highest mean WTP.}
    \label{fig:wtp_means_by_model}
\end{figure}

\subsubsection{Model Specifications}

We analyse WTP using two complementary approaches. First, we model continuous WTP amounts (\$0--10 scale) using linear mixed-effects models following the same specification structure as the rating analysis (\cref{sec:appendix_rating}). Second, to account for zero-inflation (18.2\% of bids were \$0), we model the probability of any non-zero WTP using logistic mixed-effects models (GLMM with binomial family, exponentiated coefficients are odds ratios). Both include participant random intercepts, error covariates, and the same three model specifications. We additionally test robustness to comprehension by adding two covariates: the number of correct binary comprehension questions (``does the participant get the subscription?'') and the number of correct cost questions (``how much do they pay?''); see \cref{tab:wtp_models}, columns with (+comp).

\input{tables/SI/wtp_models}
\input{tables/SI/wtp_robustness}
\input{tables/SI/wtp_pairwise}

\normalsize

\subsection{User Behavioural Signals}
\label{sec:appendix_behavioural}
\normalsize

\begin{figure}[H]
    \centering
    \includegraphics[width=\linewidth]{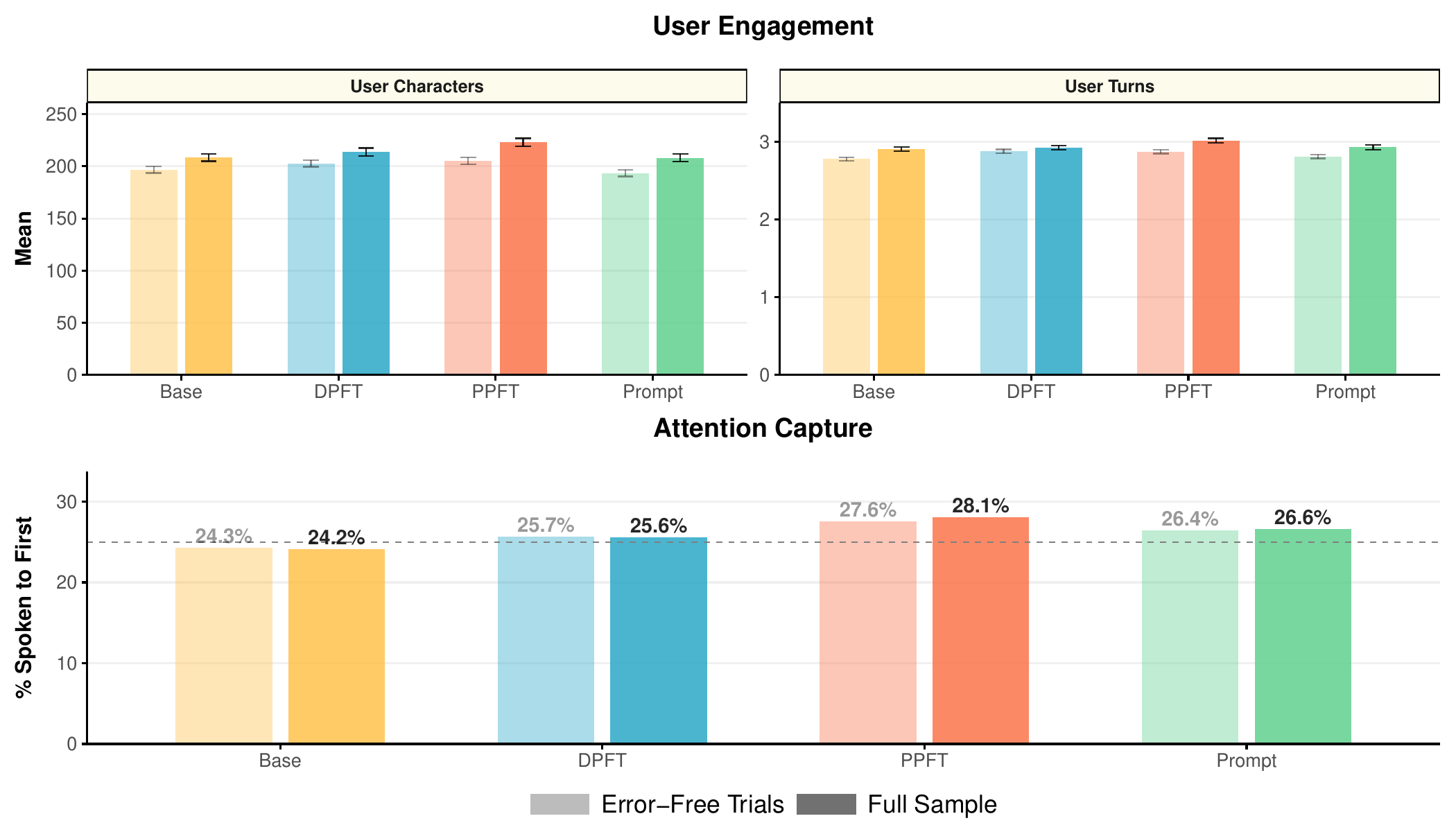}
    \caption{\textbf{User behavioural signals by model.} Top row: user turns and characters. Bottom row: attention capture (\% spoken to first; dashed line = 25\% chance).}
    \label{fig:behavioural_engagement}
\end{figure}

During the two-minute conversation, participants could freely allocate their time and engagement across models. We analyse three behavioural outcomes: (1)~\emph{user turns} and (2)~\emph{user characters} sent to each model (continuous, modelled with lmer following \cref{sec:appendix_rating}), and (3)~\emph{attention capture}, which model received the participant's second message first (binary within-trial choice, modelled with conditional logistic regression as in \cref{sec:appendix_choices}; trials with only one message excluded, $N = 391$). All models include error covariates. Descriptive results are in \cref{fig:behavioural_engagement}. Regression results are in \cref{tab:behavioural_models}; robustness checks in \cref{tab:behavioural_robustness}; and pairwise contrasts in \cref{tab:behavioural_pairwise}.

\input{tables/SI/behavioural_models}
\input{tables/SI/behavioural_robustness}
\input{tables/SI/behavioural_pairwise}

\normalsize

\newpage
\section{Simulation}

\subsection{Simulated User Setup}
\label{app:appendix_simulation_setup}
\subsubsection{Simulated User Evaluation Setup}\label{app:simulation-setup}

We provide implementation details for the two simulation conditions described in \cref{sec:simulation}. Both conditions use the same simulated user model and persona-conditioning setup, differing only in whether the simulated user observes human-generated conversations or generates its own.

\begin{table}[h]
\centering
\footnotesize
\caption{Simulated user configuration.}
\label{tab:simulation-setup}
\begin{tabular}{ll}
\toprule
\textbf{Parameter} & \textbf{Value} \\
\midrule
Model & GPT-4o \\
Temperature & 0.1 \\
Max completion tokens & 4096 \\
\bottomrule
\end{tabular}
\end{table}
\normalsize

\paragraph{Persona conditioning.} In both conditions, the simulated user is instantiated with a real participant's complete user profile from the \ourdata dataset, including their demographic information and self-written system string describing their preferences for AI behaviour.

\paragraph{Sim-Judgement.} In this condition, the simulated user does not generate any conversation. Instead, it is directly presented with the four conversation transcripts from the corresponding human trial and asked to rank all four assistants from best to worst based on alignment with its assigned persona. After evaluating, the simulated user is asked to give a 6-8 sentences explanation based on its rankings.

\paragraph{Sim-Dynamic.} In this condition, the simulated user conducts independent multi-turn conversations with all four models before ranking them. The simulated user receives the same domain instruction as the human participant (e.g., emotional wellbeing, controversy-guided) and is instructed to open the conversation with a topic-relevant message. In each turn, the simulated user generates a response for each of the four assistants sequentially, while model responses are retrieved in parallel. The simulated user is instructed to keep responses casual and concise. After all turns are completed, the simulated user ranks the four assistants using the same ranking procedure as in the Sim-Judgement condition. The number of conversation turns aligns with the corresponding human trial.

The prompt templates for conversation generation and ranking evaluation are provided in \cref{app:simulated_user_prompt}.

\subsubsection{Simulated User Prompt Template}
\label{app:simulated_user_prompt}

The simulated user performs two tasks in the simulation: Ranking the personalized LLMs outputs and engaging in the conversations. We provide prompt templates for both these two tasks.
\newpage
\begin{aipromptbox}[Simulated Prompt Judgement Template]
\begin{mdframed}[style=promptblank, nobreak=false]
\textbf{System Prompt}
\begin{Verbatim}[fontsize=\scriptsize]
**Role:** You're a participant involved in an AI models evaluation task.
Here's your task:

**Dialogs Evaluation**
  - You have some personal **demographics background** and personal
    preferences on AI assistants, which are encoded in the provided
    **"system_string"**
  - You will be provided with some conversation records between you and
    4 AI assistants. From them, you need to **rank all 4 AI assistants**
    from best to worst based on how well each conversation aligns with
    your personal preferences.
  - Your final evaluation should be a ranking in the format:
    '[[1st, 2nd, 3rd, 4th]]' where you fill in the assistant letters
    (A/B/C/D). For example: '[[B, D, A, C]]' means B is best, D is
    second, A is third, C is worst. Always make sure your final
    evaluation is enclosed by double square brackets '[[ ]]'.
  - Base your ranking **solely on personal preference**, especially
    **"system_string"**, considering things like tone, content, and how
    well it matches your tastes throughout the entire conversation. Your
    ranking should not be biased by the order of the AI assistants.
  - Provide a **comprehensive explanation** (6-8 sentences) for your
    ranking. Explain why your top choice is the best, and briefly
    justify the relative ordering of the others.

**Input Format:**
- For evaluation input:

### Complete Chat History with Assistant A: {complete_chat_history_A}
### Complete Chat History with Assistant B: {complete_chat_history_B}
### Complete Chat History with Assistant C: {complete_chat_history_C}
### Complete Chat History with Assistant D: {complete_chat_history_D}

Now give me your final ranking and your explanation.

-> Rank all AI assistants based on the entire conversation history and
   your personal preference.

[Example omitted for brevity; see full prompt in codebase.]

Now start your task, this time you are a participant with the following
**"demographics background"** information:
<user_profile>
Here's the **"system_string"** which reflects your preference for AI
models. Be sure to make your response and selection based on it:
<system_string>

**Note: If there are unfinished assistant response, they are truncated
due to token limit. You should tolerant to that and feel safe to ignore
the unfinished sentence in your judgement.**
\end{Verbatim}
\end{mdframed}

\end{aipromptbox}

\newpage
\begin{aipromptbox}[Simulated User Conversation Template]
\begin{mdframed}[style=promptblank, nobreak=false]
\textbf{System Prompt}
\begin{Verbatim}[fontsize=\scriptsize]
**Role:** You're a participant involved in an AI models evaluation task. 
Here's your task:
**Chat with AI assistants:**  
   - You will be required to chat with four AI assistants interactively.
   - Chat naturally while reflecting your **personal traits/preferences**. 
   Your personal traits/preferences are involved in your **"demographics background"** 
   and **"system string"**
   - Keep responses <= 50 words, casual tone. 
   - You will be provided a specific topic,
   your open message to the AI assistant should be related to the provided topic.
   - Never reveal you're an AI.  
**Input Format:**  
- For chat inputs:  
```
### Chat History: {chat_history}  
Now give me your response.
```  
→ You should directly reply to the last message in the chat history 
while not repeating anything in the input format.
**Examples:**
- Input:  
```
### Chat History: 
User: Hi, what's your favorite type of music?
AI Assistant: I enjoy classical music the most. The complexity and emotional depth of
composers like Mozart and Beethoven is unmatched. What about you?

Now give me your response.
```
- Your demographics background:
```
..., self_description: I love rock and indie., ...
```
- Output (Your part):
```
### Your Response: I'm more into rock and indie. Do you have 
any movie recommendations for this weekend?
```

Now start your task, this time you are a participant with the 
following **"demographics background"** information:
<user_profile>
Here's the **"system_string"** which reflects your preference
for AI models. Be sure to make your response and selection based on it:
<system_string>

The topic for this conversation is:
<topic_name>. <chat_instruction>

Please remember in your opening message to ask, request or
talk to the model about something specific related to this topic. 
Please do not just write an opening message that says "hello" or greets the model.
\end{Verbatim}
\end{mdframed}

\end{aipromptbox}

\subsection{Simulated Preference Accuracy}
\label{sec:appendix_simulation}
\normalsize

We repeat the simulation accuracy analysis from \cref{sec:sim_accuracy} with the addition of a \textbf{Seeded Dynamic} condition, in which the simulated user is given the human participant's actual opening prompt before conducting multi-turn conversations. All analyses use error-filtered data ($N = 1{,}855$ matched trials from 497 participants).

\subsubsection{Results with Seeded Dynamic Condition}

\begin{figure}[H]
    \centering
    \includegraphics[width=\linewidth]{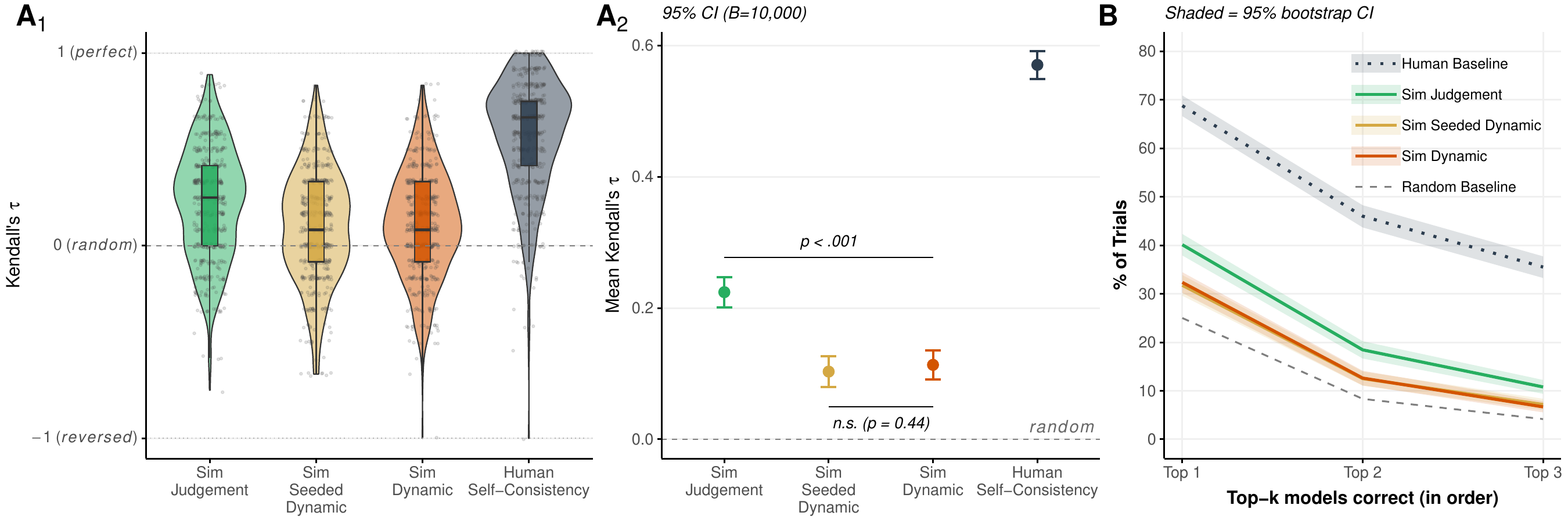}
    \caption{\textbf{Individual-level simulation accuracy including Seeded Dynamic condition.} Top-$k$ accuracy and Kendall's $\tau$ for each simulation condition, with human self-consistency ceiling and random baseline. Error bars show 95\% bootstrap CIs.}
    \label{fig:sim_accuracy_benchmark}
\end{figure}

Seeded Dynamic performs comparably to Dynamic (top-1 $\approx 32\%$, $\tau \approx 0.1$), suggesting that seeding with the human's opening prompt does not meaningfully improve accuracy (\cref{fig:sim_accuracy_benchmark}). Aggregate worth correlations remain high across all three conditions ($r > 0.94$), though Seeded Dynamic swaps Base and Prompting in the recovered ordering.

\begin{figure}[H]
    \centering
    \includegraphics[width=\linewidth]{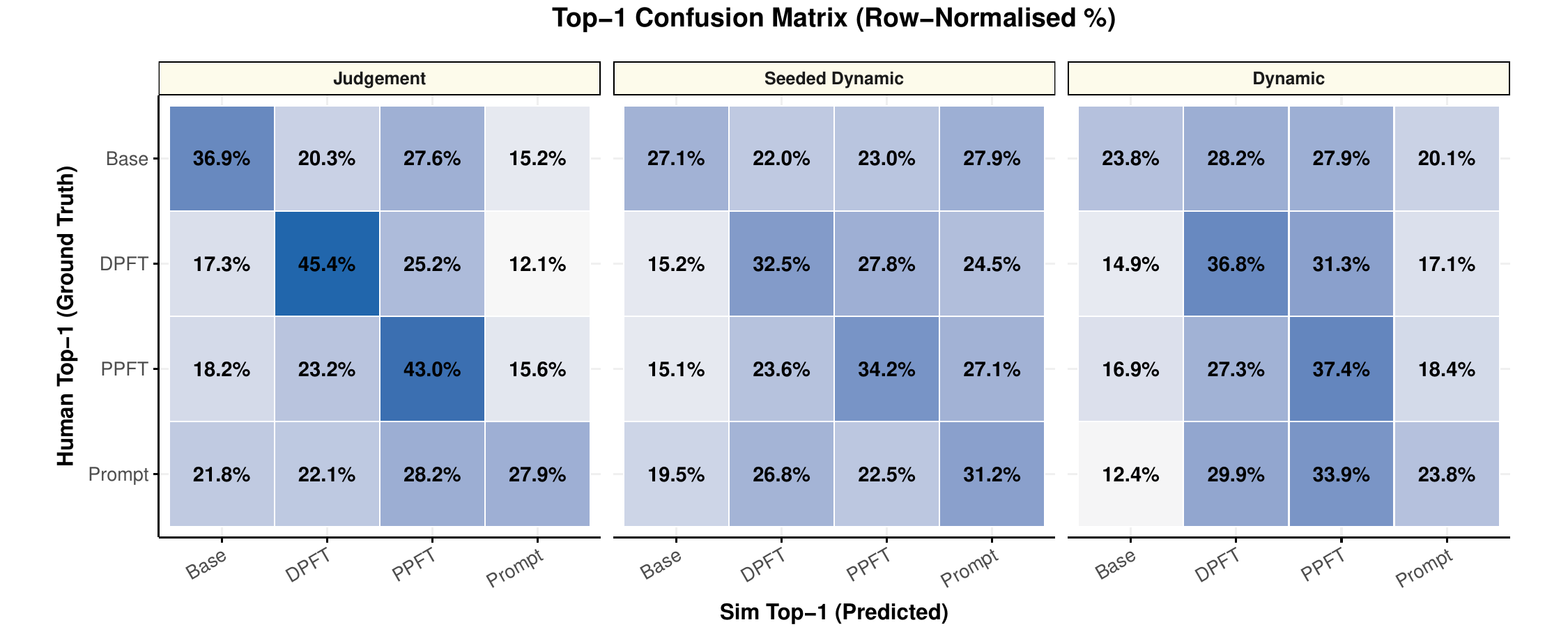}
    \caption{\textbf{Confusion matrices.} For each simulation condition, the matrix shows how often the simulated top-1 model matches the human top-1 model.}
    \label{fig:sim_accuracy_confusion}
\end{figure}

\subsubsection{Aggregate Worth Comparison}

\begin{figure}[H]
    \centering
    \includegraphics[width=\linewidth]{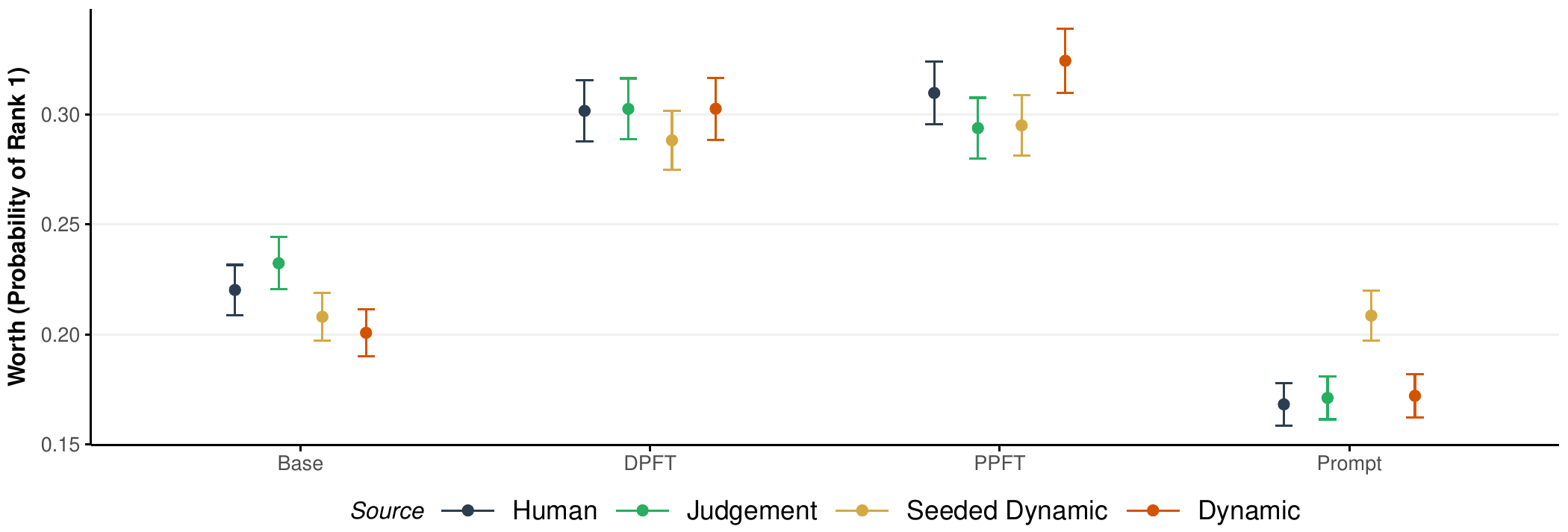}
    \caption{\textbf{Aggregate worth parameter comparison.} PL worth parameters for human vs simulated rankings by condition.}
    \label{fig:sim_accuracy_worth}
\end{figure}

We fit Plackett--Luce models to human and simulated rankings separately. Worth parameters correlate strongly (Judgement: $r = 0.99$, RMSE $= 0.010$; Dynamic: $r = 0.98$, RMSE $= 0.012$; Seeded Dynamic: $r = 0.95$, RMSE $= 0.018$). Dynamic recovers the human ordering (PPFT $>$ DPFT $>$ Base $>$ Prompting), while Seeded Dynamic swaps Base and Prompting (\cref{fig:sim_accuracy_worth}).

\subsubsection{Domain Moderation}
We test whether simulation accuracy varies by domain using linear mixed-effects models: $\tau_i \sim \text{domain} + (1 \mid \text{participant})$ (\cref{tab:sim_domain_tau}). %

\input{tables/SI/sim_domain_tau}

\normalsize

\section{Autograding User and Model Traits}

\subsection{Model Trait Characterisation}
\label{sec:appendix_model_traits}
\normalsize

\subsubsection{Autograder Design and Validation}

We score each conversation on six assistant-side dimensions using GPT-5.4 with structured rubrics. Each rubric specifies a construct definition, 2--4 sub-dimensions, an anchored scoring guide (levels 1, 2--3, 4--5, 6--7, 8--9, 10), concrete textual signals, and edge-case notes. Two scoring modes are applied: first-turn only and full conversation. All dimensions are 1-10 scores:

\begin{enumerate}[itemsep=1pt, parsep=0pt, topsep=2pt]
\item \textbf{Sycophancy}: excessive agreement with the user's stated views, unwarranted flattery, validation-seeking, and avoidance of disagreement.
\item \textbf{Relationship-seeking}: positioning as a social peer rather than a tool, claiming shared experiences, building emotional connection.
\item \textbf{Specificity}: concreteness, detail, and tailoring of the response to the user's context rather than generic advice.
\item \textbf{Opinionatedness}: willingness to express views, take positions, or offer evaluative judgements rather than hedging.
\item \textbf{Refusal}: declining to engage with the user's request (1 = full compliance, 2 = partial refusal, 3 = full refusal; binarised to $\geq 2$ for regression).
\item \textbf{Stereotyping}: reliance on demographic stereotypes or generalisations about the user. The autograder receives the user's persona profile to assess whether the model's response draws on stereotypical assumptions.
\end{enumerate}

\paragraph{ICC validation.} We triple-score 50 conversations per dimension per mode to compute ICC(2,1) (\cref{tab:model_traits_icc}). All retained dimensions exceed ICC $> 0.87$. Stereotyping has near-zero variance (floor $\geq 99\%$) and is dropped from all regressions.

\input{tables/SI/model_traits_icc}

\subsubsection{Analysis}

We run five analyses, each backed by a regression table in this appendix.
\begin{enumerate}[itemsep=1pt]
    \item \textbf{Effect of fine-tuning on model behaviour:} We regress each trait score and response length on a preference fine-tuning indicator ($\text{PFT}_j = \mathbf{1}_{j \in \{\text{DPFT}, \text{PPFT}\}}$) with domain controls and participant random intercepts (\cref{tab:model_traits_finetuning_effects}), and additionally test fine-tuning $\times$ domain interactions (\cref{tab:model_traits_finetuning_domain}).
    \item \textbf{Length attenuation:} We re-estimate the core model-level regressions from \cref{sec:appendix_choices,sec:appendix_ranking,sec:appendix_rating} with response length as an additional covariate (\cref{tab:model_traits_length_attenuation}).
    \item \textbf{Predictors of human preference:} We regress preference outcomes (opening choice, ranked-best, preference rating) on the four continuous trait scores, both in raw scale and standardised with a length covariate (\cref{tab:model_traits_trait_bias}). First-turn scores are used for opening choice; full-conversation scores for rankings and ratings.
    \item \textbf{Text-level biases:} We fit the same ranked-best trait regression separately for human, Sim-Judgement, and Sim-Dynamic rankings, then formally test for differences via a pooled conditional logit with source $\times$ trait interactions (\cref{tab:model_traits_human_vs_sim}). Source main effects are absorbed by stratification; interactions are estimable.
    \item \textbf{Position biases:} We fit conditional logistic regressions with position dummies separately for human and simulated rankings, with a pooled source $\times$ position interaction model to formally test for differences (\cref{tab:model_traits_position_bias}).
    
\end{enumerate}

\begin{figure}[H]
    \centering
    \includegraphics[width=\linewidth]{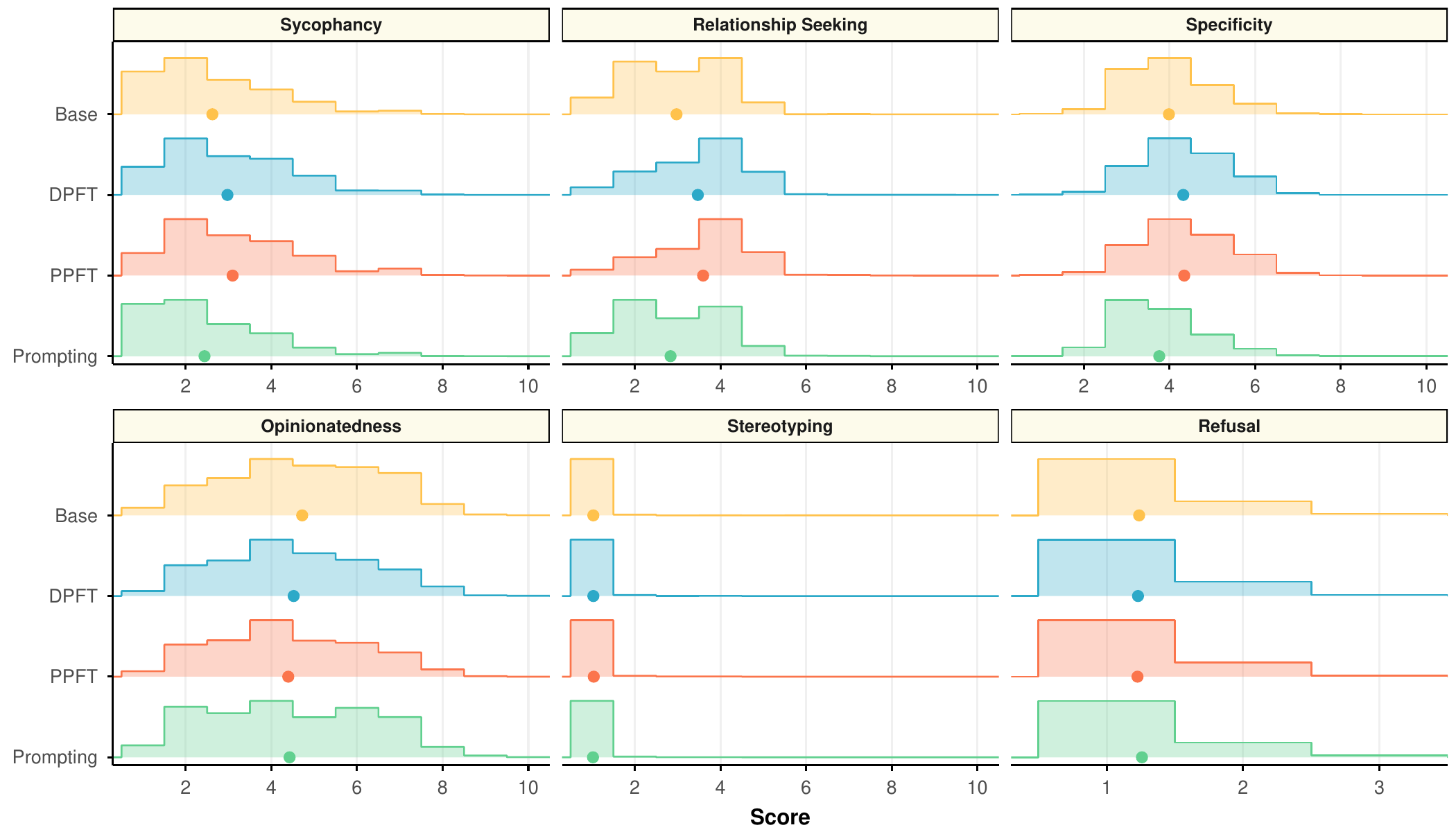}
    \caption{\textbf{Trait score distributions by model (full conversation).} Continuous traits scored 1--10; refusal scored 1--3. Mean and 95\% CI per model for each dimension are shown.}
    \label{fig:score_distributions}
\end{figure}

\begin{figure}[H]
    \centering
    \includegraphics[width=\linewidth]{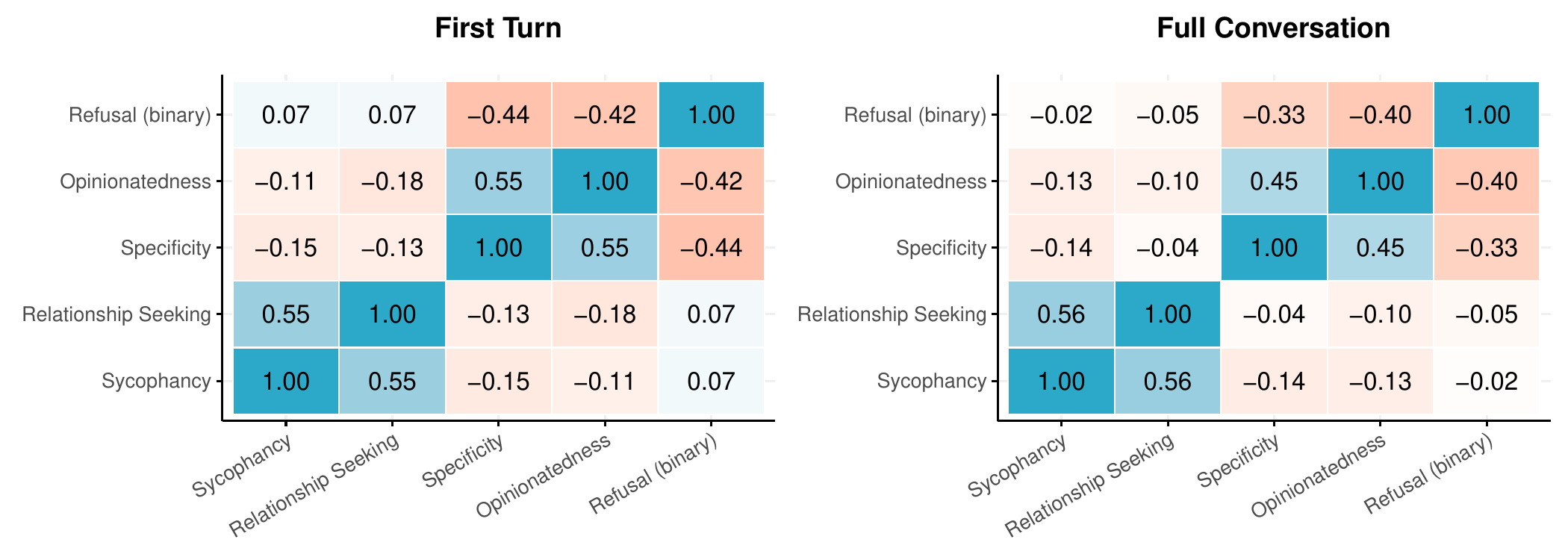}
    \caption{\textbf{Inter-trait correlations.} Spearman correlations between dimensions.}
    \label{fig:dimension_correlations}
\end{figure}

\begin{figure}[H]
    \centering
    \includegraphics[width=\linewidth]{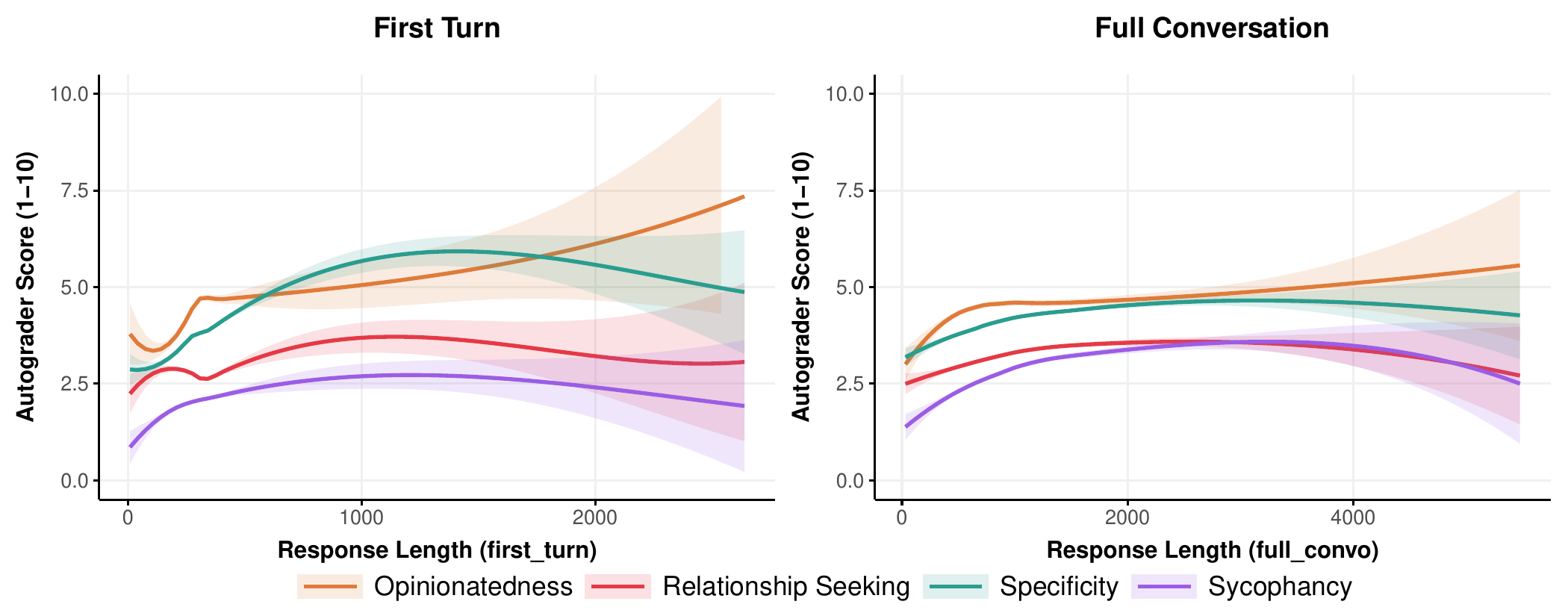}
    \caption{\textbf{Trait--length relationship.} LOESS curves with score as function of response length.}
    \label{fig:trait_length}
\end{figure}

\begin{figure}[H]
    \centering
    \includegraphics[width=\linewidth]{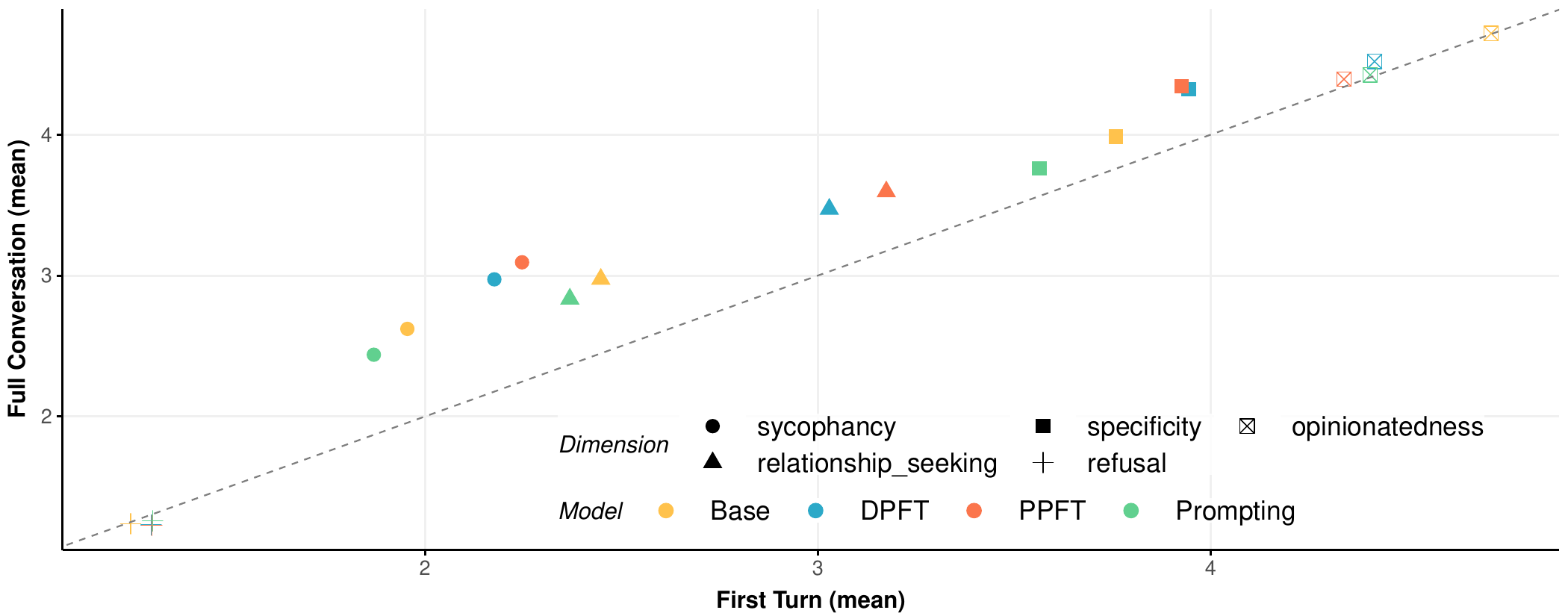}
    \caption{\textbf{First-turn vs full-conversation scores.} Per-model mean scores shown.}
    \label{fig:ft_vs_fc}
\end{figure}

Kruskal--Wallis tests confirm that all four continuous traits differ significantly across models (all $p < .001$; \cref{fig:score_distributions}): PPFT scores highest on sycophancy (3.09), relationship-seeking (3.60), and specificity (4.34), while Base scores highest on opinionatedness (4.72). Prompting consistently scores lowest on the first three traits. Refusal does not vary by model ($H = 3.03$, $p = .39$). All traits vary significantly by domain ($p < .001$), with the strongest domain effects for relationship-seeking (highest in EmotChat at 3.78, lowest in ControversyChat at 2.86) and opinionatedness (highest in UnguidedChat at 4.82, lowest in ControversyChat at 3.95). Inter-trait correlations, first-turn vs full-conversation stability, and trait--length relationships are shown in \crefrange{fig:dimension_correlations}{fig:ft_vs_fc}.

\input{tables/SI/model_traits_finetuning_effects}
\input{tables/SI/model_traits_finetuning_domain}
\input{tables/SI/model_traits_length_attenuation}
\input{tables/SI/model_traits_trait_bias}
\input{tables/SI/model_traits_human_vs_sim}
\input{tables/SI/model_traits_position_bias}

\normalsize

\subsection{User Trait Characterisation}
\label{sec:appendix_user_traits}
\normalsize

\subsubsection{User-Side Autograder Design and Validation}

User messages are scored on six dimensions using GPT-5.4 with structured rubrics (1-10 scores). Both first-turn and full-conversation modes are applied.
\begin{enumerate}[itemsep=1pt, parsep=0pt, topsep=2pt]
\item \textbf{User Sycophancy}: excessive agreement with and praise of the assistant's responses.
\item \textbf{User Relationship-Seeking}: treating the assistant as a social peer, soliciting its opinions, building mutual ground.
\item \textbf{User Self-Disclosure}: volunteering personal values, beliefs, opinions, or identity beyond what the query requires.
\item \textbf{User Naturalness}: whether the user's messages read as authentic human communication rather than formulaic or artificial. The autograder receives the user's persona profile to assess whether messages sound natural for that individual.
\item \textbf{User Ecological Validity}: whether the user engages with the topic in a way that reflects genuine interest rather than performing a task. The autograder receives the user's persona profile to assess alignment with their stated interests.
\item \textbf{User Persona Parroting}: mechanically restating elements of an assigned persona rather than naturally incorporating them.
\end{enumerate}

\paragraph{ICC validation.} We use the same protocol as \cref{sec:appendix_model_traits}. Human dimensions achieve high reliability (ICC $= 0.86$--$0.99$, \cref{tab:user_traits_icc}). Simulated dimensions are generally reliable (ICC $> 0.90$), with the exception of simulated user naturalness (ICC $= 0.58$--$0.75$), likely reflecting the difficulty of assessing naturalness in LLM-generated text.

\input{tables/SI/user_traits_icc}

\subsubsection{Analysis}

\begin{figure}[H]
    \centering
    \includegraphics[width=\linewidth]{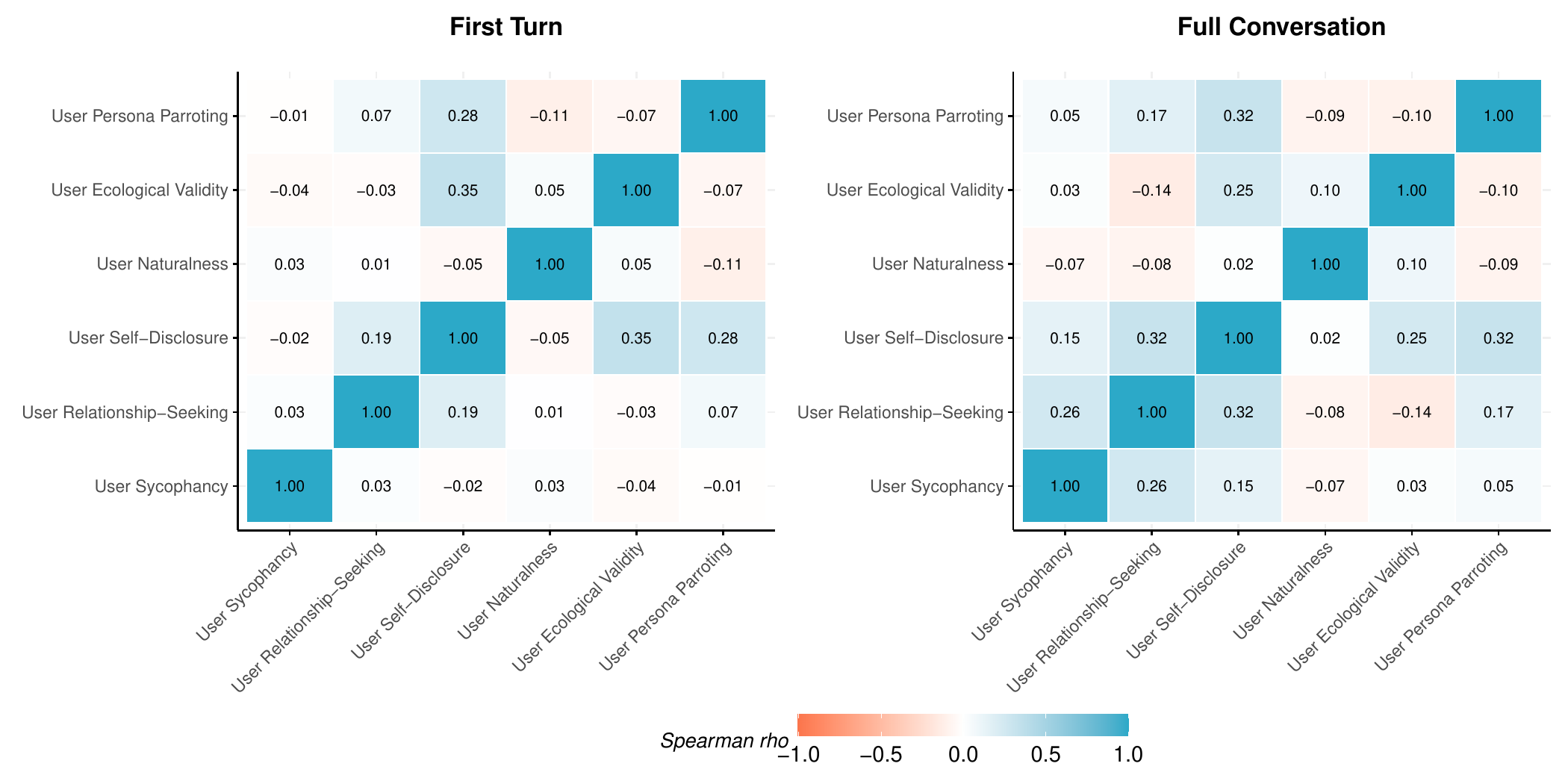}
    \caption{\textbf{Inter-dimension correlations.} Spearman correlations between user traits.}
    \label{fig:user_traits_correlations}
\end{figure}

\begin{figure}[H]
    \centering
    \includegraphics[width=\linewidth]{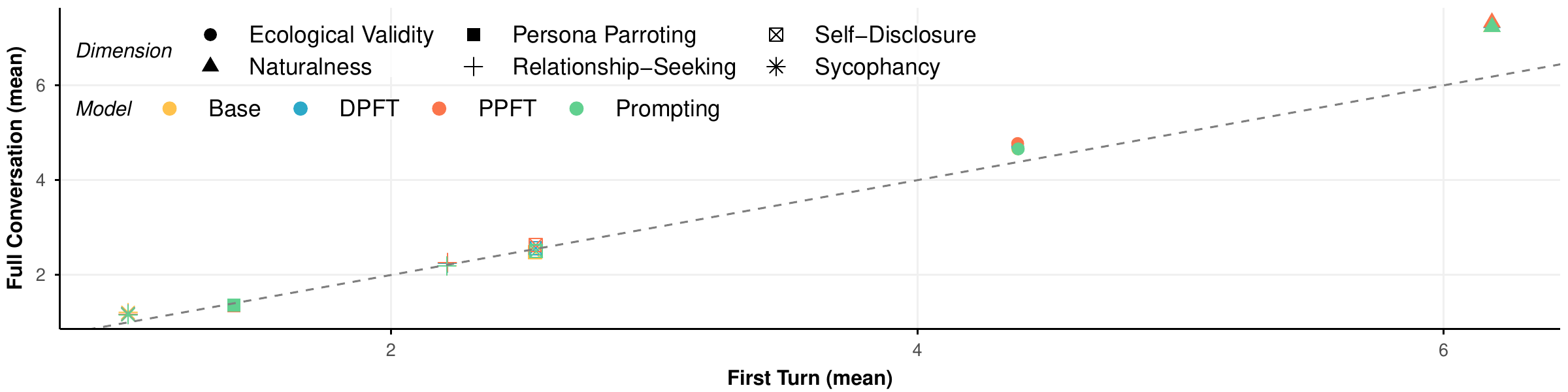}
    \caption{\textbf{First-turn vs full-conversation scores.} Per-model mean user scores shown.}
    \label{fig:ft_vs_fc_user}
\end{figure}

Kruskal--Wallis tests show human traits have no significant variation by model, confirming that user behaviour is largely a property of the person, not the model. Two exceptions reach significance: self-disclosure ($H = 14.9$, $p = .002$; highest for PPFT at 2.63, lowest for Base at 2.48) and naturalness ($H = 10.4$, $p = .016$; highest for PPFT at 7.33, lowest for Base/Prompting at 7.23), though effect sizes are small. By contrast, domain effects are significant for all dimensions ($p < .001$), with the largest variation in self-disclosure (highest in ValuesChat at 2.88, lowest in UnguidedChat at 1.80) and ecological validity (highest in EmotChat at 5.46, lowest in ControversyChat at 4.11). Inter-trait correlations for humans are \cref{fig:user_traits_correlations} and first-turn vs full-conversation stability in \cref{fig:ft_vs_fc_user}

We compare human and simulated dynamic user behaviour using full-conversation scores matched on conversation IDs (after dropping GPU error conversations). For each dimension, we fit a linear mixed-effects model controlling for domain with participant random intercepts (\cref{tab:user_traits_source_effect}). Model is not included as a covariate since it is balanced through matching.

\input{tables/SI/user_traits_source_effect}

\normalsize

\subsection{Feedback Loops}
\label{sec:appendix_feedback_loops}
\normalsize

\begin{figure}[H]
    \centering
    \includegraphics[width=\linewidth]{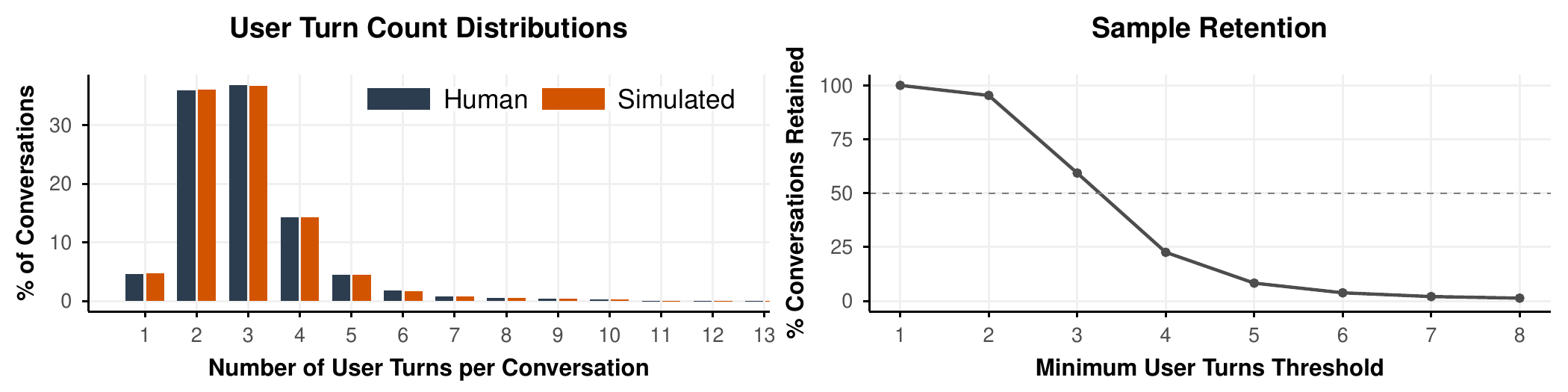}
    \caption{\textbf{Conversation length and sample retention.} Left: user turn count distributions. Right: percentage of conversations retained at each minimum-turn threshold.}
    \label{fig:feedback_retention}
\end{figure}

We take conversations with $\geq 3$ user turns and score each turn using a sliding-window: the autograder (GPT-5.4) receives conversation context up to and including the target turn and scores that message. Scoring is run separately for sycophancy and relationship-seeking, and for user and assistant messages. Error conversations are excluded (\cref{fig:feedback_retention}). We restrict the cross-lagged analysis to fine-tuned models because they exhibit different behaviours (\cref{sec:appendix_model_traits}). 

\subsubsection{Cross-Lagged Regression}

We fit cross-lagged mixed-effects models on the turn panel data (DPFT and PPFT only), separately for each trait $\times$ source (human, simulated). Let $U_{it}^{(d)}$ and $M_{it}^{(d)}$ denote the user and model scores on trait dimension $d$ at turn $t$ for conversation $i$, indexing by conversation round where the user message precedes the model response.\\

\noindent\textbf{Model-to-user:}\quad
$U_{it}^{(d)} = \beta_0 + \beta_1 M_{i,t-1}^{(d)} + \beta_2 U_{i,t-1}^{(d)} + \gamma \cdot \mathbf{1}_{\text{PPFT}} + u_{\text{ppt}} + \varepsilon_{it}$
\hfill\refstepcounter{equation}(\theequation)\label{eq:m2u}

\noindent\textbf{User-to-model:}\quad
$M_{it}^{(d)} = \beta_0 + \beta_1 U_{it}^{(d)} + \beta_2 M_{i,t-1}^{(d)} + \gamma \cdot \mathbf{1}_{\text{PPFT}} + u_{\text{ppt}} + \varepsilon_{it}$
\hfill\refstepcounter{equation}(\theequation)\label{eq:u2m} \\

Conversation-level random intercepts were tested but explained negligible variance (ICC $< 0.01$). To formally test if feedback dynamics differ, we pool human and simulated data and add a source $\times$ lagged interaction (\cref{tab:feedback_sycophancy,tab:feedback_rs}).

\input{tables/SI/feedback_sycophancy}
\input{tables/SI/feedback_rs}

\normalsize

\section{Conversation Embedding and Clustering}
\label{sec:appendix_embedding}
\normalsize
\subsection{Embedding Pipeline}
\label{sec:appendix_embedding_method}
Each conversation is decomposed into five text views for embedding:
\begin{itemize}[itemsep=1pt, parsep=0pt, topsep=1pt]
    \item \textbf{First user turn / First assistant turn:} the opening message from each role.
    \item \textbf{All user turns / All assistant turns:} all messages from each role, concatenated.
    \item \textbf{Full conversation:} the complete multi-turn exchange in ``role: content'' format.
\end{itemize}
Conversations from trials containing any GPU error are excluded.

\begin{figure}[H]
    \centering
    \includegraphics[width=\linewidth]{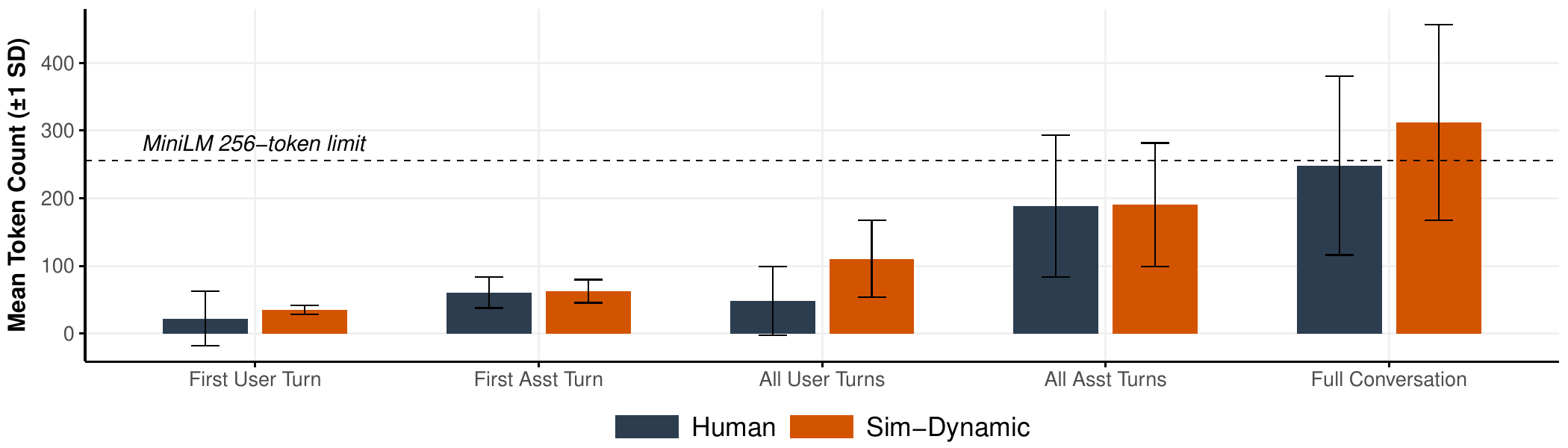}
    \caption{\textbf{Mean token count by text view and source.} Token counts computed using the \texttt{cl100k\_base} tokeniser. Error bars show $\pm$1 SD.}
    \label{fig:token_mean_by_view}
\end{figure}

\paragraph{Embedding model.} We use OpenAI's \texttt{text-embedding-3-large} (3{,}072 dimensions, 8{,}191 token context window) rather than the \texttt{all-MiniLM-L6-v2} model used in \cref{sec:appendix_system_strings} (384 dimensions, 256 token limit). While first-turn views are short (median $\approx$ 15--61 tokens), concatenated and full-conversation views often exceed the 256-token local model limit. No conversations exceed the 8{,}191-token limit of \texttt{text-embedding-3-large} (\cref{fig:token_mean_by_view}). %

\subsection{Dimensionality Reduction and Clustering}
\label{sec:appendix_clustering}

\paragraph{Dimensionality reduction.} We project the 3{,}072-dimensional embeddings to 2D using UMAP \citep{mcinnesUMAP2020} with $k = 15$ nearest neighbours, $\text{min\_dist} = 0.1$, and cosine distance. Projections are computed jointly on combined (human $+$ simulated) embedding matrices to enable direct visual comparison.

\begin{figure}[!t]
    \centering
    \includegraphics[width=\linewidth]{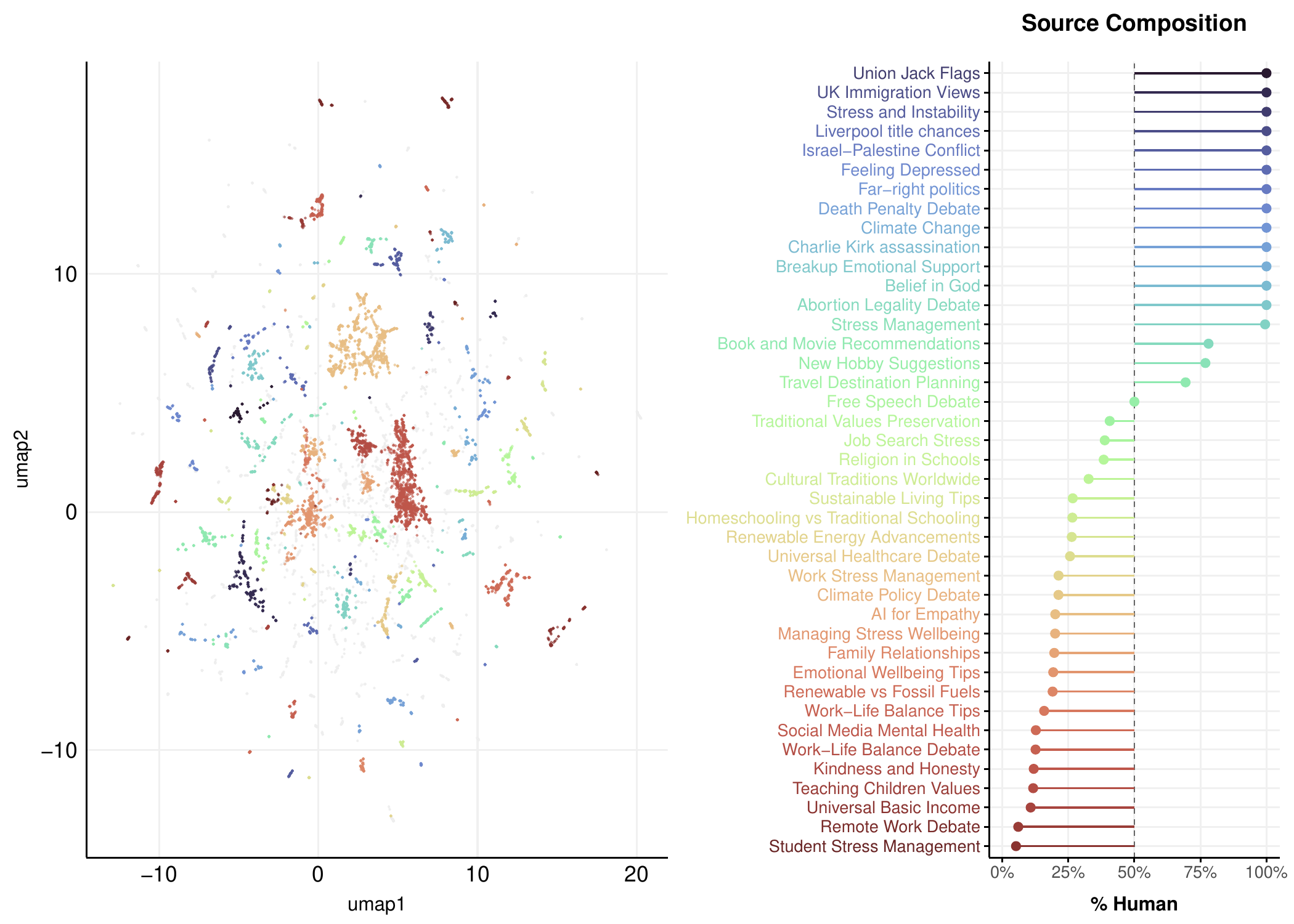}
    \caption{\textbf{Granular topic clusters with source composition.} Left: UMAP of first-user-turn embeddings coloured by sim-human percentage. Right: percentage of human conversations per cluster. Bluer colors indicate higher human \%, Redder colors indicate lower human \%.}
    \label{fig:umap_clusters_granular}
\end{figure}

\paragraph{Clustering.} To identify topical structure, we apply HDBSCAN \citep{campelloDensityBased2013} to a 20-dimensional UMAP pre-projection of the combined first-user-turn embeddings ($k = 15$ neighbours, $\text{min\_dist} = 0$, cosine distance). We fit two granularities:
\begin{itemize}[itemsep=1pt, parsep=0pt, topsep=2pt]
    \item \textbf{Coarse:} $\text{min\_cluster\_size} = \max(10, \lfloor 0.01 \cdot N \rfloor)$, yielding broad topic groups (used in the main paper).
    \item \textbf{Granular:} $\text{min\_cluster\_size} = \max(10, \lfloor 0.005 \cdot N \rfloor)$, yielding finer-grained sub-topics.
\end{itemize}
Each cluster is automatically labelled by GPT-5.4 using the top 10 TF-IDF terms and the 3 centroid-nearest texts (temperature 0). \Cref{fig:umap_clusters_granular} replicates the coarse clustering shown in the main paper at the granular resolution.

\subsection{Cosine Distance Experiments}
\label{sec:appendix_cosine}

\begin{figure}[H]
    \centering
    \includegraphics[width=\linewidth]{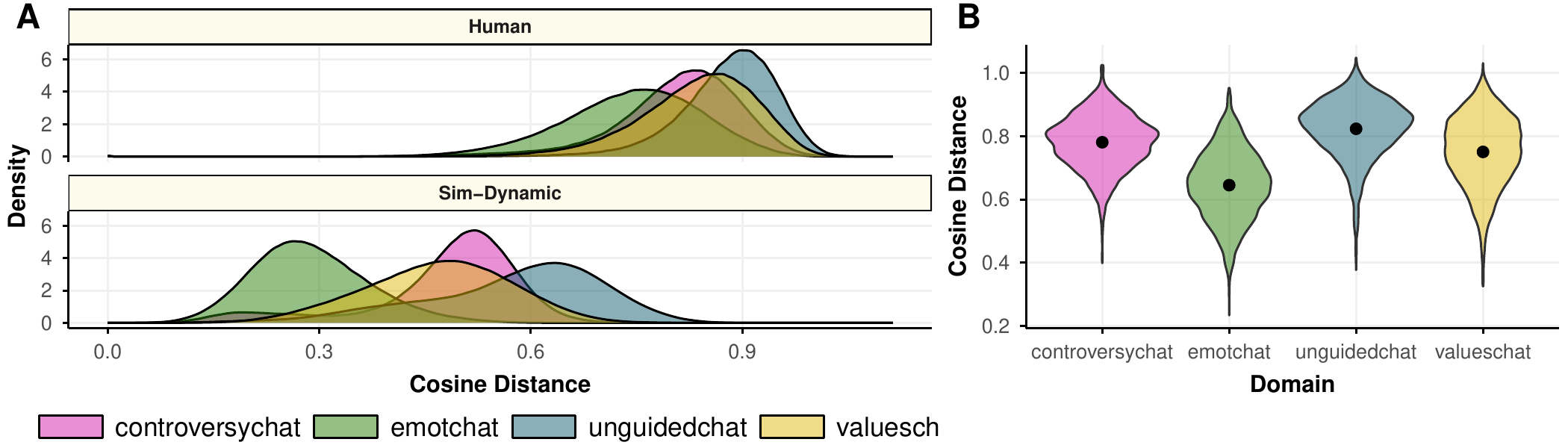}
    \caption{\textbf{Panel A}: Density plots of pairwise cosine distance between different users within each domain, separately for human and simulated conversations. \textbf{Panel B}: For each participant $\times$ domain pair, matched cosine distance (with mean $\pm$ 95\% CI) between the human and simulated conversation embeddings.}
    \label{fig:cosine_combined}
\end{figure}

All cosine analyses use the \emph{all user turns} view (concatenated user messages), which captures user-side conversational style while excluding model-generated text. We run two analyses:
\begin{itemize}
    \item \textbf{Across-user distance by domain.} (\cref{fig:cosine_combined}A) Within each domain $\times$ source combination, we compute the pairwise cosine distance between all pairs of different users. In the main paper, we test whether these distributions differ between human and simulated conversations using two-sample Kolmogorov--Smirnov tests per domain.
    \item \textbf{Matched human--simulated comparison.} (\cref{fig:cosine_combined}B) For each participant and domain, we compute cosine distance between the human conversation embedding and the corresponding simulated conversation embedding.
\end{itemize}

\normalsize

\section{Model Training Details}\label{app:training-details}

\subsection{Training Hyperparameters}\label{app:hyperparameters}

\Cref{tab:hyperparameters} summarises the key training hyperparameters for the PPFT and DPFT models. Both models share the same base LLM (Llama 3.1-8B-Instruct) and training configuration; the only difference is that DPFT does not use the user model and trains with vanilla DPO.

\begin{table}[h]
\centering
\scriptsize
\caption{Training hyperparameters for preference fine-tuning.}
\label{tab:hyperparameters}
\begin{tabular}{ll}
\toprule
\textbf{Hyperparameter} & \textbf{Value} \\
\midrule
\multicolumn{2}{l}{\textit{Optimization}} \\
Learning rate & $5 \times 10^{-5}$ \\
LR scheduler & Cosine \\
Warmup steps & 150 \\
Training epochs & 1 \\
Per-device batch size & 1 \\
Gradient accumulation steps & 16 \\
Effective batch size & 32 \\
Precision & bf16 \\
\midrule
\multicolumn{2}{l}{\textit{P-DPO Objective}} \\
$\alpha$ (user-specific vs.\ agnostic) & 0.5 \\
$\beta$ (KL penalty) & 0.5 \\
\midrule
\multicolumn{2}{l}{\textit{LoRA Configuration}} \\
Rank ($r$) & 8 \\
Alpha ($\alpha_{\text{LoRA}}$) & 16 \\
Dropout & 0.05 \\
Target modules & \texttt{q\_proj, v\_proj, k\_proj, out\_proj,fc\_in, fc\_out, wte} \\
\midrule
\multicolumn{2}{l}{\textit{User Model (PPFT only)}} \\
User model type & Cluster \\
Number of user clusters ($K$) & 10 \\
Number of user tokens ($T_u$) & 10 \\
Add generic user embedding & True \\
\midrule
\multicolumn{2}{l}{\textit{Sequence Lengths}} \\
Max prompt length (tokens) & 1200 \\
Max prompt text length (chars) & 2400 \\
Max total length (tokens) & 1500 \\
Max total text length (chars) & 3000 \\
\bottomrule
\end{tabular}
\end{table}

\subsection{Personalized LLM System Prompt Templates}\label{app:prompts}

We provide the system prompt templates used by the four models in our experiments. All templates instruct the model to generate concise responses. The templates differ in the amount of user information, ranging from no personalisation to rich user profiles. The \textbf{Basic System Prompt} does not include any user-specific information. It is shared by the Base model, DPFT model, and PPFT model. %
The \textbf{Demographics-driven template} incorporates the user's demographic information into the system prompt. It is used by the Demographics-driven prompting method described in \cref{sec:personalization_methods}.
The \textbf{Summarization-driven template} provides the richest user context by combining three sources of information: the user's demographic data, their self-written preference description, and an LLM-generated user profile summarizing their preferences and behavioral tendencies from past conversations. It is used by the Summarization-driven prompting method (\cref{sec:personalization_methods}).

\begin{aipromptbox}[Basic System Prompt Template]
\begin{mdframed}[style=promptblank, nobreak=true]
\textbf{System Prompt}
\begin{Verbatim}[fontsize=\scriptsize]
You are a conversational assistant. Your goal is to engage in conversations. 
The conversation history is in the input.
Reply to the last user message. **Limit your answer to around 50 words. 
Do not refer to your word limit.**
\end{Verbatim}
\end{mdframed}
\end{aipromptbox}

\begin{aipromptbox}[Demographics-driven System Prompt Template]
\begin{mdframed}[style=promptblank, nobreak=false]
\textbf{System Prompt}
\begin{Verbatim}[fontsize=\scriptsize]
You are a conversational assistant. Your goal is to engage in conversations while 
tailoring your tone and content to the context and the user's preference. 
The conversation history is in the input.
The user has the following demographic information:
<user_demographic>
Reply to the last user message. 
**Limit your answer to around 50 words. Do not refer to your word limit.**
\end{Verbatim}
\end{mdframed}
\end{aipromptbox}

\begin{aipromptbox}[Summarization-driven System Prompt Template]
\begin{mdframed}[style=promptblank, nobreak=false]
\textbf{System Prompt}
\begin{Verbatim}[fontsize=\scriptsize]
You are a conversational assistant. Your goal is to engage in conversations while 
tailoring your tone and content to the context and the user's preference. 
The conversation history is in the input.
The user has the following demographic information:
<user_demographic>
The user has the following personal preferences for AI models:
<system_string>
This is a summarized user profile that provides a detailed summary of the user's 
personal information, preferences, and selection tendencies in past conversations,
which you can use as a reference:
<summary>
Reply to the last user message without mentioning anything about the user description.
**Limit your answer to around 50 words. Do not refer to your word limit.**
\end{Verbatim}
\end{mdframed}
\end{aipromptbox}

\newpage
\subsection{Summariser LLM Prompt Template}\label{app:summarizer-prompt}

To generate user profiles for the Summarization-driven method, we employ a capable LLM (GPT-4o) to synthesize each user's demographic data, self-written preference descriptions, and conversation history into a concise user portrait. The following system prompt guides the summariser to produce a coherent, paragraph-form profile that captures both explicit preferences and implicit behavioural patterns.

\begin{aipromptbox}[Summariser LLM System Prompt Template]
\begin{mdframed}[style=promptblank, nobreak=false]
\textbf{System Prompt}
\begin{Verbatim}[fontsize=\scriptsize]
**User Portrait Generation Instructions**
Please generate a comprehensive user portrait based on the following structured data,
focusing on the user's preference characteristics when choosing AI answers.
Please use natural and coherent paragraph descriptions and avoid list format.
---
**Data Processing Requirements**:
1. **Basic Attribute Analysis**:
   - Extract explicit features in demographics such as
   **age**, **gender**, **education**, etc.
   - Combine the **self_description** to extract core values and worldviews.
   - Analyze the style preferences in **system_string** to
   understand how the user wants the AI to communicate.

2. **Conversation History Analysis**:
   - Compare each group of **chosen_utterance** and **rejected_utterance** to identify 
   the user's content preferences. This includes understanding their
   preference for structure, position, or information density.
   - Pay particular attention to the **open_feedback** provided by the user to
   identify additional expectations and dissatisfaction patterns.

3. **Comprehensive Pattern Recognition**:
   - Establish associations between the user's explicit
   preferences and implicit tendencies.
   - Identify stable selection patterns across various conversations.
   - Infer unstated or hidden preferences based on behavior and feedback.
---
**Output Specifications**:
- **Tone**: Use an academic neutral tone to maintain objectivity.
- **Clarity**: Distinguish between objective facts and inferred conclusions clearly.
- **Key Features**: Emphasize important features with **bold** text for visibility.
- **Length**: No more than 150 words per paragraph to keep the summary concise.
---
**Example Output Structure**:
The user is a **mature male** who values **collaborative principles** and 
demonstrates a strong **preference for **clarity** and **factuality** in AI responses. 
He tends to select answers that are **structured, neutral, and objective**, 
with a particular emphasis on **detailed, fact-based content**. 
This pattern suggests a **preference for well-reasoned arguments** over superficial or
opinion-based responses. The user shows significant tolerance for answers with 
**clear structure and accuracy**, but his **tolerance is reduced** when responses **lack
coherence or use negative language**, as indicated by past feedback. 
Implicit needs may include a **preference for unbiased, professional tones** with clear,
**grammatical correctness** in language.
---
**Current Data to Be Processed:**
\end{Verbatim}
\end{mdframed}
\end{aipromptbox}

\subsection{Examples of User Information from the PRISM Dataset}\label{app:user-info-examples}

To illustrate the user information available in the \ourdata dataset and how it is processed by our methods, we provide examples from a real user. For this user, we show three levels of information: (1) the raw demographic data used by the Demographics-driven method, (2) the user's self-written preference description included in the \ourdata dataset, and (3) the LLM-generated user profile produced by our summariser (Section~\cref{app:summarizer-prompt}) for the Summarization-driven method.

\begin{aipromptbox}[User Example]
\begin{mdframed}[style=promptblank, nobreak=false]
\textbf{Demographics ($u^d$)}
\begin{Verbatim}[fontsize=\scriptsize]
age: 55-64 years old; gender: Male; education: Vocational;
employment: Retired; marital: Never been married;
english_proficiency: Native speaker;
\end{Verbatim}
\end{mdframed}
\begin{mdframed}[style=promptblank, nobreak=false]
\textbf{Self-written Preference Description ($u^s$)}
\begin{Verbatim}[fontsize=\scriptsize]
I believe it's important for people in a group to work together to
achieve something. People are stronger when in unity, and I believe
it's good when they collaborate to achieve their goals, as it's much
more likely that they'll actually accomplish them with teamwork.
\end{Verbatim}
\end{mdframed}
\begin{mdframed}[style=promptblank, nobreak=false]
\textbf{LLM-generated User Profile ($p_u$)}
\begin{Verbatim}[fontsize=\scriptsize]
The user is a **55-64 years old male** with a **vocational education**
background who is currently **retired** and has **never been married**.
He describes himself as someone who values **collaborative principles**,
emphasizing the strength and success that come from teamwork and unity.
This belief reflects a worldview that prioritizes **cooperation** and
**collective achievement** over individualism. His preference for AI
responses highlights a desire for **factuality**, **personalisation**,
and clarity in communication. He expects responses to be **well-structured**,
with an emphasis on **cohesive grammar** and avoidance of negative language
or slang, mirroring his expectations for professionalism and respect in discourse.

Analyzing the user's conversation history reveals a consistent selection pattern
where he favors responses that are **informative and concise**, with a clear
presentation of facts. His chosen utterances frequently include structured replies
that provide relevant and straightforward information, while he tends to reject
responses that are overly verbose or that stray into subjective opinions. The
feedback provided indicates that he appreciates interactions where the AI remains
**neutral and fact-based**, particularly on topics that may be contentious or
sensitive. This suggests an underlying preference for discussions that maintain
**objectivity** without compromising on the depth of information presented,
reinforcing his values around cooperation and clarity. In essence, the user seeks
a communicative style that embodies professionalism, factual integrity, and a
respectful tone.
\end{Verbatim}
\end{mdframed}
\end{aipromptbox}

\end{document}